\documentclass[journal]{ieeetran}  

\usepackage{graphicx}
\graphicspath{{figs/}}
\usepackage{afterpage}
\usepackage{amsmath}
\usepackage{amssymb}
\usepackage{algorithm}
\usepackage{algpseudocode}
\usepackage{bm}
\usepackage{caption}
\usepackage{color}
\usepackage[colorinlistoftodos,prependcaption,textsize=tiny]{todonotes}
\usepackage{enumitem}
\usepackage{makecell}
\usepackage{mathtools}
\mathtoolsset{showonlyrefs=false} 
\usepackage{overpic}
\usepackage{physics}
\usepackage{siunitx}
\usepackage{subcaption}
\usepackage{times}
\usepackage{url}
\allowdisplaybreaks

\newcommand{\equref}[1]{\eqref{#1}}
\newcommand{\figref}[1]{Fig.~\ref{#1}}
\newcommand{\tabref}[1]{Table~\ref{#1}}
\newcommand{\secref}[1]{Sec.~\ref{#1}}

\newcommand{\subsecref}[1]{Sec.~\ref{#1}}

\newif\ifrevise
\revisefalse 
\newcommand{\revise}[1]{\textcolor{\ifrevise blue\else black\fi}{#1}}

\newcommand{\red}[1]{\textcolor{red}{#1}}

\def\paperlanguage{}

\newcommand{\switchlanguage}[2]{%
  \ifx\paperlanguage\empty%
  #1%
  \else%
  #2%
  \fi%
}

\title{Design, Control, and Motion \revise{Strategy} for DELTA: \revise{Transformable} Multilink Multirotor for Air-Ground Hybrid Locomotion and Manipulation}

\author{Kazuki Sugihara$^{1}$, Moju Zhao$^{2}$, Takuzumi Nishio$^{1}$, Kei Okada$^{1}$, and Masayuki Inaba$^{1}$
\thanks{$^{1}$Kazuki Sugihara, Takuzumi Nishio, Kei Okada, and Masayuki Inaba are with Department of Mechano-Info\revise{r}matics, The University of Tokyo, 7-3-1 Hongo, Bunkyo-ku, Tokyo 113-8656, Japan.
{\tt\footnotesize sugihara@jsk.imi.i.u-tokyo.ac.jp}}%
\thanks{$^{2}$Moju Zhao is with Department of Mechanical Engineering, The University of Tokyo, 7-3-1 Hongo, Bunkyo-ku, Tokyo 113-8656, Japan.}%
}

\begin{document}

\maketitle
\thispagestyle{empty}
\pagestyle{empty}

\begin{abstract}
In recent years, multimodal locomotion capabilit\revise{ies} have enabled robots to maneuver in both terrestrial and aerial domains.
However, most of these robots are designed only for locomotion, and few \revise{possess the} manipulation \revise{capabilities} required for practical tasks.
By adding a manipulator, ground robots can perform manipulation, and some drones with robotic arm\revise{s} have demonstrated aerial manipulation.
Nonetheless, such multirotors cannot be directly used for manipulation on the ground, and this configuration itself is unsuitable for air-ground hybrid locomotion.
This is because their \revise{thruster}-centralized structure makes it difficult to achieve both sufficient \revise{degrees of freedom (}DoF\revise{)} for manipulation and \revise{stable} motion with contact and \revise{transformation}.
Therefore, in this work, we develop a new multilink multirotor with \revise{thrusters} \revise{on} each link \revise{and} capable of contact with \revise{the} environments.
This robot can perform terrestrial \revise{rolling locomotion}, aerial \revise{flight locomotion}, and manipulation in multiple environments using joint actuation.
\revise{First}, we introduce a minim\revise{al} configuration design of the proposed robot.
\revise{We also describe a kinematic} model \revise{and propose} a design for each component based on this model.
Second, we propose a real-time control method \revise{based on nonlinear optimization that} consider\revise{s} contact and joint motion, which can be applied to various multirotors.
Third, we propose motion \revise{strategies} \revise{that} include contact constraints specific to air-ground hybrid multilink multirotor\revise{s}, \revise{and analyze the limitations of manipulation capabilities based on multi-contact model}.
Finally, we demonstrate \revise{a variety of} motions in \revise{both} domains using the implemented prototype.
To the best of our knowledge, this is the first \revise{demonstration of} air-ground hybrid locomotion and manipulation by a multilink multirotor.



\end{abstract}

\section{Introduction}
\begin{figure}[t]
    \centering
    \includegraphics[width=1.0\columnwidth]{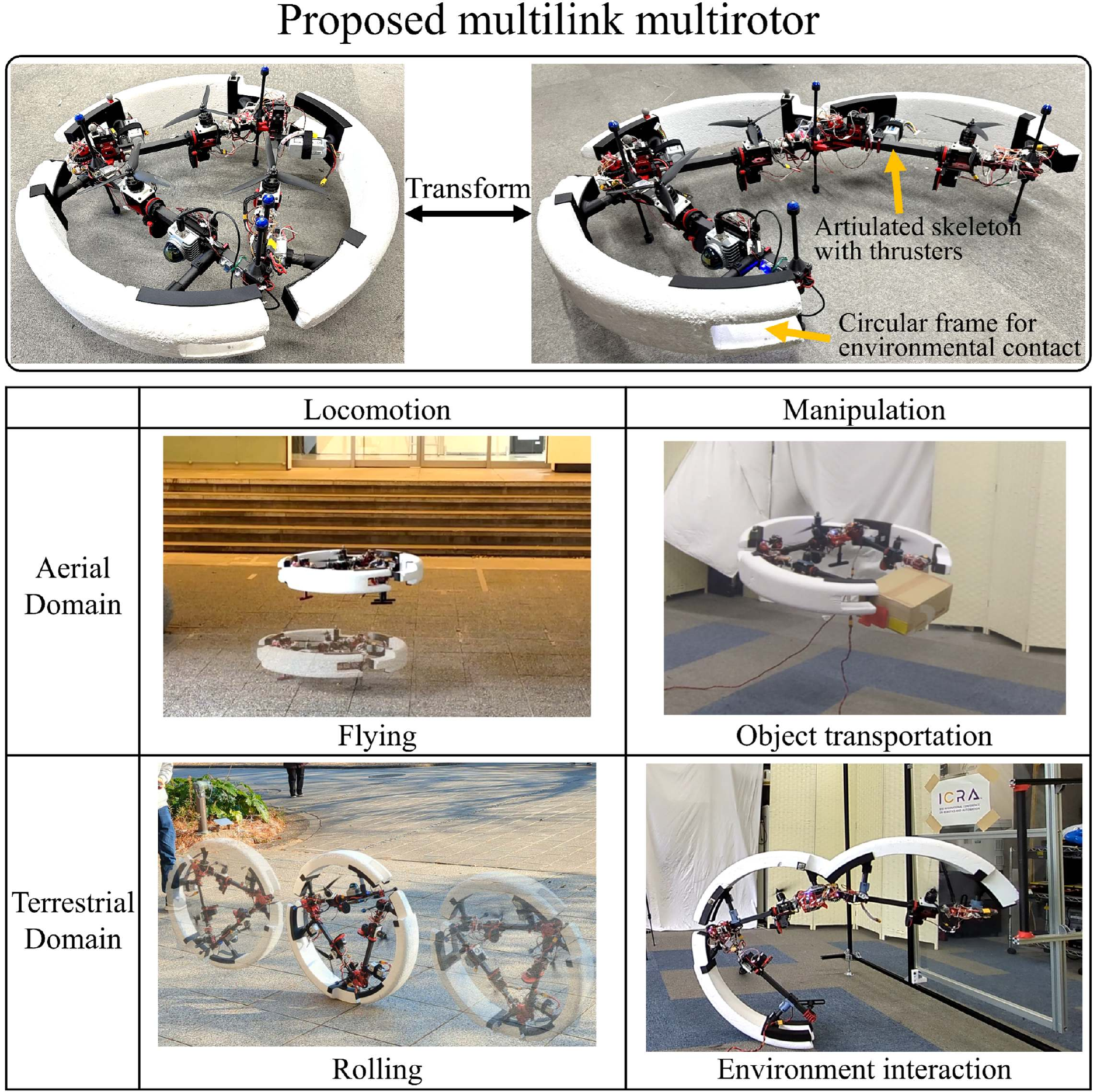}
    \caption{\textbf{DELTA}--\revise{Multilinke\textbf{D} transformabl\revise{\textbf{E}}} multirotor for \textbf{L}ocomotion and manipulation in \textbf{T}errestrial and \textbf{A}erial domains-- achieves air-ground hybrid locomotion and manipulation.}
    \vspace{-5mm}
    \label{overview}
\end{figure}
\IEEEPARstart{I}{N} recent years,
\switchlanguage{
multimodal locomotion in robotics has been investigated \cite{hamzeh2018looncopter,kawasaki2013muwa,crespi2013salamandra,kim2021bipedal,kalantari2020drivocopter}, and various applications such as exploration are expected.
\revise{In particular,} \revise{several novel} air-ground hybrid locomotion \revise{mechanisms} have been proposed \cite{kawasaki2013muwa, kim2021bipedal,page2014uavugv}.
However, most \revise{existing} multirotor platforms with multimodal \revise{locomotion capabilities lack} manipulation capabilit\revise{ies}, which \revise{are} essential for many practical tasks.
Robots with locomotion and manipulation capabilities in both domains can \revise{serve as} versatile robotic platform\revise{s} that \revise{mitigate} disadvantages \revise{such as} low mobility \revise{of} ground robots and high energy consumption \revise{of} aerial robots, as well as applications in multiple environments.

By adding manipulators, ground robots can perform manipulation tasks \cite{bohren2011pr2,ferrolho2023roloma}, and some drones equipped with arms have demonstrated aerial manipulation \cite{kim2013aerial,orsag2017dexterous}.
However, such multirotor platform\revise{s} cannot be directly used for terrestrial manipulation, and this configuration itself is unsuitable for air-ground hybrid locomotion.
This is because the\revise{y have a} \revise{thruster}-centralized structure, \revise{in} which all thrusters are mounted on the root.
When \revise{thrusters} are \revise{centralized}, it \revise{becomes} difficult to compensate the \revise{wrench} caused by external forces during aerial manipulation and to stabilize \revise{the robot} by keeping the \revise{center of gravity (}CoG\revise{)} in the support area of the thrusters.
Furthermore, during terrestrial manipulation, external forces from the environment and the mass of the manipulator segment create a large load on the joints near the root.
For these reasons, we focus on a multilink multirotor with \revise{thruster on} each link \cite{HYDRUS:Chou:IJRR2018,zhao2023spidar}.
\revise{This structure} makes it easier to surround the CoG with thrusters, allows a wide range of \revise{transformation}, and enables \revise{joint loads to be} reduce\revise{d} and distribute\revise{d} \revise{through thrust allocation}.
}{
近年, ロボティクスにおけるmultimodal locomotionの研究が盛んで, 災害現場での探索など様々な環境における活用が期待されている\cite{Jung2018hubo,hamzeh2018looncopter,kawasaki2013muwa,crespi2013salamandra,kalantari2020drivocopter}.
地上と空中における移動に着目すると, 色々なmechanismsが提案されている\cite{kim2021bipedal,page2014uavugv,kawasaki2013muwa}.
空中でも移動できるロボットは飛行による三次元的な移動により, 地上移動ロボットと比較して移動可能な空間を大きく広げることができるという利点がある.
また, ロボットにおける重要な能力として移動能力の他に操作能力が挙げられる.
脚ロボットや移動台車ロボットによる地上領域での操作行動が研究されている \cite{murooka2021locomanipulation,ferrolho2023roloma,bohren2011pr2}.
さらに, ロボットアームを搭載したドローン\cite{kim2013aerial}や多リンク構造をもつ飛行マニピュレータ\cite{zhao2022DragonValve}による, 空中領域での操作行動も実現された.
ロコモーションとマニピュレーションを地上と空中の両領域で実現できるロボットは, 災害現場や工事現場などにおける様々な応用があるだけでなく, 地上ロボットの機動力の低さや, 飛行ロボットの大きな消費エネルギといった双方の欠点を解決可能なプラットフォームになりうる.

一方で, 陸空で移動を実現した例があるにもかかわらず, 両領域で操作能力も実現した例はほとんど存在しない.
なぜなら, 従来の陸空両用マルチロータは, 根本のリンクにスラスタを集約配置する構成をとっているからである.
ロータが集約していると, 空中操作時には外力に起因する回転モーメントを補償することや, 操作時に常に重心をスラスタの支持領域にいれて飛行制御を安定化させることが難しい.
さらに, 地上操作時には環境からの外力やマニピュレータ部分の質量が根本の関節に大きな負荷を生み出してしまう.
そこで, 本研究では多リンク構造にスラスタを分散配置させたマルチリンク型マルチロータに着目し, これにより陸空での移動と操作の両立を目指す.
多リンクマルチロータは, 各リンクにスラスタを分散配置しており, 重心をロータで囲みやすく可能な変形の幅が広く, さらに各リンクの推力で関節負荷を分散させることができる.
}

{
\tabcolsep = 1pt
\begin{table*}[t]
 \centering
 \begin{tabular}[t]{|c|c|c|c|c|c|}
  \hline & Hydrus \cite{HYDRUS:Chou:IJRR2018,anzai2018halo,sugito2022contactmanipulation} & MUWA \cite{kawasaki2013muwa} & HyTaq \cite{kalantari2013hytaq} & SPIDAR \cite{zhao2023spidar,zhao2025grasping} & Ours\\ \hline
  \makecell{Concept} & \makecell{Multilink multirotor} & \makecell{Rollable quadrotor with \\rigid circular frame} & \makecell{Rollable quadrotor\\ with cage} & \makecell{Thruster-distributed\\ quadrupedal robot} & \makecell{Multilink rollable \\multirotor} \\ \hline
  \makecell{Design} & \makecell{Articulated structured\\ multirotor} & \makecell{Variable pitch \\propellers deployed\\ in circular frame} & \makecell{Quadrotor deployed \\in passively rolling\\cylindrical cage} &\makecell{Quadruped robot\\with 2 DoF\\ vectorable thrusters}& \makecell{Multilink multirotor with a small\\ number of links equipped with\\ contact frames and 1 DoF thrusters}\\ \hline
  \makecell{Control} & \makecell{Quadrotor-based flight\\ control for a \\time-varying model} & \makecell{Control allocation \\with ground contact} & \makecell{Quadrotor-based\\ control} & \makecell{Linearized thrust vectoring\\ allocation with contact \\and joint torque constraint} &\makecell{Nonlinear thrust vectoring\\ allocation considering contact\\ and joint torque constraint} \\  \hline
  \makecell{Motion\\ strategy\\ / \\Analysis} & \makecell[l]{\parbox{3.5cm}{\begin{itemize}[leftmargin=*] \item Geometric analysis of object grasping \item Transform planning for object transportation \item Inverse kinematics \end{itemize}}} & \makecell{\parbox{3cm}{\begin{itemize}[leftmargin=*]  \item Predefined attitude trajectory generation \end{itemize}}}& \makecell{\parbox{2cm}{\begin{itemize}[leftmargin=*]  \item Energy efficiency analysis \end{itemize}}} & \makecell{\parbox{3cm}{\begin{itemize}[leftmargin=*] \item Preplanned footsteps \item  Grasping force planning in null-space of the flight control\end{itemize}}} & \makecell[l]{\parbox{4cm}{\begin{itemize}[leftmargin=*] \item Mode switching strategies. \item Inverse kinematics with contact constraint \item Analysis of manipulation limitations based on multi-contact model and control allocation\end{itemize}}}\\ \hline
  \makecell{Locomotion} & Flight & Flight, Rolling & Flight, Rolling & Flight, Walking & Flight, Rolling\\\hline
  \makecell{Manipulation} & \makecell{\parbox{3.5cm}{\begin{itemize}[leftmargin=*] \item Object grasping \item Environmental interaction \end{itemize}}} & \makecell{N/A} & \makecell{N/A} & \makecell{\parbox{3cm}{\begin{itemize}[leftmargin=*] \item Object grasping \end{itemize}}} & \makecell{\parbox{4cm}{\begin{itemize}[leftmargin=*] \item Object grasping \item Environmental interaction \end{itemize}}}\\\hline
 \end{tabular}
 \caption{\revise{Comparison between existing multirotor platforms and this work.}}
 \label{table:comparison}
 \vspace{-5mm}
\end{table*}
}

\switchlanguage{
Therefore, in this work, \revise{we} develop a multilink multirotor platform capable of air-ground hybrid locomotion and manipulation\revise{.}
\revise{W}e propose a minimal configuration design and control method \revise{that} consider\revise{s} thrust, contact and joint motion.
\revise{The proposed component} design and control methodolog\revise{ies} \revise{are} applicable to other \revise{multirotor designs with} multiple \revise{Degrees of Freedom (}DoF\revise{)}.
We develop a multilink multirotor with \revise{a} contact function and realize terrestrial locomotion by rolling, aerial locomotion by flight, and manipulation in multiple domains by joint actuation as shown in \figref{overview}.
}{
したがって, 本研究では, このような多リンク型マルチロータを構築するための, 今後多自由度構成にも応用可能なminimal designと, 接触と変形を考慮した一般化可能な動作制御手法を明らかにすることを目的とする.
関節構造を有し全身で環境接触可能なマルチリンク型マルチロータを開発し, 転がりによる地上移動と飛行による空中移動および, 関節変形を利用した多環境でのマニピュレーションを実現する as shown in \figref{overview}.
}

\subsection{Related Works}
\label{section:related_works}
\switchlanguage{
In this part, we introduce related works and \revise{outline} our approach \revise{to the} design, control and motion \revise{strategy} of \revise{the} proposed robot.
Then, we \revise{summarize} key contributions.
}{
この節では, 関連研究について述べ, 機体構成, 動作制御, 動作計画において本研究で採用するアプローチを述べる.
}
\subsubsection{Air-Ground Hybrid \revise{Robot Design}}
\label{section:multimodal_locomotion}
\switchlanguage{
Robots \revise{capable} of locomotion in both terrestrial and aerial domains \revise{can be broadly categorized into} legged, wheeled, and rolling types.
Legged \revise{types include} bipedal \cite{kim2021bipedal,anzai2021design,sugihara2023flyinghumanoid} and quadrupedal \cite{zhao2023spidar} \revise{systems, where thrust was used to assist walking or enhance stability.}
Robots equipped with wheels and thrusters have also been proposed \cite{page2014uavugv, cao2023doublebee, sihite2023multi, meiri2019flyingstar}.
\revise{Some designs employed} active wheels driven \revise{by dedicated actuators} \cite{cao2023doublebee}.
\revise{In other designs, passive wheels or reduction gears were introduced, and the power of thrust motors was used for wheeling, eliminating the need for additional actuators for terrestrial locomotion \cite{page2014uavugv,meiri2019flyingstar}}.
There are also some designs that integrate flying and rolling capabilities \cite{kawasaki2013muwa, kalantari2013hytaq, zhi2023rollerquadrotor}.
For example, HyTAQ in \cite{kalantari2013hytaq} is a quadrotor equipped with a cage that can freely rotate with respect to the body and also serves as a propeller protector.
In \cite{zhi2023rollerquadrotor}, \revise{instead of surrounding the entire quadrotor,} a rotatable circular plate was mounted for rolling.
\revise{In \cite{kawasaki2013muwa}}, a fixed circular frame was used for ground contact, and \revise{the} drag moment of \revise{the built-in} propeller\revise{s} generated \revise{moment} for rolling.

Comparing these locomotion \revise{mechanisms}, legged \revise{robots} tend to be large \revise{and heavy} \revise{for} their size due to \revise{their high number of control DoF}.
\revise{In contrast}, wheel\revise{ing} and rolling, \revise{which are based on rotational motion}, can achieve relatively high\revise{-}speed movement compared to legged locomotion.
Moreover, in the case of wheeled robots, \revise{the} motors for the thrusters can \revise{contribute} to \revise{multiple locomotion modes}, \revise{which} enables lighter design.
Rolling robots \revise{often} employ a mechanism that simultaneously provides propeller protection and environmental contact function, which simplifies the structure.

\revise{In this work, we design an air-ground hybrid robot based on a multilink multirotor.
Motivated by the characteristics discussed above, we attach lightweight frames to each link that provide both propeller protection and ground-contact functionality, and enable terrestrial locomotion by rolling.
This concept allows us to minimize the required DoF while maintaining high rigidity and low mass, without the many DoF needed in legged systems.
Furthermore, the continuous contact afforded by rolling can be exploited as an additional DoF during terrestrial manipulation tasks, which is a key feature of the proposed platform.}
}{
地上と空中で移動能力を実現可能なロボットは主に脚型, 車輪型, 転がり移動型に分類される.

脚型のロボットには, 二脚型\cite{kim2021bipedal,anzai2021design,sugihara2023flyinghumanoid}や四脚型\cite{zhao2023spidar}が提案されている.
\cite{kim2021bipedal}では, 二本の平行リンク脚とクアッドロータを組み合わせたロボットが推力を利用して歩行した.
また, \cite{anzai2021design}や\cite{sugihara2023flyinghumanoid}では, 全身の自由度を有するヒューマノイドに飛行モジュールを搭載している.
\cite{zhao2023spidar}で提案された四脚ロボットは, 全リムに分散配置された推力装置を利用した歩行をした.

また, 車輪型のロボットも存在する\cite{ page2014uavugv, cao2023doublebee, sihite2023multi,meiri2019flyingstar}.
\cite{cao2023doublebee}では, 飛行のためのバイロータと走行のための能動輪を組み合わせている.
能動車輪ではなく, 受動車輪と推力を利用し, 車輪走行に追加のアクチュエータを必要としない構成もある\cite{page2014uavugv}.
\cite{meiri2019flyingstar}では車輪走行時に減速機を介してプロペラ駆動用のモータで車輪を能動的に駆動している.
車輪の内側にプロペラを搭載し, このプロペラ内蔵車輪の向きを変えることによって2輪, 4輪走行と飛行を両立している例\cite{sihite2023multi}もある.

車輪走行と同様に回転運動に基づく, 転がり移動と飛行能力を統合した素晴らしいデザインのロボットも存在する\cite{kalantari2013hytaq, zhi2023rollerquadrotor,kawasaki2013muwa}.
\cite{kalantari2013hytaq}はこのタイプのロボットのパイオニアであり,四つのプロペラを持つクアッドロータに回転可能なケージを搭載し, 地上での転がり移動を実現している.
このケージはクアッドロータに対して自由に回転可能であり, プロペラの保護としての機能も持っている.
\cite{zhi2023rollerquadrotor}では, ドローン全体を囲むのではなく, 回転可能な円形プレートを搭載したクアッドロータによる転がり移動が提案された.
また, \cite{kawasaki2013muwa}では固定フレームに可変ピッチプロペラを搭載し, 推力方向の変化によるバランス制御とプロペラ反トルクによる転がりが提案されている.

これらを比較すると, 
脚歩行型はその運動を実現するために多くの関節とリンクを有しており, 身体の大きさに対し質量が大きくなりやすい.
車輪走行や機体全体による転がり動作はその運動が無限回転運動によるものであり, 脚歩行と比較して比較的高速な移動を実現しやすいという特徴がある.
車輪型では, 推力や推力装置用のモータをmultimodal locomotionに利用し軽量化を可能にしている.
転がり型では, 飛行ロボットに必要な回転翼の保護と環境接触機能を一つの機構に担わせている.
これらにより構成の単純化と軽量化を実現している.

これらを踏まえ, マルチリンク型マルチロータの構成で地上ロコモーションをするために,
本研究では軽量な構成にしつつプロペラガード機能と環境接触機能を両立できるフレームを各リンクに搭載し, 転がりによる地上移動を実現する.
これにより, 歩行型のように多くの自由度を必要とせず, 構成を最小化し, 軽量かつ高い剛性で作ることが可能になる.
このデザインは連続的な接触を可能にするため, 接触状態をマニピュレーション時の自由度とすることができるという利点もある.
}

\subsubsection{Thrust Control Considering Contact and \revise{Transformation}}
\label{section:thrust_control_under_contact}
\switchlanguage{
Thrust control methods \revise{that consider} contact or \revise{transformation} have been studied \revise{in the context of both locomotion and manipulation}.
\revise{For} wheeled or rollable multirotors, contact forces \revise{are incorporated into the} kinematic model \cite{kawasaki2013muwa,cao2023doublebee}.
In \revise{thrust-assisted} bipedal walking \cite{kim2021bipedal}, \revise{a} model\revise{-}based method was proposed, but lightweight feet \revise{were used} and the effects of joint \revise{transformation were ignored}.
In \cite{huang2017jet}, static walking \revise{that maintains balance using} thrust for a nonlinear motion model was demonstrated; however, its motion control \revise{problem} cannot be solved in real-time.
In \revise{addition, the q}uadrupedal \revise{thruster}-distributed multirotor in \cite{zhao2023spidar} used thrust to \revise{reduce} joint load\revise{s} during walking, but it\revise{s} \revise{control framework relied on}  high redundancy to \revise{linearize the nonlinear model}, \revise{which limits its generality for other configurations}.

There are also examples where a multirotor \revise{makes} contact \revise{with the environment} for manipulation.
\revise{In \cite{sugito2022contactmanipulation,nishio2023perchingarm}, multilink multirotors established contact with walls or ceilings to stabilize their pose and generate contact forces.}
\revise{However, these approaches treated contact within \revise{a} linearized model and were designed to maintain a fixed contact state.
Such formulations are suitable for aerial manipulation but they cannot be directly applied to locomotion and manipulation scenarios where contact states change over time.}

In this work, we propose a real-time thrust control method for a nonlinear motion model considering contact and joint motion, in order to realize terrestrial locomotion and manipulation.
\revise{The control allocation is formulated as a nonlinear optimization problem that computes both thrust and its vectoring angles while explicitly imposing friction constraints and joint torque limits.}
\revise{Because the formulation is based on the generalized wrench on the CoG, the core of this control framework is applicable to other multirotors, beyond the prototype presented in this paper.}
}{
これまでに, ロコモーションやマニピュレーションのための, 接触や変形を考慮した推力制御手法が提案されている.
マルチロータによる車輪や転がりでの移動の例では接触力を考慮した安定化制御が行われている \cite{cao2023doublebee,kawasaki2013muwa}.
推力による二足歩行のための安定化では, model baseな方法\cite{kim2021bipedal}が提案されているが, 十分軽量な足を使用し, 機体のモデルにおける関節変形の影響を無視している.
\cite{huang2017jet}では非線形な運動モデルに対して推力を利用してバランスを維持しながらの静歩行をしているが実時間での求解はしていない.
\cite{zhao2023spidar}の四脚ロボットでは推力と関節トルクを利用した歩行を行っているが, 非線形な運動モデルに対し, 高い冗長性を利用してモデルを線形化し非線形性を回避している.

また, マニピュレーションのためにマルチロータが環境接触する例もある.
\cite{sugito2022contactmanipulation,nishio2023perchingarm}では, 環境接触することでマニピュレーションの安定化を行っている.
どちらも線形モデル内で接触を扱っているという前述した課題があるのと, 常に同一の接触状態を保つような制御を行っており接触状態の変化するロコモーションには活用できないという課題がある.

そこで, 本研究では, 地上での移動, 操作を実現するために, 環境接触と関節運動を考慮した非線形な運動モデルに対する実時間推力制御手法を提案する.
これは, 骨格構造やロータ配置にによらず任意の機体構成のマルチロータに適用可能な推力制御手法である.
これにより接触を伴うロコモーションとマニピュレーションを実現可能にする.
}

\subsubsection{Motion \revise{Strategies} for Multirotor}
\switchlanguage{
\revise{M}otion \revise{strategies} for multirotors have been \revise{investigated} mainly in the aerial domain.
\revise{For} rigid multirotors, model-based path planning based on environmental recognition \cite{tordesillas2019faster} and path-following flight using reinforcement learning \cite{song2021autonomousRL} have been proposed.
For multilink multirotor\revise{s}, planning method \revise{that consider} not only the whole-body \revise{trajectory} but also joint motion \revise{for collision avoidance in narrow spaces have been} proposed \cite{zhao2020differencialkinematics}.
For aerial manipulation tasks, transform\revise{ation} \revise{strategy} \revise{to maintain} stability \revise{during} object transportation \cite{anzai2018halo} and inverse kinematics \revise{methods} for multilink multirotors \revise{have been} proposed \cite{sugito2022contactmanipulation,nishio2023perchingarm}.
\revise{In contrast to these works, which primarily focused on aerial maneuvering or manipulation}, we propose motion \revise{strategies for a multilink multirotor that explicitly aim} motions in terrestrial domains.
These include a whole\revise{-}body attitude \revise{trajectory generation} to switch \revise{between multiple} locomotion modes for \revise{an} air-ground hybrid multirotor.
We also describe a whole-body inverse kinematics method \revise{that considers} environmental contact constraints for manipulation by a multilink multirotor with minimal DoF configuration.

\revise{
Furthermore, we analyze the theoretical limits of the targeted manipulation tasks.
Previous works have, for example, evaluated graspable object shapes from a geometric viewpoint \cite{HYDRUS:Chou:IJRR2018}, analyzed reachability in aerial manipulation \cite{nishio2023perchingarm}, or planned grasping forces in null-space of \revise{the} flight control \cite{zhao2023spidar}.
In contrast, we combine a multi-contact model with the nonlinear optimization based control allocation, which is the core of the proposed control framework.
This allows us to explicitly consider \revise{the} wrench generated by thrusters, multiple contact forces, and joint torque constraints simultaneously.
Then we clarify the theoretical payload limits and reachability of the manipulation tasks on the ground.
By appropriately specifying task-specific constraints on external forces, the proposed analysis framework can be extended to other manipulation tasks and robot configurations. 
}}{
マルチロータにおける動作計画については, 飛行領域において主に行われてきた.
剛体型マルチロータにおいては, 環境認識に基づくモデルベースなパスプランニング\cite{tordesillas2019faster}や, 強化学習を利用した経路追従飛行\cite{song2021autonomousRL}などが行われてきた.
また, 多リンクマルチロータにおいては, 狭小空間をすり抜けのために, 重心の軌道だけでなく, 関節変形も考慮した計画\cite{zhao2020differencialkinematics}が提案されている.
空中マニピュレーションのためには物体把持飛行を安定化するための変形計画法や\cite{anzai2018halo}や, \cite{sugito2022contactmanipulation, nishio2023perchingarm}では飛行時の逆運動学手法が提案されている.
一方, 本研究では, これまで注目されてこなかったマルチロータによる地上移動・操作行動を実現するための, 動作計画法を提案する.
複数のロコモーションモードを使い分けるための重心軌道の計画法や,
最小構成のマルチリンク型マルチロータにより地上で操作を実現するための, 環境接触維持制約を考慮した, 全身逆運動学を提案する.

本研究ではさらに, 目的とするマニピュレーション動作について, 理論的限界の解析を行う.
把持可能な物体形状を幾何的に評価する手法\cite{HYDRUS:Chou:IJRR2018}や，空中でのリーチャビリティを解析した例\cite{nishio2023perchingarm}は報告されている．
これに対して本研究では，多点接触モデルと，提案する制御手法のコアである非線形最適化に基づくcontrol allocationを組み合わせることで，接触および関節トルクの制約を明示的に考慮しつつ，ペイロードと地上におけるリーチャビリティの理論的限界を明らかにする．
この解析手法は，タスク固有の制約を与えるだけで他のタスクや機体構成にも適用可能な，一般性の高い手法となっている．
}

\subsection{Contributions}
\revise{A comparison between previous works and this work is summarized in \tabref{table:comparison}}.
The key contributions of this work are as follows:
\switchlanguage{
\begin{enumerate}
    \item We propose a minimal \revise{DoF} configuration for a multilink multirotor capable of terrestrial \revise{rolling}, \revise{aerial flight}, and manipulation in both aerial and terrestrial domain\revise{s}. \revise{We also present} design methods for \revise{the} thruster \revise{layout} and \revise{skeleton} joint\revise{s} based on \revise{a} kinematic model.
    \item We propose \revise{a} real-time thrust control \revise{framework} for multilink multirotor\revise{s} under environmental contact.
    \revise{The control allocation is formulated as a nonlinear optimization problem that can compute thrust and thrust vectoring angles while explicitly considering joint torque limits and an arbitrary number of contact forces.
    In addition, this framework \revise{is} applicable to other multirotors.}
    \item \revise{We propose motion strategies for locomotion and manipulation with the minimal DoF rollable multilink multirotor.
    These include standing up motions to switch locomotion modes, and whole\revise{-}body inverse kinematics with environmental contact constraints.
    We also analyze the limitations of manipulation tasks by incorporating \revise{a} multi-contact model and \revise{a} nonlinear control allocation method, which is the core of control framework.
    }
    \item \revise{Based on the above proposed methods, we experimentally demonstrate a variety of motions in aerial and terrestrial domains with an implemented prototype.}
\end{enumerate}
}{
\begin{enumerate}
    \item 地上と空中でロコモーションとマニピュレーションができるマルチリンク型マルチロータの最小構成法と, 運動モデルに基づくスラスタ, 関節の設計手法.
    \item 環境接触力と関節トルクを考慮した非線形性のあるマルチリンク型マルチロータの運動モデルに対する, 非線形最適化に基づく実時間での推力制御手法.
    \item 最後に, ロコモーション, マニピュレーションを実現するための動作計画について提案する. そして実装したロボットプラットフォームによる地上と空中における多様な動作を実現する.
\end{enumerate}
}


\section{Design and Modeling of the proposed robot}
\label{section:design}
\switchlanguage{
In this section, we introduce \revise{a} minimal configuration design \revise{methodology for} a multilink multirotor \revise{that can perform both} locomotion and manipulation in terrestrial and aerial domains.
\revise{We first discuss the minimal skeleton structure and the number of internal joints.
Next, we consider the control degrees of freedom (DoFs) of the thrusters  from a wrench-controllability viewpoint.}
\revise{Then}, \revise{we describe} the kinematics model.
Finally, \revise{based on this model}, \revise{we derive} \revise{the thruster} arrangement \revise{that improves} control stability and the joint \revise{design requirements}.
}{
本節では、地上と空中の領域におけるロコモーションとマニピュレーションを両立可能なマルチリンク型マルチロータプラットフォームを構築するためのminimal designを紹介する.
地上において転がりによるロコモーションやマニピュレーションを行うための骨格のデザインと推力装置の自由度構成について述べる.
次に運動モデルについて述べ, それに基づき制御安定性を大きくできるようなロータ配置法やリンク間関節の設計要件について述べる.
}

\begin{figure}
    \centering
    \includegraphics[width=1.0\columnwidth]{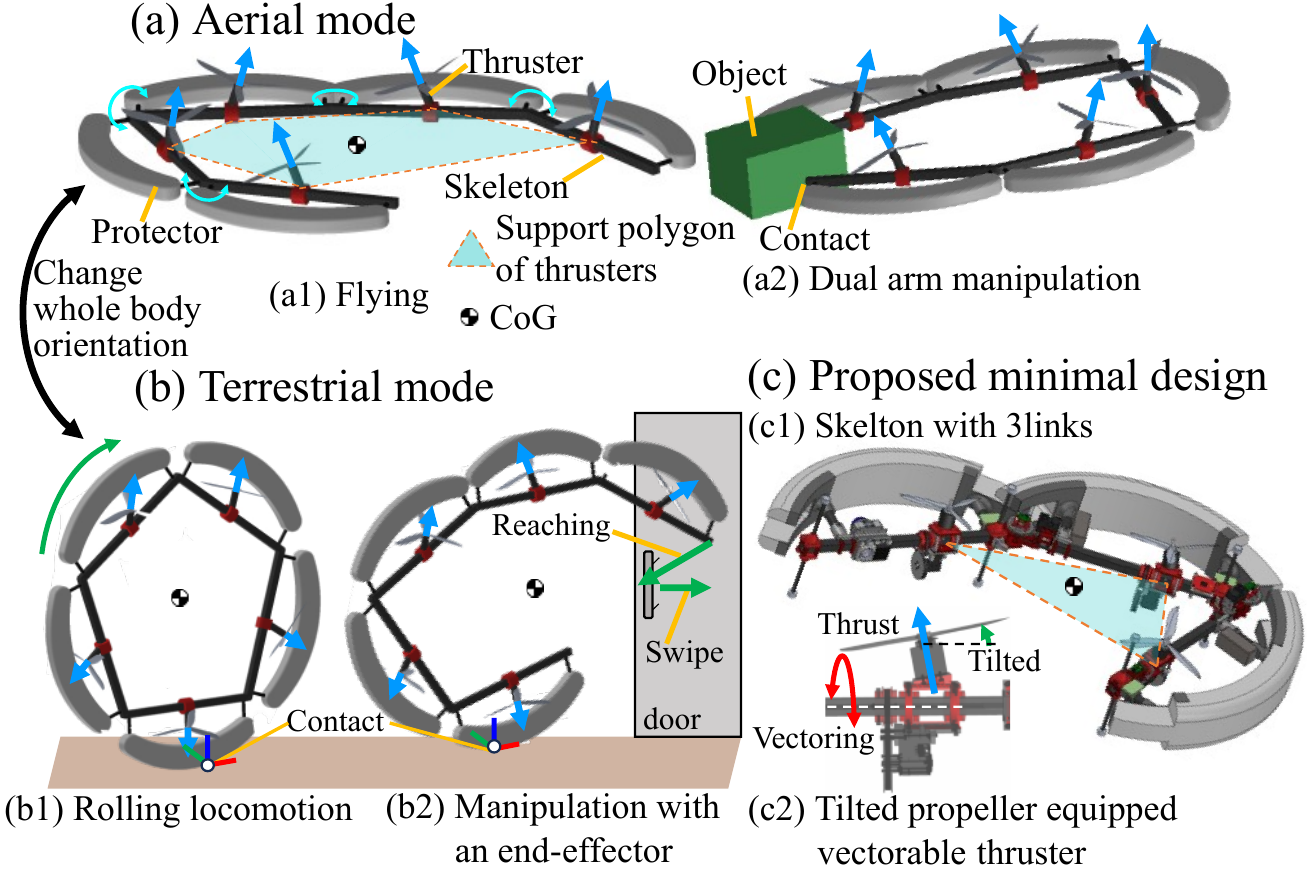}
    \caption{\revise{(a), (b): General} concept of the proposed \revise{air-ground hybrid multilink multirotor}. (a): Aerial mode. All links are \revise{on} horizontal plane and the CoG is within the support polygon area of the \revise{thrusters}. (b): Terrestrial mode. \revise{All links are on vertical plane.} (c): Proposed minimal design. The robot consists of three links with one thruster on each link. Each propeller is tilted and 1 DoF thrust vectoring mechanism can rotate it around the link.}
    \label{figure:robot_configuration}
    \vspace{-5mm}
\end{figure}

\subsection{Minimal Configuration Design Methodology}
\label{section:design methodology}
\subsubsection{Minimal Skel\revise{e}ton Configuration}
\label{section:link_number}
\switchlanguage{
In this work, we deploy frame\revise{s} with circular \revise{side} shape\revise{s} \revise{to a multilink multirotor so that every link can keep} continuous \revise{rolling} contact \revise{with the ground}, as shown in \figref{figure:robot_configuration}(b).
\revise{In addition, we assume that each link is equipped with one thruster module.}
During \revise{terrestrial} motion, the robot stand on the ground using the\revise{se} frame\revise{s}, and \revise{performs} rolling locomotion and \revise{transformation-based} manipulation, as \revise{illustrated} in \figref{figure:robot_configuration}(b).
\revise{Therefore, we connect the links in series by 1 DoF revolute joints with parallel axes, which ensures continuous ground contact.}

Under this joint design, we first consider the number of links \revise{required to realize rolling} locomotion.
\revise{To keep uninterrupted ground contact during rolling, the links in the skeleton are required to form a polygon.}
\revise{Therefore, at least three links are required to construct such a closed shape with a serial skeleton.}

\revise{Next}, we consider the DoFs of \revise{skeleton joints} for manipulation.
In this work, we aim to realize a configuration \revise{that enables to control of} \revise{the} three-dimensional position of the end-effector.
With this DoF configuration, \revise{the robot can perform} dual-arm manipulation in the air and tasks \revise{composed} of primitives such as \revise{Reaching} and Swiping on the ground, as shown in \figref{figure:robot_configuration}(a2) and (b2).
These motions enable the robot to perform tasks such as object transportation in the air and door opening on the ground.

\revise{In this subsection, we first focus on manipulation in the terrestrial domain, where the robot stands on the ground while the circular frames remain in rolling contact, as shown in \figref{figure:robot_configuration}(b).
Because all internal joint axes are designed to be parallel, the motion of the links in the terrestrial mode is confined to a single plane $\Sigma$ that is orthogonal to the rotational axis of skeleton joints.
The horizontal lateral motion of the robot in the world frame can be generated simply by rotating the whole body around the gravity axis (yaw).
Therefore, for the analysis of the DoFs of skeleton joints, we concentrate on the reachable set of the end-effector in the cross-section $\Sigma$.}

\revise{We denote the number of internal joints by \(N_q\).}
\revise{
Let the in-plane coordinates of the end-effector on $\Sigma$ be denoted by $(x_{ee}, z_{ee})$, where $z_{ee}$ is aligned with gravity and $x_{ee}$ is along the rolling direction.
For a given root pose and ground contact configuration, the end-effector position in this cross-section \(\bm{p}_{ee}^{\Sigma}\) can be written as}
\revise{
\begin{equation*}
    \bm{p}_{ee}^{\Sigma}
    =
    \begin{bmatrix}
    x_{ee} \\ z_{ee}
    \end{bmatrix}
    =
    f_{\Sigma}(\bm{q}) \in \mathbb{R}^{2},
\end{equation*}}
\revise{where $\bm{q} \in \mathbb{R}^{N_q}$ is the vector of internal joint angles.}
\revise{The differential kinematics in this plane is given by}
\begin{equation*}
    \revise{
    \delta \bm{p}_{ee}^{\Sigma}
    = \frac{\partial f_{\Sigma}}{\partial \bm{q}} \delta \bm{q} = J_{\Sigma}\delta \bm{q}, \hspace{3mm} J_{\Sigma} \in \mathbb{R}^{2 \times N_q},}
\end{equation*}
\revise{where $J_{\Sigma}$ is the Jacobian mapping the joint variations to the motion of the end-effector in the cross-section \(\Sigma\).}

\revise{To locally control both the height $z_{ee}$ and the horizontal position $x_{ee}$ of the end-effector in \(\Sigma\), the Jacobian must satisfy}
\begin{equation*}
    \revise{\mathrm{rank}\, J_{\Sigma} = 2.}
\end{equation*}
\revise{Because $\mathrm{rank}\, J_{\Sigma} \le \min(2, N_q)$, this condition implies}
\begin{equation*}
    \revise{N_q \ge 2.}
\end{equation*}
\revise{Intuitively, with only a single joint ($N_q = 1$), the end-effector trajectory for a fixed root pose is restricted to a one-dimensional curve in the $(x_{ee},z_{ee})$ plane, and it is impossible to control these quantities independently.
By introducing at least two joints with parallel axes, the multilink skeleton can generate a two-dimensional reachable region in $\Sigma$ and allows independent adjustment of these quantities.}

\revise{Moreover, if \(N_q \ge 2\), this is also sufficient for the targeted aerial manipulation tasks such as object grasping and transportation.
Combined with at least two internal joints, this allows the end-effectors to be positioned around a payload and their relative configuration to be adjusted as required by grasping and carrying motions.}
}{
本研究では, \figref{figure:robot_configuration}に示すように多リンクマルチロータに対し, 連続的な環境接触を可能にする弧状の側面をもつフレームを搭載する.
地上動作時は機体を立てて接地することが可能で転がり移動と変形によるマニピュレーションが可能となる.
連続的な接触をする機能を果たすための最小構成のため, 各リンクは回転軸が水平な1自由度の関節で接続する.
このもとでロコモーションとマニピュレーションを実現するのに必要な最小リンク数を考える.
まず, 転がり運動による移動を実現するためには, 回転に伴って途切れることなく接地可能である必要がある.
そのため, 骨格をつくるリンクによって多角形を形成可能である必要があるため, 少なくとも3リンク必要である.

そして, マニピュレーションのための自由度を考える.
本研究では操作を行うエンドエフェクタの3次元位置を制御可能な構成を実現し, \figref{figure:robot_configuration}にあるような, 空中での双腕マニピュレーションや, 地上でのPointContactやSwipeといったプリミティブからなる動作を目標とする.
これらにより, 高所物体運搬や, 地上のドア開けといった動作が可能になる.
関節数を\(N_q\)とする.
飛行時にはルートリンクの3次元位置とyaw角を制御可能なため, 自由度空間は\(\mathbb{SE}(2)\times\mathbb{R}\)である. 
さらに\(N_q\)個の関節があるので制御可能変数からエンドエフェクタ位置への写像は
}

\switchlanguage{


}{
ただし, \(\phi_{root}\), \(\theta_{root}\), \(\psi_{root}\)はそれぞれルートリンクのroll角, pitch, yaw角を表す.
\(\bm{q}\)はリンク間関節の角度ベクトルを表す.
環境接触制約のある地上でのマニピュレーションにおいても, \secref{section:ik}で示すルートリンクの自由度を利用したマニピュレーションをすることで, 関節数を少なくすることができる.
したがって, 関節数\(N_q\)は2あれば目標の自由度数を確保可能なため, 本研究では3リンクを1自由度の関節で接続する最小構成を提案する.

マルチリンク型飛行ロボットでは，リンク数が増加するとともに機械剛性が低下していくことが知られている\cite{ElasticVibrationControlforHydrus:MakiToshiya:RAL}.
機体を環境接触させながらの動作を行うため，低剛性な機体では環境から働く力によって意図しない変形や故障が引き起こされ, モデル化誤差が生じ制御不安定性につながる.
よって上述した最小リンク数構成は, 環境との接触を伴う動作を安定に行うのに寄与できると考えられる.
}

\subsubsection{Control DoF of Thrusters}
\label{section:thruster_configuration}
\switchlanguage{
\revise{The proposed robot must generate different force-torque patterns depending on its operation mode.}
During flight, all links are in the horizontal plane so that the CoG of the body is within the support polygon of the \revise{thrusters} as shown in \figref{figure:robot_configuration}(a), \revise{and a large thrust in the anti-gravity direction is required to float the body.}
While during ground motion, the frames mounted on each link contact \revise{with} the ground as shown in \figref{figure:robot_configuration}(b).
And the robot has to exert thrust in each direction to keep balance for rolling locomotion and manipulation.
\revise{Hence}, the robot needs to exert force and torque in different directions \revise{independently}.
\revise{To satisfy this condition with a small number of actuators, we determine the thruster configuration from the viewpoint of wrench-controllability.}

\revise{We assume each link has one thruster module. Thus, there are $N_q + 1$ thruster modules.}
\revise{Let $\bm{\eta}_i \in \mathbb{R}^{d_i}$ be the input vector of the $i$-th thruster module.
In general, the input for thruster includes the thrust magnitude and DoFs which control thrust direction.
The stacked input vector is as follows,
\begin{equation*}
    \bm{\eta}
    =
    \begin{bmatrix}
        \bm{\eta}_1^T, & \cdots , & \bm{\eta}_{N_q + 1}^T
    \end{bmatrix}^T
    \in \mathbb{R}^{n_\eta}, \hspace{3mm}
    n_\eta = \sum_{i=1}^{N_q+1} d_i.
\end{equation*}
}
\revise{The net wrench on CoG $\bm{w}\in\mathbb{R}^6$ can be written as $\bm{w} = \bm{w}(\bm{q},\bm{\eta})$.
Linearizing around a configuration $(\bm{q},\bm{\eta})$ yields}
\begin{equation*}
    \revise{\delta \bm{w}
    = G(\bm{q},\bm{\eta})\,\delta \bm{\eta},\quad
    G \in \mathbb{R}^{6\times n_\eta},}
\end{equation*}
\revise{where $G$ is the local wrench-input Jacobian.}
\revise{Since $\mathrm{rank}\,G \le n_\eta$, a necessary condition for locally controlling all six components of the body wrench is}
\begin{equation*}
    \revise{n_\eta = \sum_{i=1}^{N_q+1} d_i \;\ge\; 6.}
\end{equation*}

Hence, we consider introducing the thrust vectoring mechanism that make\revise{s} it possible to change the thrust direction.
By using this mechanism, it is possible to increase the number of control inputs without increasing the number of \revise{thrusters}.
There are two \revise{examples} for this mechanism, with 1 DoF \cite{kamel2018voliro} and with 2 DoF \cite{zhao2023spidar}.
In the 1 DoF design, the thruster can be rotated around a certain axis, and in the 2 DoF configuration, the thruster can be rotated around two orthogonal axes.
With the 2 DoF design, it is possible to exert thrust in any direction, but it is necessary to use a small diameter propeller to avoid \revise{collision} with other parts of the body.
\revise{
With 1 DoF thrust vectoring modules, each thruster has two inputs ($d_i = 2$): the thrust magnitude and a single vectoring angle around a fixed axis.
In this case, $n_{\eta} = 2(N_q+1)$, and the condition $n_{\eta} \ge 6$ is satisfied for the first time at $N_q = 2$.}
In this work, it is sufficient to align the directions of thrusts only during flight, where a large thrust in the anti-gravity direction is required, so we adopt the 1 DoF \revise{design}.
Furthermore, in order to enable the thruster to control the exerted force not only in the direction that can be affected by thrust vectoring, but also in the direction along with the vectoring axis, the propeller is tilted as shown in \figref{figure:robot_configuration}(c2).\\
}{
飛行時には\figref{figure:robot_configuration}(a)に示すように, 機体の重心がプロペラの支持領域内に入るようにするため、リンクが水平面上に広がるような構成になる。
そして飛行時には推力方向を揃えて抗重力方向に大きな推力を発揮して機体を浮遊させる必要がある.
一方で, 地上動作時には\figref{figure:robot_configuration}(b)に示すように各リンクに搭載したフレームが接地するように動作する
バランス制御や転がり動作のため各方向に推力を発揮可能にする必要がある.
つまり, 機体姿勢の異なる複数の形態で, 飛行のための大出力と, 地上でのバランス制御や転がりのためのモーメントを発揮できる必要がある.
そのためには, 制御自由度を増やす必要があるが, 最小構成の骨格デザインではロータ数を多くすることが難しい.

そのため，本研究ではロータ推力の発揮方向を変化させる自由度である，推力偏向自由度を導入する.
これにより, ロータ数を増やすことなく, 制御自由度を増加させることができる.
推力偏向機構には1自由度の構成\cite{HydrusSingularityFreeFlight:ChouBakui:RAL}と2自由度の構成\cite{zhao2023spidar}が考えられる.
1自由度の構成では, ある軸周りに推力装置を回転可能な構成で, 2自由度のものは, 直交する2軸回りに推力装置を回転可能な構成である.
2自由度の構成では任意の方向に推力を発揮可能であるという一方で, マルチリンク型機体への導入は機体の他の部分との干渉を避けるため小さい径のプロペラを使う必要がある.
また, 本研究では, 抗重力方向に大きな推力を発揮する必要のある飛行時にのみ推力の発揮方向を揃えられれば良いかつ, 最小構成にするため, \figref{figure:robot_configuration}に示すような1自由度の構成を採用する.
さらに, 推力偏向機能によって干渉可能な軸周りだけでなく, 回転軸にそった方向への発揮力も推力によって制御可能にするため, \figref{figure:robot_configuration}に示すように, プロペラの回転面はチルトさせておく.\\
}

\switchlanguage{
Considering \secref{section:link_number} and \secref{section:thruster_configuration}, the robot proposed in this work consists of three links with one \revise{thruster} on each link, and each \revise{thruster has} 1 DoF thrust vectoring mechanism around the link as shown in \figref{figure:robot_configuration}(c).
}{
\subsecref{section:link_number}, \subsecref{section:thruster_configuration}を踏まえ，本研究で構成する機体は各リンクに1つのロータを有する3リンクの機体とし，各ロータをリンク周りに回転させる推力偏向自由度を導入する。
これによって制御自由度数が6となるので力とトルクを独立に制御することが可能になりうる.
}

\subsection{Modeling}
\label{section:modeling}
\switchlanguage{
In this part, we describe the kinematic model of the robot.
\revise{Throughout this paper, the matrix \(^{\{X\}}R_{\{Y\}}\) denotes the rotation matrix that converts a vector in \(Y\) to a vector in \(X\).
Besides, \(*^{des}\) represents the desired value of \(*\).
Moreover, when \(\bm{a}\) is a three-dimensional vector, \(a_x\), \(a_y\) and \(a_z\) represent the first, second, and third components of \(\bm{a}\), respectively.
}
}{
各物理量を\tabref{table:definition_of_physical_quantity}に示す.
}

\begin{figure}
    \centering
    \includegraphics[width=1.0\columnwidth]{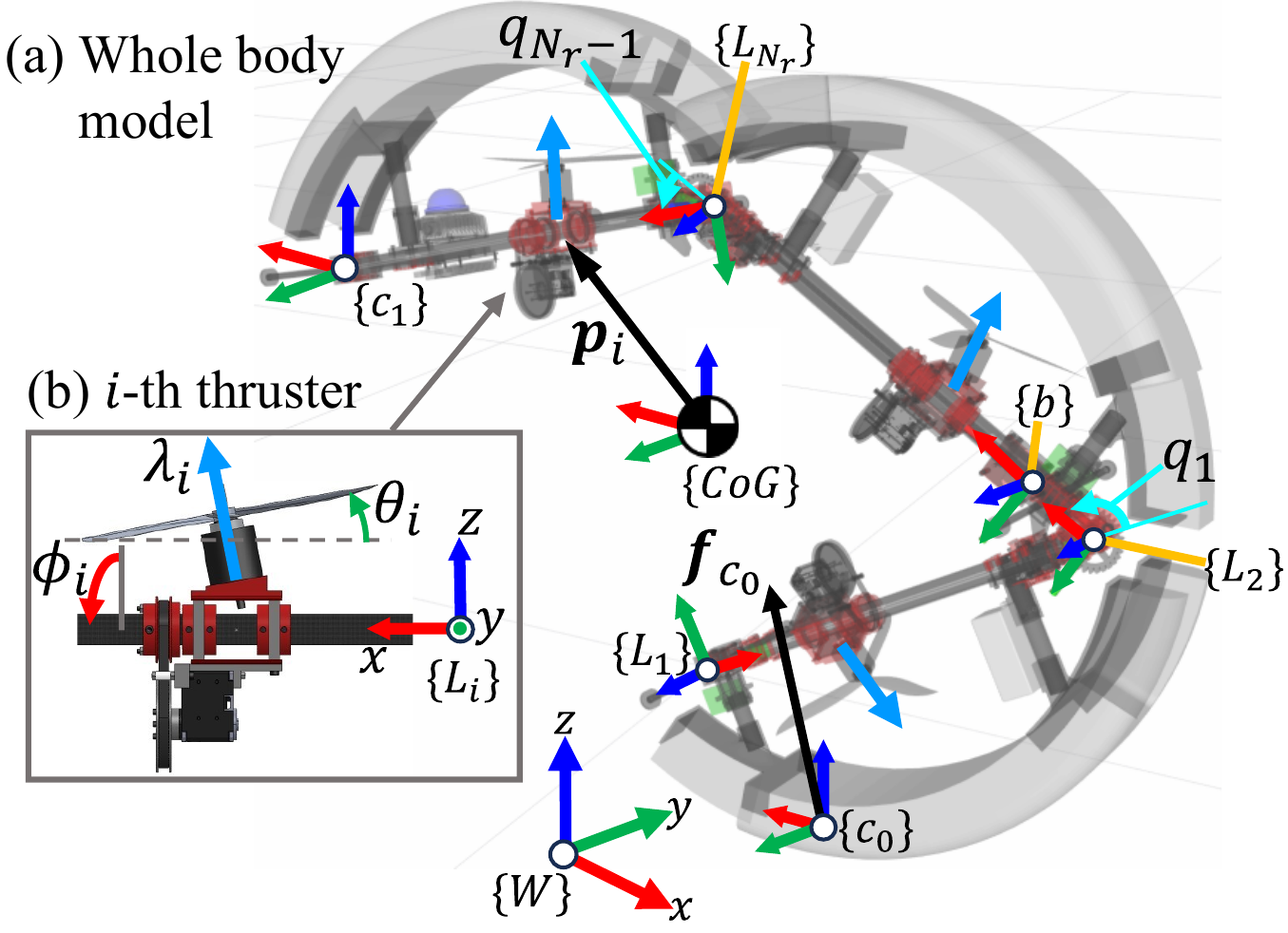}
    \caption{\revise{(a): Kinematics model of whole body. (b): \(i\)-th thruster model. The \(x\)-, \(y\)-, and \(z\)-axes of each frame are shown in red, green, and blue, respectively.}}
    \label{figure:robot model}
    \vspace{-5mm}
\end{figure}

\subsubsection{Definition of Coordinate Systems}
\switchlanguage{
\revise{In this work, we assume that the robot works on a rigid, level ground surface during ground mode.}
We define the coordinate systems as \revise{follows and } shown in \figref{figure:robot model}\revise{(a)}.
\begin{itemize}
  \item \(\{W\}\): World frame. Fixed on the ground\revise{.} \revise{Its} \(z\)\revise{-}axis point\revise{s} vertically upward.
  \item \(\{b\}\): Baselink frame. It\revise{s pose} is corresponding to \revise{that of} IMU \revise{on} the flight controller.
\revise{Desired attitude of this frame w.r.t. \(\{W\}\) is defined as follows,
\begin{equation}
 ^{\{W\}}R_{\{b\}}^{des} = R_{z}(\psi_{b}^{des})R_{y}(\theta_{b}^{des})R_x(\phi_{b}^{des}).
 \label{eq:baselink_rot}
\end{equation}
In \eqref{eq:baselink_rot}, \(R_{*}(\alpha)\) is a rotation matrix that rotates around the \(*\) axis by \(\alpha\).
\revise{Here, we use ZYX Euler angle. \(\phi_{b}^{des}\), \(\theta_{b}^{des}\), and \(\psi_{b}^{des}\) are desired roll, pitch, and yaw angles of baselink, respectively.}
To change the attitude of robot, these angles are commanded.
}
  \item \(\{\revise{C}o\revise{G}\}\): Center of gravity frame. The origin of the frame coincides with the CoG of the robot and is calculated from the joint angles at that time. The relative attitude with \(\{b\}\) is defined as follows,
}
{
\figref{figure:robot model}に示す座標系を以下のように定義する.
\begin{itemize}
    \item \(\{W\}\): 世界座標系. 地上に固定されており\(z\)軸が鉛直上向きを向いている．
    \item \(\{b\}\): ベースリンク座標系. フライトコントローラのIMUの座標系を表す.
    \item \(\{cog\}\): 重心座標系. 座標系の原点はそのときの関節角度から計算される機体重心と一致しており, \(\{b\}\)との相対姿勢が以下のように定義される. 
}
\begin{equation}
 {}^{\{\revise{C}o\revise{G}\}}R_{\{b\}} = R_y(\theta_{b}^{des})R_x(\phi_{b}^{des}).
  \label{eq:cog_rot}
\end{equation}
\switchlanguage{
\revise{From \eqref{eq:baselink_rot} and \eqref{eq:cog_rot}, the attitude control is performed to keep the \(xy\) plane of \(\{CoG\}\) horizontal because}
\begin{equation}
 {}^{\{W\}}R_{\{\revise{C}o\revise{G}\}}^{\revise{des}}={}^{\{W\}}R_{\{b\}}^{\revise{des}}{}^{\{\revise{C}oG\}}R_{\{b\}}^{-1}=R_{z}(\psi_{b}^{des}).\nonumber
\end{equation}
 \item \revise{\(\{c_i\}\): \(i\)-th contact point frame.}
       \revise{Its origin is the target position of the \(i\)-th contact point.
       Its attitude is defined by how the contact occurs at that point which will be detailed at the time of its introduction.}
 \item \revise{\(\{c_0\}\)}: Ground contact point frame.
       \revise{We assume that \(0\)-th contact point is ground contact point.
       By utilizing the assumption that the robot work on a level surface, \(\{c_0\}\) corresponds to the point where the \(z\) coordinate in the \(\{CoG\}\) frame is smallest at the contactable point formed by the outer surface of the propeller protectors.
       This point is calculated geometrically using the robot state.
       The \(z\)\revise{-}axis points vertically upward, and the yaw angle is the same as that of the \(\{\revise{C}o\revise{G}\}\) frame.}
 \item \(\{L_i\}\): \(i\)-th link frame. Fixed at the origin of the \(i\)-th link\revise{.}
       \revise{Its \(x\)-axis is oriented toward the \(\qty(i+1)\)-th link frame \(\{L_{i+1}\}\), so it is aligned with long axis of the its link.}
       \revise{Its \(z\)-axis is oriented vertically relative to the plane in which the links exist.}
\end{itemize}
}{
ただし\(\phi_{b^{des}}\), \(\theta_{b}^{des}\)はそれぞれベースリンク座標系の目標roll角とpitch角, \(R_{*}(\alpha)\)は\(*\)軸周りに角度\(\alpha\)だけ回転する回転行列を表す.
ベースリンク座標系の目標姿勢をZYXオイラー角\(\qty(\phi_b^{des}, \theta_b^{des}, \psi_b^{des})\)を用いて表現するとき, このように定義することで, \(^{\{W\}}R_{\{cog\}} = ^{\{W\}}R_{\{b\}}\hspace{0mm}^{\{cog\}}R_{\{b\}}^{-1} = R_z(\psi_b)\)となる. 
そのため, 重心を制御中心としたとき, その姿勢制御は重心座標系を水平に保つよう制御することとなる.
    \item \(\{cp\}\): 接地点座標系. 地面との接触点の目標位置を原点とし, \(\{cog\}\)の状態を使って幾何学的に決定される. \(z\)軸が鉛直上向きを向き, yaw角が\(\{cog\}\)のそれと一致する.
    \item \(\{L_i\}\): \(i\)番目リンク座標系. \(i\)番目のリンクの原点に固定されており, \(x\)軸がリンクの長軸方向と一致する.
\end{itemize}
}

\subsubsection{Thrust Model}
\switchlanguage{
Rotating propeller generates thrust force along to rotational axis and drag moment around rotational axis, so the force and torque generated by thrusts on \(\{\revise{C}o\revise{G}\}\) are represented as follows,
}{
回転するプロペラは回転軸に沿った推力と軸周りの反トルクを生み出すので, 推力が\(\{cog\}\)座標系に発生させるレンチは以下のように表される.
}
\begin{equation}
 \begin{gathered}
  \begin{bmatrix}
   \bm{f}_{\lambda}\\
   \bm{\tau}_{\lambda}
  \end{bmatrix}
  =
  Q(\bm{q}, \bm{\phi})\bm{\lambda},\\
  Q(\bm{q}, \bm{\phi}) =
  \begin{bmatrix}
   \bm{u}_1, & \cdots, & \bm{u}_{N_{r}}\\
   \bm{v}_1, & \cdots, & \bm{v}_{N_{r}}
  \end{bmatrix},\\
  \bm{\lambda} =
  \begin{bmatrix}
   \lambda_1 & \cdots \ &  \lambda_{N_r}
  \end{bmatrix}^T,\\
  \bm{v}_i = \bm{p}_{i} \cross \bm{u}_i + \sigma_i \bm{u}_i.
 \end{gathered}
 \label{eq:Q_definition}
\end{equation}
\switchlanguage{
\revise{In \eqref{eq:Q_definition}, \(\bm{f}_{\lambda}\) and \(\bm{\tau}_{\lambda}\) are force and torque generated by thrust on \(\{CoG\}\).}
\revise{\(N_r\) is the number of thrusters.}
\(\bm{u}_i\) \revise{and} \(\bm{p}_i\) \revise{are rotational axis and position of \(i\)-th thruster expressed in \(\{CoG\}\) frame, respectively.}
\revise{They} are calculated \revise{by} forward kinematics using the current joint angles \revise{\(\bm{q}\)} and thrust vectoring angles \revise{\(\bm{\phi}\)}.
\revise{\(\sigma_i\) is ratio of drag moment to thrust of \(i\)-th thruster and \(N_r\) is the number of thrusters.}
}{
\(\bm{u}_i\), \(\bm{p}_i\)は現在の関節角及び推力方向角を用いて順運動学に基づき計算される. 
}

\subsubsection{Motion of CoG and Joints}
\label{section:motion_equation}
\switchlanguage{
\revise{
In this work, we focus on tasks in which skeletal joints move slowly.
Actually, in order to perform environmental contact motions by lightweight aerial robot, joint actuators with high reduction ratios are necessary to exert high torque, leading to difficulties in increasing joint angular velocity.
Furthermore, while the thrust vectoring angles are used for real-time control, the inertia of the rotating sections are small compared to that of the entire body.
Therefore, in the equations of motion, we set \(\dot{\bm{q}}\), \(\ddot{\bm{q}}\), \(\dot{\bm{\phi}}\), and \(\ddot{\bm{\phi}}\) to \(\bm{0}\) and omit all terms related to joint velocities and accelerations.
On the other hand, for the motion of the CoG, we consider the full dynamics, including the velocity and acceleration.}
\revise{Based on these assumptions and} kinematic model depicted in \figref{figure:robot model}\revise{(a)}, translational and rotational motions of CoG, \revise{as well as the} joint \revise{dynamics}, are described as follows,
}{
本研究では関節の運動を静的と仮定し, 関節の速度と加速度をゼロとみなす.
\figref{figure:robot model}に示す運動モデルとニュートン・オイラー方程式から, 重心の並進と回転, 関節運動の運動方程式は以下のように表される.
}
\begin{gather}
 m \ddot{\bm{r}} = -m\bm{g} + ^{\{W\}}R_{\{\revise{C}o\revise{G}\}} \bm{f}_{\lambda} + \sum_{i=0}^{N_c} {}^{\{W\}}R_{\{c_{\revise{i}}\}} \bm{f}_{c_i},\label{eq:cog_latelal}\\
 I\dot{\bm{\omega}} = -\bm{\omega} \times I \bm{\omega} + \bm{\tau}_{\lambda} + \sum_{i=0}^{N_c} {}^{\{\revise{C}o\revise{G}\}}\bm{p}_{\{c_{\revise{i}}\}} \times \qty(^{\{\revise{C}o\revise{G}\}}R_{\{c_{\revise{i}}\}}\bm{f}_{c_{\revise{i}}}),\label{eq:cog_angular}
\end{gather}
\begin{align}
 \begin{aligned}
  \sum_{i=1}^{N_{seg}} J^T_{seg_i} m_i
  \begin{bmatrix}
   \bm{g}\\
   \bm{0}
  \end{bmatrix}
  = \bm{\tau}_{joint} + &\sum_{i=1}^{N_{r}} J^T_{\lambda_i}
  \begin{bmatrix}
   \bm{u}_i\\
   \sigma_i \bm{u}_i
  \end{bmatrix}
  \lambda_i
  \revise{+ \sum_{i=0}^{N_c} J^T_{c_{i}}
  \begin{bmatrix}
   \bm{f}_{c_{i}}\\
   \bm{0}
  \end{bmatrix}}.
  \label{chap4::eq::joint_motion}
 \end{aligned}
\end{align}
\switchlanguage{
In \eqref{eq:cog_latelal}--\eqref{chap4::eq::joint_motion}, \revise{\(\bm{r}\), \(m\), \(\bm{g}\), \(I\), \(\bm{\omega}\)}, \revise{are position of CoG in world frame, mass of robot, gravitational vector, inertia matrix expressed in CoG frame, angular velocity of CoG.}
 \(\bm{f}_{c_{\revise{i}}}\) \revise{is contact force at \(\{c_{\revise{i}}\}\) and it is expressed in \(\{c_i\}\) frame.}
\(^{\{\revise{C}o\revise{G}\}}\bm{p}_{\{c_{\revise{i}}\}}\) is position of \(\{c_{\revise{i}}\}\) frame expressed in \(\{\revise{C}o\revise{G}\}\) frame.
\(\bm{\tau}_{joint}\) \revise{and \(N_{seg}\) denote joint torques and the number of segments in the robot, respectively.}
\revise{\(N_{c}\) is and the number of contact points other than ground contact point.}
\revise{\(J_{seg_i}\), \(J_{\lambda_{i}}\) and \(J_{c_{\revise{i}}}\) are jacobian matrices of the CoG of \(i\)-th segment in robot, the \(i\)-th thruster frame, and the \revise{\(i\)-th} contact point, respectively.}
\revise{L}eft-hand side of \equref{chap4::eq::joint_motion} represents the effect of gravity.
\revise{Second term of right-hand side of \equref{chap4::eq::joint_motion} represents the wrench generated by thrusters, and the third term represents the effect of the contact forces.}
}{
\(\bm{\tau}_{joint}\), \(\bm{f}_{cp}\)は関節トルク, \(\{cp\}\)における接触力を表す．
\(^{\{cog\}}\bm{p}_{\{cp\}}\)は\(\{cog\}\)からみた\(\{cp\}\)の位置ベクトルを表す.
また, \equref{chap4::eq::joint_motion}の左辺第二項は推力装置が生み出すレンチによる影響, 第三項は重力による影響, 第四項は接触力による影響を表す．
}

\subsubsection{Minimum Thrust Model}
\label{section:thrust_minimization_model}
\switchlanguage{
In this subsection\revise{,} we \revise{formulate} the thrust minimization model in the nominal state of aerial mode and terrestrial mode based on \revise{the model described in} \secref{section:motion_equation}.
This model define\revise{s} the joint design \revise{requirement} and \revise{form} the core of the optimization problem for real-time control in \secref{section:control}.
}{
ここでは, \secref{section:motion_equation}に基づく, ノミナルな状態における推力最小化モデルについてのべる.
この推力最小化モデルは\secref{section:design_joint}における関節の設計要件の定義に使われたり, \secref{section:control}の実時間制御においてとく最適化問題のコアになる.
}

\switchlanguage{
The nominal state during flight is the state in which the \revise{wrench} equivalent to gravity is applied to the CoG by \revise{all of external forces including thrust}.
We can find the \revise{optimal thrusts, thrust vectoring angles, and external forces} that minimize the thrust while satisfying this constraint by solving the following optimization problem,
}{
飛行時のノミナルな状態は, 与えられたコンフィギュレーションにおいて, 推力により自重に相当する力を重心に作用させている状態である.
この制約をみたしつつ発揮推力が最小になるアクチュエータ入力は, 以下の最適化問題を解くことで求められる.
}
\begin{align}
 &\underset{\bm{\lambda}, \bm{\phi}, \bm{\tau}_{joint}, \revise{\bm{f}_{c_{0}} \cdots \bm{f}_{c_{N_{c}}}}}{\text{minimize}} && \norm{\bm{\lambda}}^2,\label{chap4::eq::aerial_mode_cost_function}\\
 & \text{subject to} && \eqref{chap4::eq::joint_motion},\nonumber\\
 &&& \bm{f}_{c_{0}} = \bm{0},\label{eq:aerial_f_c0_constraint}\\
 &&&
 \begin{aligned}
  \bm{w}^{des} =
  \begin{bmatrix}
   m\bm{g}\\
   \bm{0}
  \end{bmatrix}
  = Q(\bm{q}, \bm{\phi}) \bm{\lambda}\\
  & \hspace{-24mm} \revise{+ \sum_{i=0}^{N_c} {}^{\{CoG\}}R_{\{c_i\}} \bm{f}_{c_i}.}
 \end{aligned} \label{chap4::eq::aerial_mode_wrench_constraint}
\end{align}
\switchlanguage{
\revise{In aerial mode, we assume no contact force acts at \(\{c_0\}\) and derive the constraint in \eqref{eq:aerial_f_c0_constraint}.
 \(\bm{w}^{des}\) is desired wrench on CoG to cancel the gravity.}
}{
本研究で提案する機体構成の制御入力数6は空間の自由度数と等しいため, 位置と姿勢に関する等式制約条件のみを課している. 
}

\switchlanguage{
The nominal state during ground motion is the state in which the rotational moment caused by the contact force\revise{s} and gravity is canceled by thrust.
In addition, the \revise{ground} contact force \revise{must remain} inside the friction cone in order to keep the ground contact.
Furthermore, \revise{when the joint angle is not corresponding to its boundary,} the \revise{gravity} generate loads on the joints during ground motion, which is different from flight.
\revise{To prevent overloading the joint actuators}, \revise{we penalize} the \revise{joint} torques in the objective function and \revise{explicitly constrain them in the optimization}.
The optimization problem to obtain the actuator input is formulated as follows,
}{
地上動作時におけるノミナルな状態は, 接触力と重力に起因する回転モーメントをより打ち消している状態である.
また, 接地を維持するためには接触力が摩擦錐の内部にある必要がある.
これらの制約に加え, 地上動作時には飛行時と異なり接触力が関節への負荷を生み出すので, 関節での発揮トルクも目的関数と制約条件に追加する.
アクチュエータ入力を得るための最適化問題を以下のようにする.
}
\begin{align}
 & \underset{\bm{\lambda}, \bm{\phi}, \bm{\tau}_{joint}, \revise{\bm{f}_{c_{0}} \cdots \bm{f}_{c_{N_c}}}}{\text{minimize}} && w_1\norm{\bm{\lambda}}^2 + w_2\norm{\bm{\tau}_{joint}}^2,\label{eq:sqp_original_cost}\\
 & \text{subject to} && \eqref{chap4::eq::joint_motion}, \nonumber\\
 &&& \begin{aligned}
  \bm{\tau}_{\lambda}^{des} -\revise{\sum_{i=0}^{N_c}} {}^{\{\revise{C}o\revise{G}\}}\bm{p}_{\{c_{\revise{i}}\}} \cross \bm{f}_{c_{\revise{i}}}\\
   = Q_{rot}(\bm{q}, \bm{\phi}) \bm{\lambda},
 \end{aligned} \label{eq:sqp_tau_lambda}\\
 &&& f_{\revise{c_0}, z} > 0,\label{chap4::eq::sqp_z_constraint}\\
 &&& \qty|f_{\revise{c_0}, x}| < \mu f_{\revise{c_0}, z}, \hspace{1mm}\qty|f_{\revise{c_0}, y}| < \mu f_{\revise{c_0}, z},\label{chap4::eq::sqp_y_constraint}\\
 &&& 0 < \lambda_i < \lambda_{max},\label{chap4::eq::sqp_lambda_limit}\\
 &&& \qty|\tau_{joint, i}| < \tau_{joint, max}, \label{chap4::eq::sqp_joint_torque_limit}
\end{align}
\switchlanguage{
\revise{I}n \eqref{eq:sqp_original_cost}, \revise{the first term minimizes the total thrust, and the second term reduces the joint torque load}, where \(w_1\), \(w_2\), are the \revise{corresponding} weights.
\(Q_{rot}\qty(\bm{q}, \bm{\phi})\) \revise{in \eqref{eq:sqp_tau_lambda}} is the lower \revise{three} rows of \(Q\qty(\bm{q}, \bm{\phi})\) \revise{in \eqref{eq:Q_definition}}.
\eqref{chap4::eq::sqp_z_constraint}--\eqref{chap4::eq::sqp_y_constraint} express \revise{the linearized} friction cone constraint to prevent slipping and floating.
\(\mu\) is the static friction coefficient.
\eqref{chap4::eq::sqp_lambda_limit} and \eqref{chap4::eq::sqp_joint_torque_limit} are constraints on thrust and joint torque, respectively.
\(\lambda_{max}\) is the maximum thrust, and \(\tau_{joint, max}\) is the stall torque of joints.
}{
ここで,  \eqref{eq:sqp_original_cost}の各項はそれぞれ発揮推力の最小化, 関節の負荷トルクの軽減を目的とする項であり, \(w_1\), \(w_2\)が各項の重みを表す．
\eqref{chap4::eq::sqp_torque_constraint}における\(Q_{rot}\qty(\bm{q}, \bm{\phi})\)は\(Q\qty(\bm{q}, \bm{\phi})\)の下3行を取り出したものである.
\eqref{chap4::eq::sqp_z_constraint}--\eqref{chap4::eq::sqp_y_constraint}は四角錐に近似した摩擦錐制約で, 浮遊と滑りを防いでいる.
\(\mu\)は静止摩擦係数である.
\eqref{chap4::eq::sqp_lambda_limit}や\eqref{chap4::eq::sqp_joint_torque_limit}は発揮推力や関節負荷トルクに関する制約で, ハードウェアによって規定される上限値を超えないようにしている.
ただし\(\lambda_{max}\)は最大推力, \(\tau_{joint, max}\)は最大関節トルクを表す.
}

\begin{figure}[t]
    \centering
    \includegraphics[width=1.0\columnwidth]{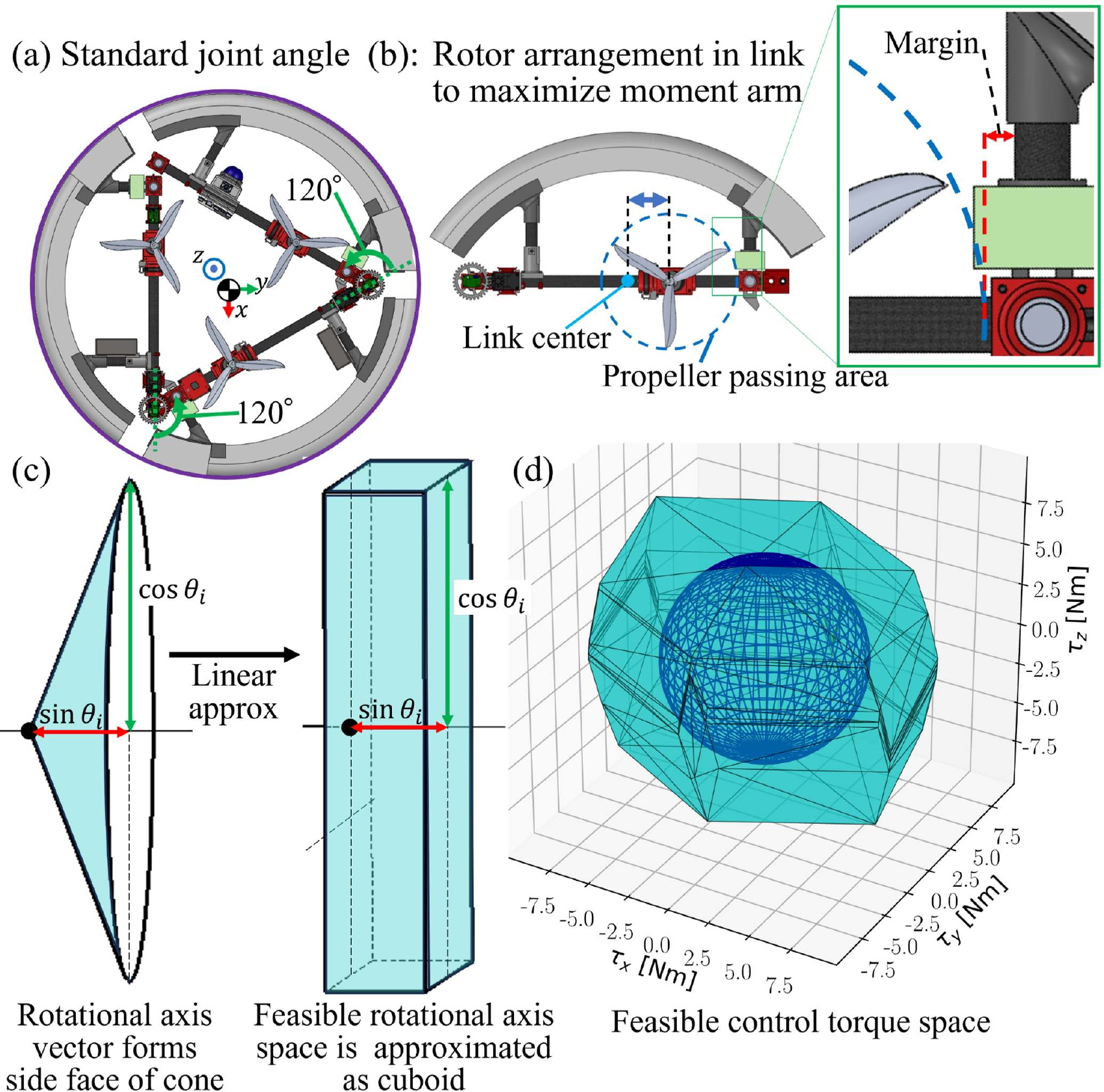}
    \caption{(a): Top view of whole robot under standard joint angle. (b): \revise{Thruster} arrangement in each link. (c): Approximation adopted in calculation of feasible torque space. (d): The lightblue convex hull is the feasible torque space with the selected tilt angles. The darkblue sphere is inscribed it and the radius is the minimum feasible torque.}
    \label{figure:fct space}
    \vspace{-5mm}
\end{figure}

\subsection{\revise{Thruster} Design}
\label{section:rotor_configuration}
\switchlanguage{
As described in \secref{section:design methodology}, the robot \revise{developed} in this work consists of three links and a thrust vectoring mechanism that can rotate \revise{thruster} around \revise{its} link.
The number of control inputs in this configuration is \revise{six}; however, \revise{in order to achieve motions that involve attitude changes, the \revise{thrusters} must be placed so that they can generate sufficient force and torque in all required directions.}
In this \revise{subsection}, we describe \revise{how the positions and orientations of the \revise{thrusters}  are determined}.

\revise{
In this work, the arrangement of the thrusters and their tilt angles is designed for a single representative configuration, which we refer to as the standard joint angle configuration.
For this robot, this configuration is defined as in which the two hinge joints are set to \(120^\circ\) and the propeller guards form a circle, as shown in \figref{figure:fct space}(a).
This robot can realize multiple configurations through joint actuation, and in principle one could evaluate control stability for several configurations and optimize over their combined performance.
However, this is not strictly required for the control objectives.
Therefore, we prioritize locomotion stability in the configuration that is most frequently used for normal flight and rolling locomotion, and so we consider the thruster design for this configuration.
}}{
\secref{section:design methodology}で述べたように，本研究では3つのリンクと, それらをそれぞれリンク周りに回転可能な推力偏向機構をもつ構成とした．
また, 各プロペラは推力偏向自由度が寄与できない1軸方向にも力を発揮できるように, プロペラの回転面がチルト角をもっている.
この構成では制御入力数が6となっているが, 姿勢の変化を伴う動作を実現するためには各方向に力とトルクを十分に発揮できるようロータを配置する必要がある.
そこで, 制御を安定化させやすいロータの位置と, プロペラ傾斜角の決定法を述べる.
この項では飛行, 転がりのロコモーションを行う, 最も基本的な関節角度である, \figref{figure:fct space}(a)に示す標準関節角度での安定性を考える.
この標準関節角度は, 全リンクのフレームで真円が形成されるときで, 本研究の機体ではすべての関節角度が\(120\)度となる.
}
\subsubsection{Arrangement of \revise{Thrusters}}
\label{section:rotor_arrangement}
\switchlanguage{
First, we consider the arrangement of \revise{thrusters} in the link.
By changing the thrust direction, the force exerted on the link can be controlled in two directions perpendicular to the \revise{\(x\)-}axis of the \revise{\(i\)-th link frame \(\{L_{i}\}\)}.
However, if the \revise{thrusters} are placed at the center of the link, the length of moment arm of the force contributing to the rotation around the \(z\)\revise{-}axis of CoG in \figref{figure:fct space}(a) is short.
Therefore, the \revise{thrusters} are placed as far away as possible from the center of the link to avoid interference with other parts of the body, as shown in \figref{figure:fct space}(b).
\revise{In the prototype}, the margin shown in \figref{figure:fct space}(b) was set \revise{to} \SI{10}{mm}.
}{
まず, リンク内におけるロータの配置を考える.
推力方向を変化させることによって, リンクの長軸に直行する2方向に発揮する力を制御できる.
ただし, ロータをリンク中心に配置した場合，標準関節角度(\figref{figure:fct space}(a))で, CoG座標の\(z\)軸周りの回転に寄与する力のモーメントアームが短くなる.
そのため，\figref{figure:modules}(b1)に示すようにロータは機体の他の部分と干渉しないようリンク中心から最大限離れた位置に配置する．
実際に, \figref{}にしめすマージンはhogehogeにした.
}

\subsubsection{\revise{Design} of Propeller Tilt Angle}
\label{section:rotor_tilt}
\switchlanguage{
In the \revise{proposed} thruster \revise{design}, the propeller rotation plane is tilted with respect to the link as shown in \figref{figure:robot_configuration}(c2).
This allows the thrust to generate force \revise{along} the \revise{link's} longitudinal axis, and enables the thrust to \revise{contribute to} attitude control around the \(z\)\revise{-}axis of CoG, shown in \figref{figure:fct space}(a).

\revise{Based on the thruster arrangement described in \secref{section:rotor_arrangement}, we select the propeller tilt angles \(\theta_i\) using a simple performance index.}
\revise{Following previous work on thruster design or transformation planning for multirotors in \cite{ODAR:Park:2018,sugihara2023trady,HydrusSingularityFreeFlight:ChouBakui:RAL,nishio2024singularity}, we consider the minimum feasible torque.}
The minimum feasible torque is \revise{defined as} the smallest value that can be exerted among the torque spaces exerted by thrust.
In this work, \revise{we use this index to determine the tilt angles of the propellers, considering the thrust vectoring angle.}
We \revise{extend the definition of} feasible torque space as follows,
}
{
我々の推力装置の設計では, \figref{figure:robot_configuration}(c2)に示すように, プロペラ回転面をリンクに対して傾斜させている.
これにより, 推力の発揮によりリンク長軸方向へも力を発生させることが可能であり, \figref{figure:modules}(a)に示すCoG座標\(z\)軸周りの姿勢制御に推力を利用することができる.

\secref{section:rotor_arrangement}で述べたロータの配置において, 最小保証トルクを最大化するようにこのプロペラの傾斜角\(\theta_i\)を最適化する.
最小保証トルクとは, 推力により発揮可能なトルク空間の中でもっとも小さいトルクである.
この方法をつかって, プロペラ固定式ドローンにおいて最適ロータ配置手法\cite{ODAR:Park:2018, sugihara2023trady}が提案されていたり, マルチリンク型マルチロータにおいて制御の特異点回避\cite{HydrusSingularityFreeFlight:ChouBakui:RAL, nishio2024singularity}に利用されていたりする.
我々は推力偏向角度を考慮して, 設計パラメータを最適化する.

推力偏向角度を考慮したトルク空間を以下のように定義する.
}
\begin{equation}
  \begin{gathered}
    \mathcal{V}_{T} \coloneqq \qty{\bm{\tau} \in \mathbb{R}^3 | \bm{\tau} = \sum_{i=1}^{N_r}\sum_{j=1}^{5} \lambda_{max}\alpha_j\bm{v}_{ij}},\\
    0 \leqq \alpha_j \leqq 1\\
    \begin{bmatrix}
        \bm{v}_{i1} & \cdots & \bm{v}_{i5}
    \end{bmatrix}
    =
    (\qty[\bm{p}_i \cross] + \sigma_i E_3) ^{\{\revise{C}o\revise{G}\}}R_{\{L_i\}}
    U\label{chap3::eq::v_definition}\\
    U  =    \begin{bmatrix}
        S & 0 & 0 & 0 & 0\\
        0 & C & -C & 0 & 0\\
        0 & 0 & 0 & C & -C\\
    \end{bmatrix},\\
    S = \sin\theta_i, \hspace{4mm} C = \cos\theta_i,
\end{gathered}
\end{equation}
\switchlanguage{
\(\begin{bmatrix}
    *\cross
\end{bmatrix}\) is the representation matrix of the cross product of three dimensional vector.
In \eqref{chap3::eq::v_definition}, the moments exerted on the CoG by unit thrust in each axis of the link frame are calculated.
\(\bm{v}_{i2}, \cdots, \bm{v}_{i5}\) are linearized as in \cite{zhao2020enhanced} by decomposing the exerted force in each direction.
In other words, the space of forces distributed on the sides of the cone is \revise{approximated} by a \revise{circumscribed cuboid}, as shown in \figref{figure:fct space}(c).
Here, we redefine \(\bm{v}_i\) \revise{as the collection of vectors} \(\bm{v}_{ij}\) \revise{indexed by the set} \(\mathcal{N}_T = \{1, 2, \cdots, 5N_r\}\).
\revise{The} feasible torque space \(\mathcal{V}_{T}\) is \revise{then} the \revise{convex hull} spanned by these \(5N_r\) vectors.
\revise{Because the conical surface is approximated by a \revise{cuboid}, this space becomes a convex polytope.}
Using the fact that \(\bm{v}_{i}\) is three dimensional vector, distance to the face of \(\mathcal{V}_{T}\) where \(\bm{v}_{i}\) and \(\bm{v}_{j}\) is spanning from the origin of \(\mathcal{V}_{T}\) along \(\bm{v}_i \times \bm{v}_j\) is calculated as follows.
}{
\(\begin{bmatrix}
    *\cross
\end{bmatrix}\)は3次元ベクトルによる外積を表す表現行列である.
\equref{chap3::eq::v_definition}では各リンク座標系の各軸に対して, 単位推力を発生させたときの, 重心に作用するモーメントを計算している.
なお、\(\bm{v}_{i2}\)から\(\bm{v}_{i5}\)は\cite{zhao2020enhanced}と同様に, \figref{figure:fct space}(c)に示すように, 発揮力を各方向に分解することで線形化している.
これにより円錐面の重ね合わせという非線形な計算の回避と計算量の削減をしている.
つまり, ここでは円錐の側面上に分布する発揮力の空間をそれに外接する直方体に近似して簡単化している. 
ここで、長さ\(5N_r\)の集合\(\mathcal{N}_T = \{1, 2, \cdots ,5N_r\}\)の要素と\(\bm{v}_{ij}\)を一対一対応させる.
feasibleトルク空間\(\mathcal{V}_{T}\)はこれら\(5N_r\)個のベクトルが張る空間であり, 凸包になる.
\(\bm{v}_{i}\)が3次元ベクトルであることを利用して, \(\bm{v}_i \times \bm{v}_j\)に沿った, \(\bm{v}_i\)と\(\bm{v}_j\)が張る\(\mathcal{V}_{T}\)の面と\(\mathcal{V}_{T}\)の原点までの距離は以下のようになる.
}
\begin{equation}
    d_{ij}^{\tau} = \sum_{k=1}^{N_T} \max\qty(0, \dfrac{\qty(\bm{v}_i \cross \bm{v}_j)^T}{\qty|\bm{v}_i \cross \bm{v}_j|}\bm{v}_k), \hspace{2mm} i, j \in \mathcal{N}_{T},
    \label{eq::fct_distance}
\end{equation}
\switchlanguage{
We used different definition from \cite{ODAR:Park:2018} which deploy the same \revise{two thrusters} at the same place in opposite direction.
Also, in \equref{chap3::eq::v_definition}, we assumed that the force can be exerted in both directions of the \(y\) and \(z\) axes, however, since the inner product of these two vectors and \(\bm{v}_i \times \bm{v}_j\) is opposite sign, by taking the maximum value with \(0\) in \equref{eq::fct_distance}, only one of the directions is evaluated to exert force.
Then, the minimum feasible torque is defined as follows.
}{
同じ場所に回転軸ベクトルの符号が逆のプロペラを搭載する構成としている\cite{ODAR:Park:2018}とは定義が異なることに注意する.
また, \equref{chap3::eq::v_definition}で\(y\)方向と\(z\)方向に関して, 正負の両方向に力を発揮可能と仮定したが, これら2つのベクトルと\(\bm{v}_i \times \bm{v}_j\)との内積は逆符号となり, \equref{eq::fct_distance}で\(0\)との最大値を取ることによりどちらか一方のみが\(d_{ij}^{\tau}\)に影響するため, 正負方向同時に発揮可能であると評価されはしないことに注意する.
そして, 最小保証トルクは以下のように定義される.
}
\begin{equation}
    \tau_{\min} = \lambda_{\max}\min d^{\tau}_{ij}, \hspace{4mm} i,j \in \mathcal{N}_T. \label{eq:fct_tau_min}
\end{equation}

\switchlanguage{
We aim to maximize \(\tau_{\min}\).
On the other hand, the larger the tilt angle of the \revise{thruster}, the smaller the force that can be exerted in the vertical\revise{ly} upward direction during flight, resulting in lower the energy efficiency.
Hence, we also consider the magnitude of \(\theta_i\) not to be too large.
The optimization problem is formulated as follows.
}{
我々は\(\tau_{\min}\)を最大化することを目的とする.
一方で，ロータの傾斜角が大きいほど飛行時に鉛直上方向に発揮可能な力成分が小さくなり飛行時のエネルギ効率が低下するため, \(\theta_i\)の大きさが大きくなりすぎないようにする.
最適化問題として定式化すると以下のようになる．
}
\begin{equation*}
    \underset{\bm{\theta}}{\text{maximize}} \hspace{4mm} w_1 \dfrac{\tau_{\min}}{\lambda_{\max}} - w_2 \|\bm{\theta}\|^2.
a\end{equation*}
\switchlanguage{
where \(w_1\) and \(w_2\) are the weights of each optimization term, and in this work, we set \(\qty(w_1, w_2) = \qty(4.0, 1.0)\).
We used the ISRES algorithm \cite{ISRES} to solve this problem.
We obtained the \revise{thruster} tilt angles as \(\bm{\theta}_{opt} = \begin{bmatrix} -0.0728 & 0.179 & -0.0802 \end{bmatrix}\) rad.
Since we assigned only the second rotor rotates in the opposite direction to others, it is considered that the tilt angle of its thruster becomes larger than others, and the sign is reversed.
A lightblue convex and radius of darkblue sphere in \figref{figure:fct space}(d) show the feasible torque space and minimum feasible torque (\(\tau_{min}\) in \eqref{eq:fct_tau_min}), at the standard joint angles and optimal propeller tilt angles.
\revise{These selected tilt angles are adopted to real platform in experimental validation (\secref{section:system} and \secref{section:experiment}).}
}{
\(w_1\)と\(w_2\)は各最適化の項の重みであり, 本研究では\(w_1 = 4.0\), \(w_2 = 1.0\)と設定した.
この問題を解くためにISRESアルゴリズム\cite{ISRES}を用いた.
最適なロータチルト角
\(\bm{\theta}^{opt} =
\begin{bmatrix}
    -0.0728 & 0.179 & -0.0802
\end{bmatrix}\) radを得た.
我々は2番目のロータをほかと逆に回転するようにアサインした。
二番目のロータのみほかのロータと逆回転にしているため、z軸周りに十分な力をだそうとしてこのロータの傾きが他と比べて大きくなり、符号が逆になったと考えられる。
このときの発揮可能なトルク空間を\figref{figure:fct space}(d)に示す.
}

\subsection{\revise{Model-based Requirement} for Joint}
\label{section:design_joint}
\switchlanguage{
\revise{The links are connected by actuated joints that} hold and control the joint angles.
Since joints are loaded due to exterted thrust or environmental contact force, actuators must be able to exert enough torque.
In this work, we \revise{compute} the required \revise{joint} torque\revise{s} \revise{over a range of} joint angles and attitudes, and deploy actuators \revise{that} can \revise{generate} torque greater than the maximum \revise{required value}.
To calculate the required torque\revise{s}, we use thrust minimization model described in \secref{section:thrust_minimization_model}.

\subsubsection{\revise{Calculation Method of Required Joint Torque}}
During flight, we solve the optimization problem in \eqref{chap4::eq::aerial_mode_cost_function}--\eqref{chap4::eq::aerial_mode_wrench_constraint} for \revise{a} given configuration to get optimal thrusts and thrust vectoring angles, and joint torques.
\revise{In this case, the calculation is performed under the assumption that no contact force acts on the robot.
The required torque is defined as the largest absolute value of torques obtained from this optimization.
}

During terrestrial motion, we solve the optimization problem in \eqref{eq:sqp_original_cost}--\eqref{chap4::eq::sqp_joint_torque_limit} for \revise{a} given configuration.
\revise{In this case, contact force is considered only at \(\{c_0\}\) which is \revise{the} ground contact point.}
When solving the optimization problem, the joint torque\revise{s} \revise{are treated} as a constrained optimization variable\revise{s}, so \revise{we gradually tightened the torque bounds to determine the} maximum value of the required torque.

\subsubsection{\revise{Results of the Required Torque Calculation}}
\revise{We solved above optimization problems to compute \revise{the} control inputs and contact forces.
The configurations used for the calculations in both mode are shown in the video in the supplemental material.}
\revise{If the solver fails to converge, or if} the suboptimal solution at maximum \revise{number of} iteration\revise{s} reaches actuator \revise{limits}, we \revise{regard} optimization \revise{problem as unsolved}.

The calculated required torque during flight \revise{without contact force was \SI{5.4}{Nm}}.
\revise{Despite assuming that the joint rotation axis is oriented in the same direction as gravity, the values were relatively large.
This is because, as shown in the video, in some singular configurations, the thrust direction angle and thrust increased, leading to greater joint load.
In other most configurations, required torques were less than \SI{1}{Nm}.
}

The calculated required joint torque \revise{during ground motion only with the ground contact force was \SI{7.2}{Nm}}.
During motion on the ground, the required joint torque is larger than that of during flight due to the contract force from the ground and mass of \revise{each segment of robot}.
}{
リンク同士を結合する関節部には, 相対位置を保持・制御するためのアクチュエータを搭載する. 
推力を発揮したり, 環境と接触したりするとこの関節に負荷が生じるため, 十分なトルクが発揮できる必要がある. 
そこで, 本研究では様々な関節角度, 機体姿勢の組に対して必要なトルクを求め, それらの最大値よりも大きな関節トルクを発生させられるように機体を構成する.
必要なトルクの算出には\secref{section:thrust_minimization_model}に述べた, 推力最小化ノミナルモデルを用いる.
以下で飛行時, 地上動作時における必要なトルクの計算方法を述べる.

飛行時には, \eqref{chap4::eq::aerial_mode_cost_function}--\eqref{chap4::eq::aerial_mode_wrench_constraint}に示す最適化問題を解く.
そして計算した推力, 推力方向角を多リンク系の運動方程式\eqref{chap4::eq::joint_motion}に代入することで関節で発揮する必要のあるトルクを逆算する.

地上動作時には, \eqref{eq:sqp_original_cost}--\eqref{chap4::eq::sqp_joint_torque_limit}に示す最適化問題を解く.
地上動作時の求解には関節トルクを制約つきの最適化変数とするため, その制約を徐々に強めていくことで必要なトルクの最大値を求める.

目標のアクチュエータ入力の計算には凸最適化の手法による求解が含まれる.
なので, 最適化がそもそも解けなかったまたは, 最大iterationで得たsuboptimalな解がハードウェアによって規定される境界条件と一致している際に解けなかったと判断する.

この方法で
飛行動作時に必要な関節トルクを求めると, \SI{5.7}{Nm}となった.
地上動作時に必要な関節トルクを求めると, \SI{8.3}{Nm}となった.
地上動作時は環境からの接触力が作用するため, 飛行時と比較して必要な関節トルクが大きくなったと考えられる.
}

\section{Thrust Control of \revise{Transformable} Multilink Multirotor under Environmental Contact}
\label{section:control}
\begin{figure}[t]
 \centering
 \includegraphics[width=0.99\columnwidth]{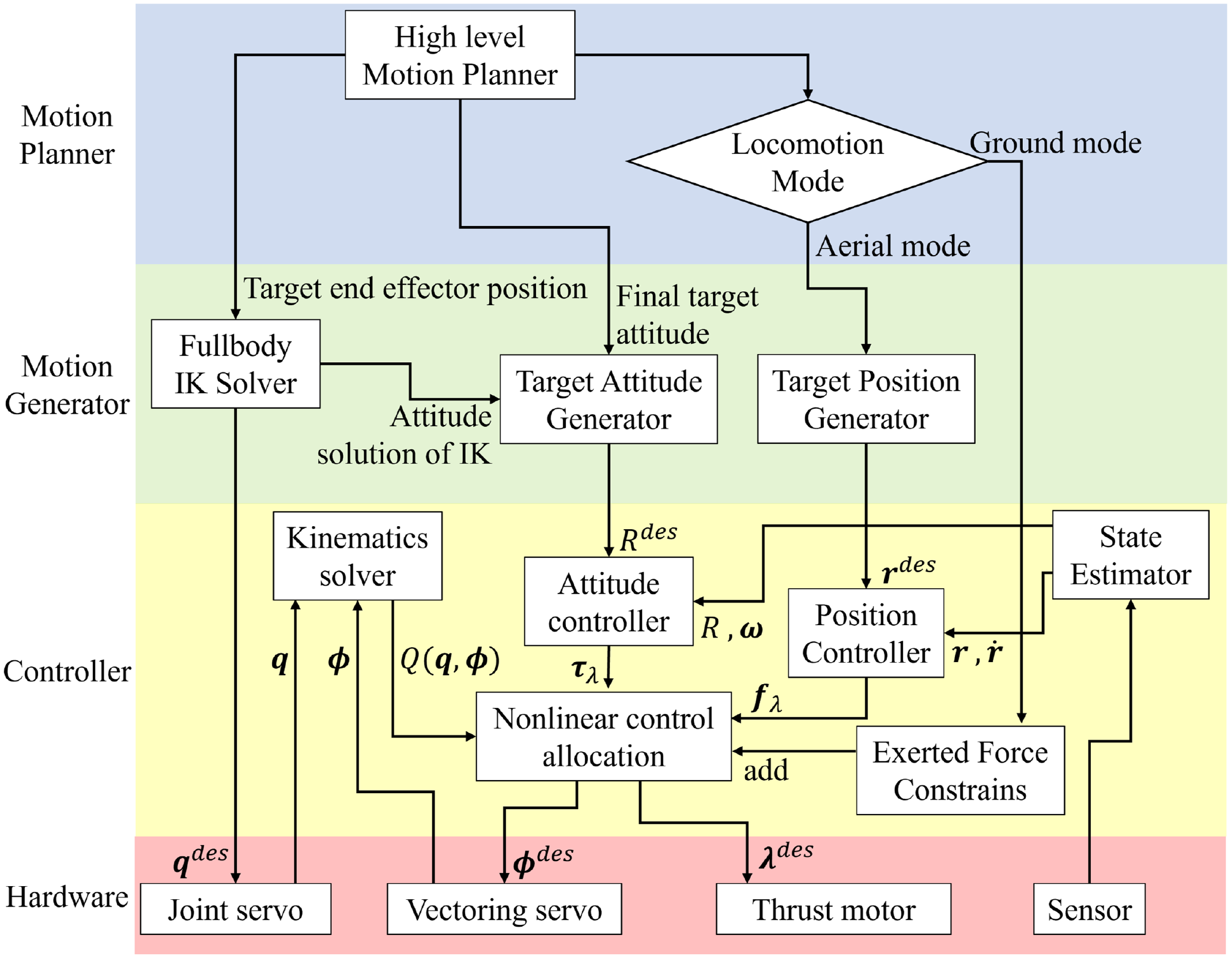}
 \caption{Motion control framework for the proposed robot platform.}
 \label{figure:control system}
 \vspace{-5mm}
\end{figure}

\switchlanguage{
The motion control framework for the proposed robot platform is shown in \figref{figure:control system}.
In this section, we describe the real-time motion control method shown in the Controller layer.
We determine the target thrusts and thrust vectoring angles using \revise{a} nonlinear optimization based on the model described in \secref{section:modeling}.
\revise{Please note that the controller for the actual experiment only considered the ground contact at \(\{c_0\}\).
However, by optimizing contact forces alongside task-specific constraints, it is possible to control the system so that the desired contact forces are explicitly reproduced.
In \secref{section:motion_strategy}, we analyze the limitations of manipulation tasks for the proposed robot based on this approach.}
}{
本研究における, motion control frameworkを\figref{figure:control system}に示す.
本節では, \figref{figure:control system}のControlレイヤに示す実時間動作制御手法を述べる.
\secref{section:modeling}で述べた, 変形と接触を考慮した運動モデルに基づき, 推力と推力方向角を非線形最適化によりリアルタイムに計算をする.
}

\subsection{Flight Control}
\switchlanguage{
During flight, we ignore the contact force \revise{from} the ground, \revise{and set}  \(\bm{f}_{c_{\revise{0}}}\) in \eqref{eq:cog_latelal}--\eqref{chap4::eq::joint_motion} to \(\bm{0}\).
\revise{The f}light control \revise{consists of} attitude control, position control, and control allocation.
}{
飛行制御では接触力を無視するので, \(\bm{f}_{cp}\) in \eqref{eq:cog_latelal}--\eqref{chap4::eq::joint_motion} は \(\bm{0}\)と見なす.
そして, 位置制御, 姿勢制御, control allocationからなる.
}

\subsubsection{Attitude Control}
\label{section:flight_attitude_control}
We \revise{apply PID control to calculate the target angular acceleration of CoG.}
\revise{From \eqref{eq:cog_angular}}, target \revise{moment generated} by thrusts on the CoG \revise{are calculated} as follows,
\begin{equation}
 \begin{split}
    \bm{\tau}^{des}_{\lambda} &= I\qty(K_{\tau, p} \bm{e}_R + K_{\tau, i} \int \bm{e}_R \dd t + K_{\tau, d} \bm{e}_{\omega})\\
    &+ \bm{\omega}\cross I \bm{\omega} \label{chap4::eq::aerial_mode_attitude_pid}
 \end{split}
\end{equation}
where \(K_{\tau, *} \in \mathbb{R} ^{3\times3}\) are PID gain diagonal matrices.
\(\bm{e}_R,\bm{e}_{\omega} \in \mathbb{R}^3\) are tracking errors of attitude and angular velocity.

\subsubsection{Position Control}
\revise{We apply PID control to compute the target translational acceleration of CoG.}
\revise{From \eqref{eq:cog_latelal}, }The target force \revise{generated} by thrusts on the CoG is calculated as follows,
\begin{equation}
 \bm{f}^{des}_{\lambda} = m {}^{\{W\}}R_{\{CoG\}}^{-1} \qty(K_{f, p} \bm{e}_{r} + K_{f, i} \int \bm{e}_{r} \dd t + K_{f, d} \dot{\bm{e}}_{r}),\label{eq:aerial_position_pid}
\end{equation}
where \(K_{f, *} \in \mathbb{R}^{3\times3}\) are PID gain diagonal matrices.
\revise{\(\bm{e}_{r}\in \mathbb{R}^3\) is the tracking error of CoG position in world frame.}

\subsubsection{Formulation of Flight Control Allocation}
\switchlanguage{
We determine the actuator inputs that realize the target \revise{f}orce and torque obtained from \eqref{chap4::eq::aerial_mode_attitude_pid}--\eqref{eq:aerial_position_pid}.
\revise{The} optimization problem to \revise{obtain the} solution is formulated by substituting \(\bm{w}_{des}\) in \eqref{chap4::eq::aerial_mode_wrench_constraint} \revise{with}
\revise{\(
\begin{bmatrix}
    (\bm{f}^{des}_{\lambda})^T & (\bm{\tau}^{des}_{\lambda})^T
\end{bmatrix}^T
\)}.
\revise{
The values used for the whole-body inertia in \eqref{chap4::eq::aerial_mode_attitude_pid} and thruster position w.r.t. \(\{CoG\}\) frame in \eqref{chap4::eq::aerial_mode_wrench_constraint} are calculated from the actual configuration at the time of each control cycle.
Although these values are changed due to the variations in the thrust vectoring angles, which are optimization variables, these changes are small relative to their original values.
Therefore, during optimization, their influence is ignored, and the control inputs are calculated assuming the initial configuration at each control step.
}
}{
位置制御器, 姿勢制御器で求めた目標加速度, 目標角加速度を実現するアクチュエータ入力を求める.
ここでは, \eqref{chap4::eq::aerial_mode_cost_function}--\eqref{chap4::eq::aerial_mode_wrench_constraint}に示す最適化問題において, 
\eqref{chap4::eq::aerial_mode_wrench_constraint}の\(\bm{w}_{des}\)を
\(
\begin{bmatrix}
    \bm{f}^{des}_{\lambda}\\
    \bm{\tau}^{des}_{\lambda}
\end{bmatrix}
\)で置き換えて解くことで解を得る.
thrusterの位置や慣性行列は各制御周期の時点の実際のコンフィギュレーションから求まる値をもちいる.
最適化変数である推力方向角の変化によって重心位置や慣性行列が変化するが, もとの重心とthrusterの距離やinertiaに対してその変化は小さいため, 最適化中にはその影響を無視して制御入力を計算する.
}

\subsection{Ground Control}
\label{section:ground control}
\subsubsection{Attitude Control in Ground Motion}
\switchlanguage{
During terrestrial mode, it is necessary to \revise{explicitly} consider the contact force with the ground.
In addition to the PID control as in flight control \revise{(}\secref{section:flight_attitude_control}\revise{)}, \revise{we also consider a} feed-forward angular acceleration and moment generated by the contact force around CoG.
The target \revise{moment generated by thrusts} is calculated as follows,
}
{
地上における動作制御では地面との接触力\(\bm{f}_{cp}\)を考慮する必要がある．
姿勢制御では, 飛行制御で行ったようなPID制御に加え, 接触力による重心周りの回転モーメントを考慮し, 目標の発揮トルクを以下のように計算する.
}
\begin{equation}
 \begin{aligned}
  \bm{\tau}_{\lambda}^{des} = I \qty(K_{\tau, p} \bm{e}_R + K_{\tau, i} \int \bm{e}_R \dd t + K_{\tau, d} \bm{e}_{\omega} + \dot{\bm{\omega}}_{ff})\\
  + \bm{\omega} \times I \bm{\omega} - ^{\{\revise{C}o\revise{G}\}}\bm{p}_{\{c_{\revise{0}}\}} \times \qty(^{\{\revise{C}o\revise{G}\}}R_{\{c_{\revise{0}}\}}\bm{f}_{c_{\revise{0}}}),\label{eq:ground_mode_attitude_pid}
 \end{aligned}
\end{equation}
where \(\dot{\bm{\omega}}_{ff}\) is the feed-forward angular acceleration.

\subsubsection{Formulation of Control \revise{A}llocation \revise{C}onsidering \revise{C}ontact \revise{F}orce}
\switchlanguage{
As mentioned in \secref{section:thrust_minimization_model}, during terrestrial motion, \revise{the} contact force \revise{must lie within} friction cone to prevent slipping and floating.
Since this is \revise{expressed as a set of} inequality constraints, the feasible region is larger than that of flight control case, which imposes equality constraints on all components of \revise{the} wrench.
Taking advantage of this \revise{larger feasible region}, \revise{we additionally consider} minimization of joint torque\revise{s} in \eqref{eq:sqp_original_cost}.
Furthermore, during control, we also \revise{penalize} changes in the thrust vectoring angles, \revise{since rapid variations can introduce} vibration\revise{s} \revise{and degrade the attitude control performance}.
Therefore, the optimization problem to calculated target actuator inputs is formulated as follows,
}{
\secref{section:thrust_minimization_model}で述べたように, 地上動作時は接触力が摩擦錐の内部になるように制御する.
これは不等式制約であり, 6次元のレンチの全成分に等式制約をかける飛行制御と比較して解空間が広がる.
これを利用して\eqref{eq:sqp_original_cost}では関節トルクの最小化も評価関数で行った.
制御中には, これに加え, 振動が姿勢制御の安定に影響する推力偏向角度の変化を抑制することも同時に考える.
よって, アクチュエータの目標入力を求める問題を以下の最適化問題として定式化する.
}
\begin{align}
  & \underset{\bm{\lambda}, \bm{\phi}, \bm{\tau}_{joint}, \bm{f}_{c_{\revise{0}}}}{\text{minimize}} &&w_1\norm{\bm{\lambda}}^2 + w_2\norm{\bm{\tau}_{joint}}^2 + w_3\norm{\bm{\phi} - \bm{\phi}_{curr}}^2,\label{eq:sqp_control_cost_function}\\
  & \text{subject to}&& \eqref{chap4::eq::joint_motion}, \eqref{chap4::eq::sqp_z_constraint}-\eqref{chap4::eq::sqp_joint_torque_limit}, \eqref{eq:ground_mode_attitude_pid},\nonumber\\
  &&& \bm{\tau}_{\lambda}^{des} = Q_{rot}(\bm{q}, \bm{\phi}) \bm{\lambda},\label{eq:sqp_control_torque_constraint}
\end{align}
\revise{where \(\bm{\phi}_{curr}\) and \(w_3\) are actual thrust vectoring angle vector, and its weighting coefficient, respectively.}

\subsubsection{\revise{Additional} \revise{C}onstraint to \revise{T}hrust \revise{V}ectoring \revise{A}ngle\revise{s}}
\label{section:additional_gimbal_constraint}
\switchlanguage{
\revise{With} the control \revise{law} based on \revise{\eqref{eq:sqp_control_cost_function}}--\revise{\eqref{eq:sqp_control_torque_constraint}}, the exerted force \revise{satisfies} the friction cone constraint\revise{s}.
\revise{However, o}nly with th\revise{ese} constraint\revise{s}, the target direction\revise{s} of \revise{the} thrusts may \revise{become nearly aligned when generating} attitude control \revise{moments}.
In the worst case, the thrust vectoring angle can \revise{oscillate} back and forth by nearly 180 degrees.
Since \revise{the actuators for thrust vectoring} have limit\revise{ed angular} velocity, if the \revise{required} change is too large, the target state cannot be achieved.
Moreover, the cable length connected to the actuator\revise{s} \revise{is} limit\revise{ed}, so it is difficult to rotate \revise{a} thruster \revise{continuously} in one direction.
Therefore, we add constraint\revise{s} that all propellers \revise{must} face outward when the robot is near\revise{ly} vertically standing state.
\revise{In this way, the rotation around} the frame surface \revise{normal is controlled} mainly by thrust, and attitude around other axes mainly by thrust vectoring.
Thus, the following constraints are added in nearly vertically standing state,
}{
\eqref{chap4::eq::sqp_cost_function}-\eqref{eq:sqp_friction_cone_y}に基づく制御では, 発揮力は摩擦錐の制約を満たす範囲になる.
この制約では, 姿勢の制御のために, 目標の推力方向が同じような方向を向くことが起こりうる.
最悪の場合推力方向角度が180度近く行ったり来たりする.
推力方向角を制御するアクチュエータの速度には限界があるため，変化量が大きすぎると目標状態を実現できなくなる．
また，アクチュエータに繋がれたケーブル長の制約もあるため，推力偏向機構を一方向に回転させ続けることは難しい．
そこで，機体が垂直近傍の状態ではすべてのプロペラがリンクの外側を向くようにする制約を加える．
これによって目標推力方向角の過剰な変化や，推力偏向機構が一方向に回り続けてケーブル長などのハードウェア的制約を満たせなくなることを防ぐ．
そのため, フレーム表面に沿った方向の回転は主に推力により，その他の軸周りの姿勢は推力偏向角を主に利用して制御する.
よって, 垂直に立っている付近の姿勢では以下の制約条件を追加する.
}
\begin{equation}
  0 < \phi_i < \pi, \hspace{2mm} \qty(i = 1, \cdots, N_r). \label{chap4::eq::rolling_gimbal_range_constraint}
\end{equation}
\switchlanguage{
Please \revise{note} the definitions of the coordinate system and the thrust vectoring angle \revise{shown in \figref{figure:robot model}(b)}.
}{
  なお, 座標系と推力方向角の定義に注意されたい.
}

\subsection{Optimization Approach for Nonlinear Control Allocation}
\label{section:nonlinear wrench allocation}
\switchlanguage{
\revise{In multirotors,} the control allocation is generally \revise{formulated as the} problem of converting the target wrench into the \revise{each thrust commands} based on the robot model.
In this work, the problem is further complicated by \revise{explicitly} considering the thrust vectoring angles.

The \revise{optimization} problem\revise{s} to be solved in this work \revise{are given by} \eqref{chap4::eq::aerial_mode_cost_function}--\eqref{chap4::eq::aerial_mode_wrench_constraint} and \revise{\eqref{eq:sqp_control_cost_function}}--\revise{\eqref{chap4::eq::rolling_gimbal_range_constraint}}.
\revise{In} some previous works on multilink multirotor\revise{s}, the \revise{nonlinear thrust vectoring} model is linearly approximated \revise{by exploiting} \revise{redundant configuration}  \cite{zhao2023spidar,nishio2023perchingarm,zhao2020enhanced}.
However, the matrix \(Q(\bm{q}, \bm{\phi})\) that converts the thrust vector \(\bm{\lambda}\) into the exerted wrench in the body frame is a nonlinear transformation of the control input variable \(\bm{\phi}\), so it is generally difficult to solve analytically.
Therefore, we solve it in real-time using a nonlinear optimization method.
Since the objective function and constraints are both twice differentiable, we use the Sequential Quadratic Programming (SQP) method.
\revise{To solve the problem, we provide gradients of cost function and constraints.
Furthermore, in the actual experiments, the robot works on a horizontal surface, and we assumed that the only contact point is the ground at \(\{c_0\}\).
Under this assumption, the contact force is treated as a function of the actuator inputs, which reduces the number of optimization variables.
The influence of this reduction is discussed in Appendix-A.
}
}{
マルチロータにおける推力配分は一般に, 力学モデルに基づいて目標とするレンチを各ロータの推力に変換する問題である.
本研究ではさらに推力方向角も考慮する問題となる. 

本研究で解く問題は\eqref{chap4::eq::aerial_mode_cost_function}-\eqref{chap4::eq::aerial_mode_wrench_constraint}および\eqref{chap4::eq::sqp_cost_function}-\eqref{eq:sqp_friction_cone_y}である.
自由度の多いマルチロータにおいて, 運動モデルを線形化している例\cite{zhao2020enhanced,zhao2023spidar,nishio2023perchingarm}があるが, 
推力ベクトル\(\bm{\lambda}\)を機体座標系における発揮レンチに変換する行列\(Q(\bm{q}, \bm{\phi})\)は制御入力である変数\(\bm{\phi}\)の非線形な変換であるため, 通常は解析的に解くことが難しい.
そこで, 非線形最適化の手法により実時間で求解を行う.
目的関数および制約条件がともに二階微分可能なため, 逐次二次計画法(SQP)を用いることとした. 
}

\section{Motion Strategies and \revise{Analysis of} Locomotion and Manipulation}
\label{section:motion_strategy}
\switchlanguage{
In this section, we describe motion \revise{strategies} for locomotion and manipulation based on the design and control \revise{framework introduced} in \secref{section:design} and \secref{section:control}.
\revise{We explicitly consider} contact constraints specific to the \revise{proposed} air-ground hybrid multirotor, which \revise{have} not been taken into account in motion planning for conventional multirotors.
\revise{Furthermore, we analyze the limitations of manipulation tasks based on the multi-contact model introduced in \secref{section:modeling}.}
}{
本節では\secref{section:design}, \secref{section:control}で提案した機体構成, 動作制御手法に基づき, ロコモーションとマニピュレーションを行うための動作の生成法について述べる.
マルチロータによる空中での動作計画においてこれまで考えられてこなかった, air-ground hybrid multirotorに特有の接触を考慮する.
}

\subsection{Rolling \revise{M}otion \revise{G}eneration}
\label{section:rolling}
\switchlanguage{
\revise{In terrestrial mode, we set the desired roll angle of baselink (\(\phi_b^{des}\)) \revise{to} \(\pi/2\) rad.
\revise{The} desired pitch angle of baselink (\(\theta_{b}^{des}\)) is \revise{then} updated to realize rolling motion.
}
In \revise{this} motion, \revise{we control} the \revise{angular} velocity around the rolling axis.
The desired angular velocity of pitch angle of \revise{the} baselink pitch \(\omega_y^{des}\) is \revise{provided} \revise{by} the motion planner, and the target pitch angle of baselink is increased in each control loop as follows,
}{
ここで, 転がり動作時には\(\{cog\}\)の\(y\)軸が転がり回転軸で, \(x\)軸が進行方向を向いているとする.
転がり移動時には転がる回転軸周りについて速度制御を行う.
動作計画機において目標の角速度\(\omega_y^{des}\)を与え, 各制御周期において, 目標角度については以下のように制御周期分だけ増加させる.
}
\begin{equation*}
  \theta_b^{des}(t + dt) = \theta_b^{des}(t) + \omega_y^{des} \dd t\revise{.}
\end{equation*}

\subsection{Stand\revise{-}up \revise{M}otion to \revise{S}witch \revise{to Terrestrial} \revise{M}ode}
\label{subsec::standing_up}
\begin{figure}
  \centering
  \begin{tabular}{cc}
      \begin{minipage}[t]{0.45\columnwidth}
          \centering
          \includegraphics[width=1.0\columnwidth]{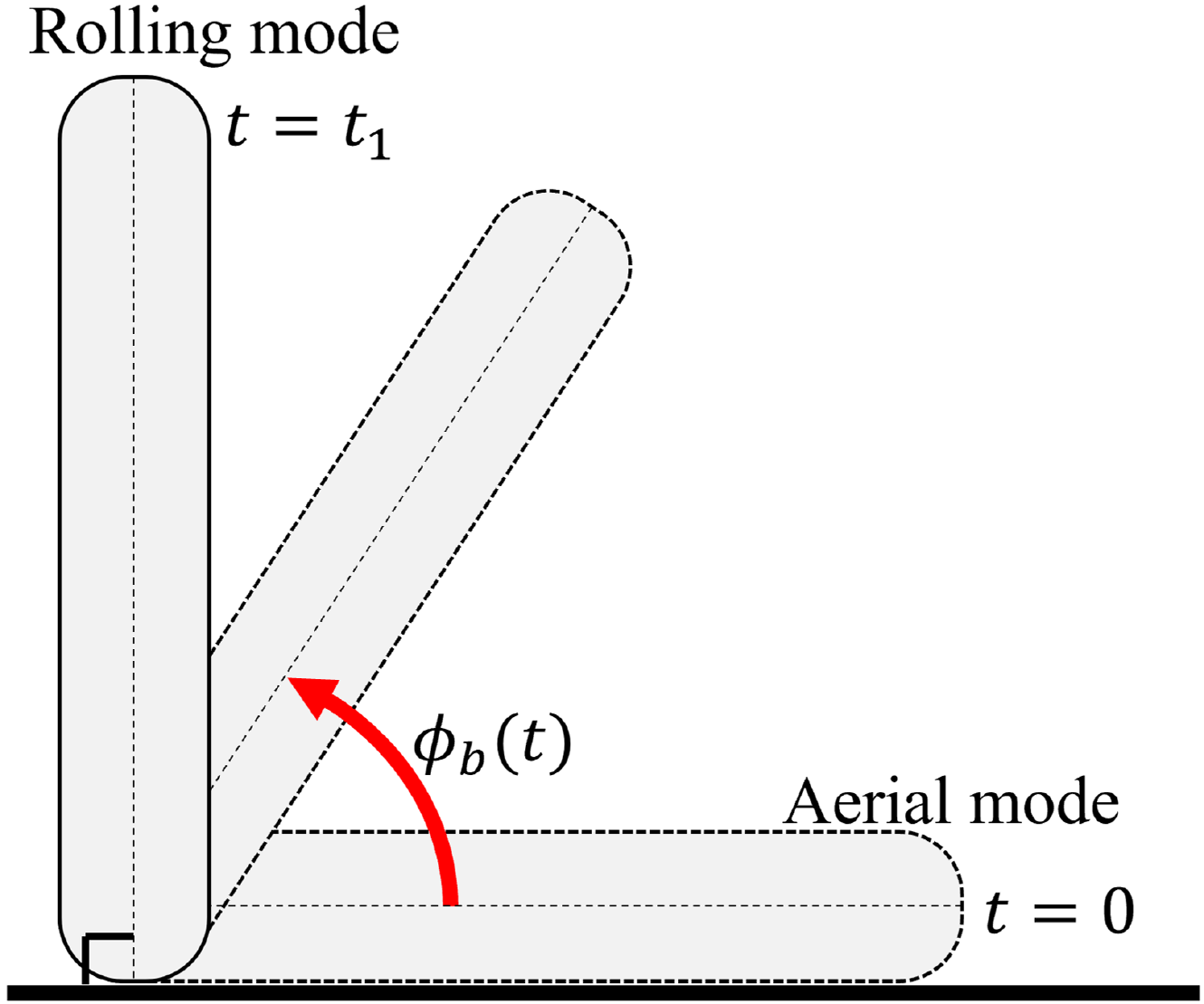}
          \subcaption{Diagram to show the standing up motion.}
          \label{figure:standing_up_trajectory}
      \end{minipage}
      \begin{minipage}[t]{0.45\columnwidth}
          \centering
          \includegraphics[width=1.0\columnwidth]{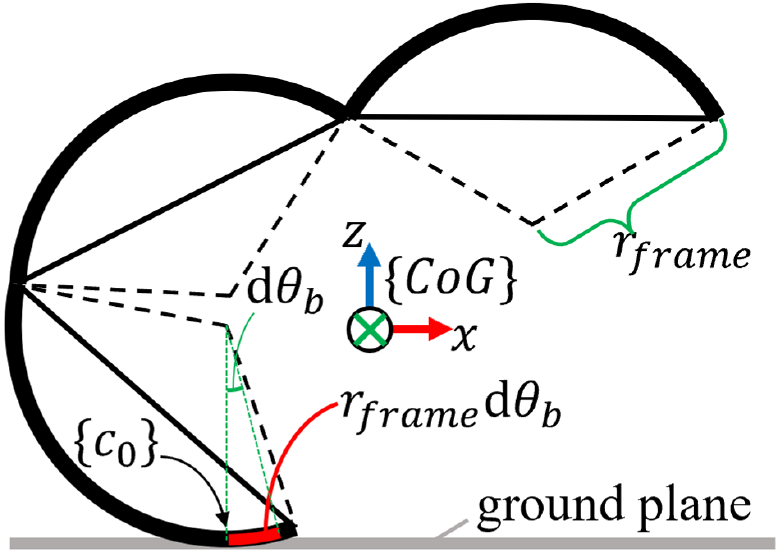}
          \subcaption{Diagram to show the \revise{rolling jacobian}.}
          \label{figure:rolling_jacobian}
      \end{minipage}
  \end{tabular}
  \caption{Motions in terrestrial mode.}
 \vspace{-5mm}
\end{figure}

\switchlanguage{
The pose of the robot \revise{differs between aerial and terrestrial} mode\revise{s}.
In the \revise{aerial} mode shown in \figref{figure:robot_configuration}(a), the robot keeps a \revise{configuration} in which all links \revise{lie in} the horizontal plane \revise{so that} the CoG is \revise{surrounded by} \revise{the thrusters} for stable flight.
\revise{In contrast}, in the \revise{terrestrial} mode shown in \figref{figure:robot_configuration}(b), all links \revise{lie in} the vertical plane to keep contact between \revise{the} frames and \revise{the} ground.
To switch \revise{from the aerial mode to the terrestrial mode}, \revise{the robot must execute the} stand-up motion as \revise{illustrated} in \figref{figure:standing_up_trajectory}.
In this part, we propose its \revise{strategy}.
}{
本研究で提案する機体は, そのlocomotion modeに応じて, とる姿勢が大きく異なる.
具体的には, \figref{figure:robot_configuration}(a)に示すように, 飛行時には安定した飛行のため機体重心をロータで囲むため全リンクが水平面上に存在する姿勢を取る一方で, 転がり動作を行う際には, \figref{figure:robot_configuration}(b)に示すように各リンクの外側に配置したプロペラガードで接地するためリンクが鉛直面上に存在する姿勢を取る.
このlocomotion modeの変更のためには\figref{figure:standing_up_trajectory}のような立ち上がり動作が必要であり, この項ではその動作の生成方法について述べる.
}

\subsubsection{Attitude \revise{T}rajectory \revise{G}eneration}
\label{sec::minimum_jerk}
\switchlanguage{
\revise{During} the stand\revise{-}up motion, the robot has to keep in the standing pose, so  overshoot \revise{in attitude should be suppressed}.
\revise{Because} \revise{e}ach actuator has limit\revise{ed} responsiveness, it is difficult to \revise{track} motion\revise{s} that \revise{require rapid} changes \revise{in} the target acceleration.
In addition, near the target attitude, \revise{the desired velocity should converge to zero}, so it is necessary to exert force \revise{for deceleration}.
For th\revise{ese} reason\revise{s}, we generate a\revise{n attitude} trajectory that minimizes the jerk.
In the aerial robotics field, trajectory generation \revise{methods} that minimize the snap have been \revise{applied} \cite{kumar2011minimumsnap}.
In this work, we apply \revise{such a formulation}  to the whole\revise{-}body \revise{attitude}.

Roll angle trajectory of baselink (\(\phi_b\)) during standing up motion is defined \revise{as the} following \revise{\(N\)-th order} polynomial\revise{,}
}{
この姿勢遷移では起き上がった姿勢でとどまる必要があり, オーバーシュートを抑制する必要がある.
各アクチュエータにはその応答性に限界があり, 急激に目標加速度が変わるような動作をすることは難しい.
また, 目標の姿勢付近では, 目標速度を\(0\)にしていくことと減速する方向に力を出すことが有効である.
以上の理由から, 本研究では, 躍度(加加速度)を最小にするような軌道に沿って, 姿勢遷移を行う.
飛行ロボットの軌道生成ではsnap(加加加速度)を最小化するような軌道生成\cite{kumar2011minimumsnap}が提案されてたり, 軸駆動ロボットで躍度を最小化する関節角軌道での変形\cite{kyriakopoulos1988minimumjerk}が提案されていたりする.
本研究ではこれを機体姿勢に適用する.

機体を起き上がらせる際の\(\phi_b\)の軌道を以下の多項式で定義する. 
}
\begin{equation}
    \label{eq:baselink_trajectory}
    \phi_b^{des}(t) = \sum_{i=0}^{N}a_i t^i.
\end{equation}
\switchlanguage{
We define the trajectory that minimizes the jerk under the following conditions.
}{
このもとで, ジャークが最小になる軌道を以下のように定義する.
}
\begin{equation}
    \label{eq:minjerk_trajectory}
    \underset{a_0, \cdots, a_N}{\text{minimize}} \hspace{2mm} \int_{0}^{t_1} \dddot{\phi}_b^2(t) \dd t \\
\end{equation}
\switchlanguage{
The time when the body stands up is set as \(t_1\), and the angle at this time is set as \(\phi_{b}(t_1) = \pi / 2\).
The angular velocity and angular acceleration are 0 at \(t=0\) and \(t=t_1\), so the following boundary conditions are obtained.
}{
境界条件として機体が起き上がる時間を\(t_1\)とし, このときの角度を\(\phi_{b}(t_1) = \pi / 2\)とする.
\(t=0\)と\(t=t_1\)で角速度, 角加速度は0となるべきなので以下の境界条件を得る.
}
\begin{equation}
 \begin{aligned}
   &\phi_b(0) = 0, &&\dot{\phi}_b(0) = 0, &&\ddot{\phi}_b(0) &= 0 \\
   &\phi_b(t_1) = \pi / 2 , &&\dot{\phi}_b(t_1) = 0, &&\ddot{\phi}_b(t_1) &= 0
 \end{aligned}
 \label{eq:standing_trajectory_constraint}
\end{equation}
\switchlanguage{
We solve the linear least squares problem with \(N+1\) variables and \revise{six} constraints \revise{in \eqref{eq:minjerk_trajectory}--\eqref{eq:standing_trajectory_constraint} }to \revise{calculate desired trajectory}.
}{
これらの条件を満たすように係数\(a_i\)を求める.
変数が\(N\)個, 制約条件はここでは6個の線形最小二乗問題を解く.
}

\subsubsection{State \revise{T}ransition}
\label{section:state transition}
\switchlanguage{
During ground motion, \revise{when} the attitude is near the vertical, \revise{we impose} constraint\revise{s} on thrust vectoring angles to suppress excessive movement as described in \secref{section:additional_gimbal_constraint}.
Although th\revise{ese} constraint \revise{improves} stability, \revise{it also} narrows the feasible region, and may \revise{hinder} solvability \revise{when} the \revise{required moment} changes with the attitude.
Therefore, th\revise{ese} constraint\revise{s} \revise{are} removed during the stand\revise{-}up motion.
We \revise{switch} the control mode to VERTICAL\_STATE \revise{(}with \revise{the} constraint\revise{)} and STANDING\_STATE \revise{(}without \revise{the} constraint\revise{)} according to the baselink \revise{roll angle}.
Depending on the roll angle of the baselink \(\phi_{b}\), we change the mode as follows,
}{
地上動作において姿勢が垂直近傍のときは\secref{section:additional_gimbal_constraint}で述べたように, 推力方向角の過剰な変化を抑制するための角度制約を設けることとしている.
一方で, この制約は, 安定性には寄与するが, 解空間を狭めており, 姿勢の変化に伴って目標発揮トルクが変化するような動作時には可解性を損なう可能性がある.
そのため, 起き上がり動作中にはこの制約を取り除くべきである.
そこで, ベースリンクの姿勢に応じて, この制約を追加するVERTICAL\_STATEと, そうでないSTANDING\_STATEの制御モード変更を行う.
ベースリンクのroll角\(\phi_{b}\)応じて以下のようにモードを変える.
}
\begin{equation*}
 \label{eq:state transition}
 \text{STATE} =
  \left\{
   \begin{array}{ll}
    \text{STANDING\_STATE}  & \qty(|\abs{\pi/2 - \abs{\phi_b}} > \phi_{\alpha})\\
    \text{VERTICAL\_STATE} & \qty(|\abs{\pi/2 - \abs{\phi_b}} < \phi_{\alpha})
   \end{array}
  \right. ,
\end{equation*}
where \(\phi_{\alpha}\) is the threshold angle to switch the state and determined experimentally.

\subsection{Fullbody \revise{I}nverse \revise{K}inematics for \revise{T}errestrial \revise{M}anipulation}
\label{section:ik}
\switchlanguage{
\revise{
This \revise{sub}section proposes a fullbody inverse kinematics method for articulated robots with circular frame.
The robot in this work can change not only the joint angles, but also the orientation of contacting link while keeping \revise{ground} contact.
So we compute both the joint angles and the contacting-link \revise{orientation} to achieve the desire\revise{d} end-effector position.}
}{
このパートでは, 円形のプロペラガードを有する多関節ロボットのための全身逆運動学手法を提案する.
本研究のロボットは関節角度だけでなく, 環境接触を維持した状態であれば接触しているリンクの姿勢を変えられる.
そこで, 目標手先位置を実現する関節角と接触しているリンクの姿勢生成を行う.
}

\switchlanguage{
In this work, \revise{we consider} a two-dimensional \revise{transformable} multilinked multirotor.
\revise{As we stated in \secref{section:design methodology}, we  consider the situation} that all links \revise{lie in} a plane perpendicular to the ground during \revise{terrestrial} manipulation.
\revise{Consequently}, we control position of the end-effector \revise{with}in \revise{this} plane.

Using the target end-effector position \(\bm{p}^{des}\) and the actual position \(\bm{p}(\bm{q})\), we define the residual \revise{\(\bm{e}_{p}(\bm{q})\)} as follows,
}{
本研究では, 二次元変形マルチリンク型マルチロータを用いる. 
そして, 地上でのマニピュレーション動作時にはリンクが地面に垂直な平面上に広がっているとする. 
そのため, リンクの存在する平面内におけるエンドエフェクタ位置を制御する. 

横方向への力の発揮には機体全体の鉛直軸周りの回転を利用する。
\cite{sugihara2011solvabilityunconcernedIK}と同様に目標のエンドエフェクタ位置と実際のエンドエフェクタ位置を用いて, we define the residual of position of the end effector as follows,
}
\begin{equation*}
 \bm{e}_p(\bm{q}) = \bm{p}^{des} - \bm{p}(\bm{q}).
\end{equation*}
\switchlanguage{
Considering the rotation of the whole body around the ground contact point, the small variation of the end-effector due to the small variation of the pitch angle can be expressed as \(J_{c_{\revise{0}, pitch}} \revise{\delta} \theta_b\).
\revise{\(J_{c_{0}, pitch}\) is the Jacobian of the end-effector  w.r.t the pitch angle around the contact point.}
Furthermore, when the circular shaped frame rolls keeping contact with ground, the position of the body is changed as shown in \figref{figure:rolling_jacobian}.
So the change of the end-effector position due to rolling is:
}{
接地点における回転関節を考えると, 姿勢の微小変位によるエンドエフェクタの微小変位は
図のように微小に回転するとき, 接地点に仮想の回転関節を考えると,\(J_{cp, pitch} \dd \theta_b\)となる.
また, この回転角\(\dd \theta_b\)と接地を維持するという条件をみたすとき, この回転にともなう変位に加えて, 接地点が動くことを考慮して,
}
\begin{equation*}
 \revise{\delta} \bm{p} = J_{cp, pitch} \revise{\delta} \theta_b +
  \begin{bmatrix}
   r_{frame} \revise{\delta} \theta_b & 0 & 0
  \end{bmatrix}^T
  := J_{\theta} \revise{\delta} \theta_b,
\end{equation*}
where \(r_{frame}\) is radius of the circular shaped link surface.

\switchlanguage{
The small variation of the joint angle\revise{s} and the whole\revise{-}body attitude \revise{required} to achieve the target end-effector position is calculated as follows,
}{
そして, 機体姿勢と関節変形を組み合わせ, 目標のエンドエフェクタ位置を実現するための関節角及び全身の姿勢角の微小変位は以下のようになる
}
\begin{equation*}
 \begin{bmatrix}
  \revise{\delta}\theta_b\\
  \revise{\delta}\bm{q}
 \end{bmatrix}
 =
 \begin{bmatrix}
  J_{\theta} & J_{q}
 \end{bmatrix}^{\#} \revise{\delta} \bm{p},
\end{equation*}
where \(J_q\) is jacobian of end-effector w.r.t. joint angle \(\bm{q}\).
\([*]^{\#}\) denotes Moore-Penrose inverse matrix.
\revise{Desired} joint angles and whole\revise{-}body attitude are updated sequentially until convergence to the target position.

\subsection{\revise{Payload Analysis for Object Grasping Flight}}
\label{section:payload_analysis}
\switchlanguage{\revise{
One aerial task envisioned for the proposed robotic platform is object grasping and transport.
Because this robot incorporates a thrust vectoring mechanism, it grasps objects by sandwiching them between its end links to avoid interference between this mechanism and the grasped object.
Previous studies have proposed grasping via joint position control and geometric analysis of graspable object \cite{HYDRUS:Chou:IJRR2018}, and a method that separately calculates thrust to achieve grasping force within the null-space while satisfying the wrench required for flight control \cite{zhao2025grasping}.
However, to the authors' knowledge, no framework has been reported that simultaneously computes joint torques and the CoG wrench for flight while generating internal forces for object grasping in a multilink multirotor.
Based on the multi-contact model described in \secref{section:modeling}, we analyze the conditions that simultaneously satisfy gravity compensation and the static equilibrium of the multilink structure, and we investigate the payload limits for object grasping flight.
}}
{
本研究で提案するロボットプラットフォームにおける空中タスクの一つとして，物体の把持・運搬タスクを想定している．
本ロボットは推力偏向機構を搭載しているため，この機構と把持対象との干渉を避ける目的で，両端のリンクで物体を挟み込む形で把持を行う．
従来研究では，関節の位置制御による把持 \cite{HYDRUS:Chou:IJRR2018} や，飛行制御に必要なレンチを満たす範囲内で，そのヌル空間上において把持力を実現するための推力を別途計算する手法 \cite{zhao2025grasping} が提案されている．
しかし，multilink multirotorにおいて物体把持のための内力を発生させながら，関節トルクと飛行制御のための重心レンチを統合的に計算する枠組みは，著者らの知る限りこれまで報告されていない．
本研究では，\secref{section:modeling} で述べた複数の接触点における接触力を考慮したモデルに基づき，重力補償と多リンク構造の静力学的つり合いを同時に満たす条件を解析し，物体把持飛行におけるペイロード限界について検討を行う．
}

\begin{figure}[t]
 \centering
 \begin{tabular}[t]{cc}
  \begin{minipage}[t]{0.5\columnwidth}
   \centering
   \includegraphics[width=1.0\columnwidth]{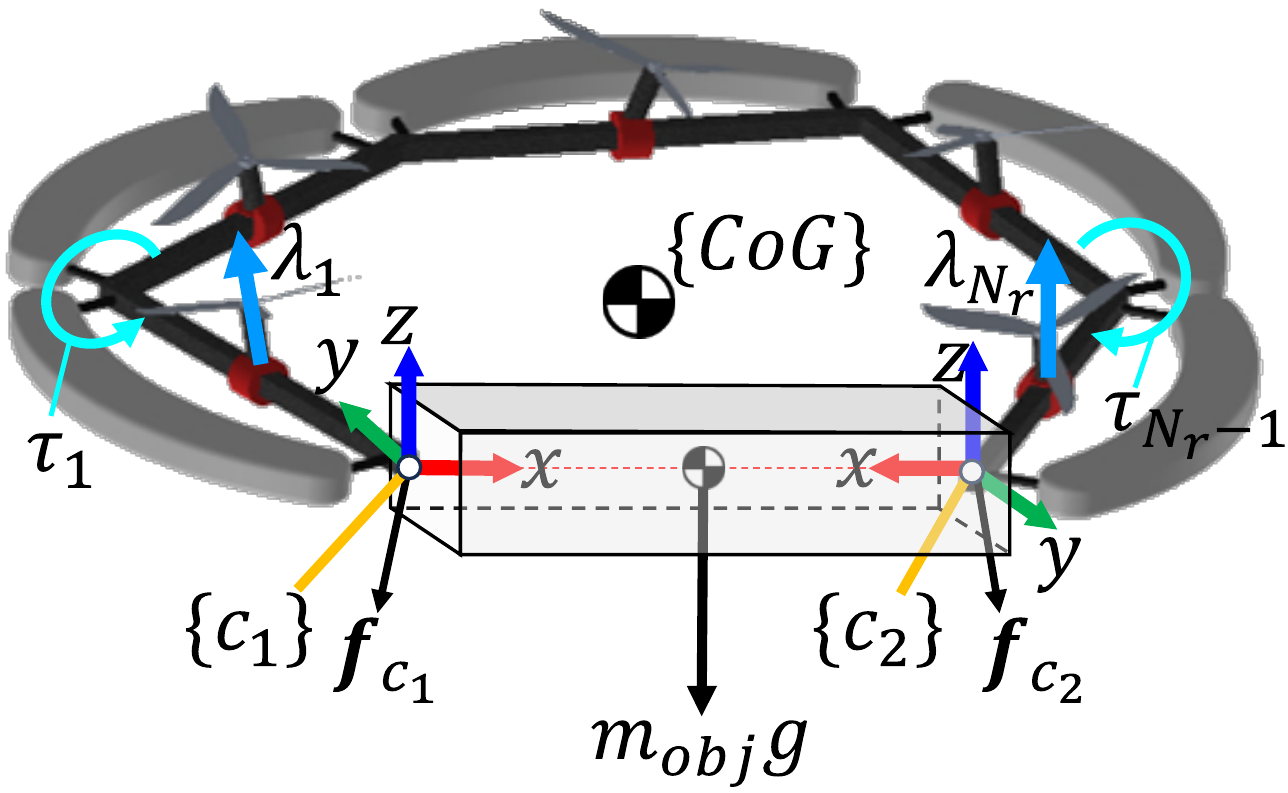}
   \subcaption{\revise{Definitions of contact point frames for ideal object grasping.}}
   \label{figure:grasping_model}
  \end{minipage}%
  \begin{minipage}[t]{0.5\columnwidth}
   \centering
   \includegraphics[width=1.0\columnwidth]{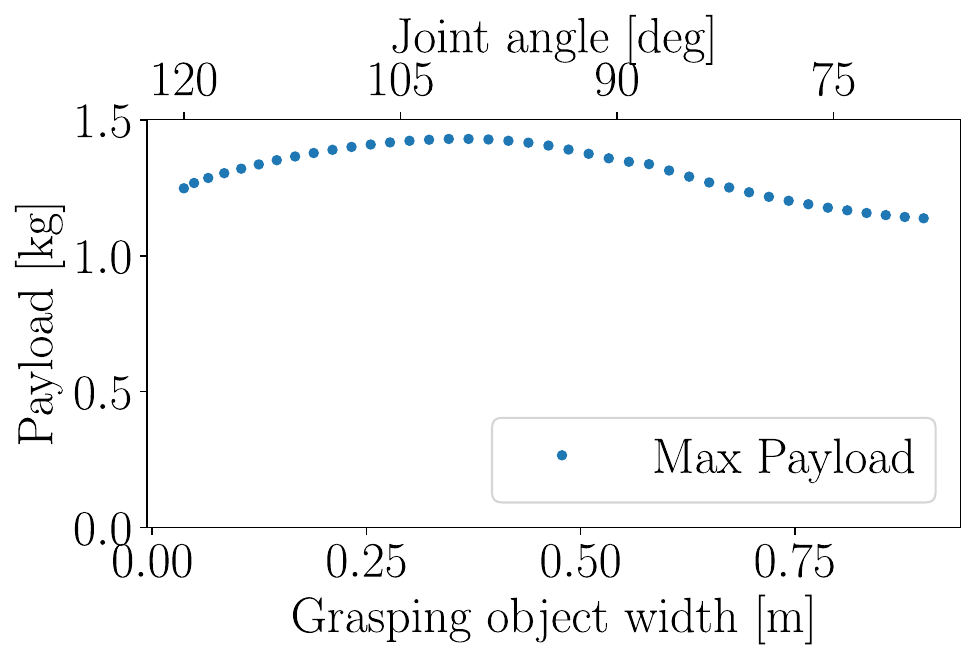}
   \subcaption{\revise{Theoretical payload for three link model.}}
   \label{figure:theoretical_payload}
  \end{minipage}
 \end{tabular}
 \caption{\revise{Analysis of theoretical payload with multi-contact model.}}
 \vspace{-5mm}
\end{figure}

\switchlanguage
{\revise{
To analyze the theoretical limits of payload, this work assumes an ideal grasping state.
In this state, the line segment connecting the two contact points passes through the object's CoG and is perpendicular to each contact surface, as shown in \figref{figure:grasping_model}.
The conditions for maintaining this grasp are: (i) the frictional force compensates for the object's gravitational force (mass), (ii) the two contact forces are in equilibrium, and (iii) each contact force lies within the friction cone, preventing slipping and separation.
Because the robot receives the reaction forces acting on the object, formulating these conditions leads to the following constraints,
}
}{
ペイロードの理論的限界を解析するため，本研究では \figref{figure:grasping_model} に示すように，2つの接触点を結ぶ線分が物体の重心を通り，かつその線分がそれぞれの接触面に垂直となる理想的な把持状態を仮定する．
このとき，把持を維持できる条件は，(i) 摩擦力により物体の重力（質量）を補償できること，(ii) 2つの接触力のつり合いが満たされていること，(iii) それぞれの接触力が摩擦錐の内部にあり，滑りおよび剥離が生じないことである．
ロボットは物体に作用する力の反力を受けるため，これらの条件を定式化すると，把持を維持するための制約条件は次のように与えられる．
}
\begin{equation}
 \begin{gathered}
  \label{eq:grasping_force}
  \revise{f_{c_{1}, x} = f_{c_{2}, x} < 0}, \hspace{5mm} \revise{f_{c_{1}, y} = f_{c_{2}, y} = 0},\\
  \revise{f_{c_{1}, z} = f_{c_{2}, z} = -\dfrac{1}{2} m_{obj} g}, \hspace{5mm} \revise{\|f_{{c_{i}}, z}\| < \mu \|f_{c_{i}, x}\|}.
 \end{gathered}
\end{equation}
\switchlanguage{
\revise{
The frame \(\{c_1\}\) and \(\{c_2\}\), that are contact points with the object, are defined as illustrated in \figref{figure:grasping_model}.
Constraints in \eqref{eq:grasping_force} are incorporated into the optimization problem shown in \eqref{chap4::eq::aerial_mode_cost_function}--\eqref{chap4::eq::aerial_mode_wrench_constraint}.
If a feasible solution is obtained as a result of solving this problem, the object of mass \(m_{obj}\) is considered capable of flight while being grasped.
}
}{
これらの制約条件を \eqref{chap4::eq::aerial_mode_cost_function}--\eqref{chap4::eq::aerial_mode_wrench_constraint} に示す最適化問題に組み込み，これを解いた結果としてfeasible solutionが得られた場合，質量\(m_{obj}\)の物体を把持した状態で飛行可能であるとみなす．
}

\switchlanguage{
\revise{
By varying the combinations of joint angles, the maximum mass that could be transported in each configuration was determined using binary search.
Here, for the three link model, the two skeletal joint angles were set to be identical.
The distance between the end links, determined by the joint angles, was regarded as the width of the grasped object.
}}{
関節角度の組み合わせを変化させ，各コンフィギュレーションにおいて把持可能な最大質量を二分探索によって求めた．
ここでは 3リンクのモデルに対して，2 つの骨格関節角度は同一とした．
関節角度によって一意に定まる両端リンク間の距離を把持対象物の幅とみなし，その幅に対して得られた最大ペイロードをプロットした結果を \figref{figure:theoretical_payload} に示す.
本プラットフォームにおけるペイロードの理論的な限界は約\SI{1.42}{kg}であることがわかった.
なお, 物体の幅が大きくなるにつれて, ペイロードは小さくなった.
これは関節角度が大きいほど, 把持物体の質量によりシステム全体の重心がスラスタがなす多角形の境界付近に位置しやすくなるためであると考えられる.
なお，本計算では推力，推力方向角，関節トルク，2 つの接触力を合わせた\(N_r + N_r + (N_r - 1) + 3 * 2 = 3N_r + 5\)個の変数を同時に最適化している.
feasible solutionが得られたコンフィギュレーション全体の平均した求解時間は\SI{0.2413}{ms}であった.
}

\revise{
The relationship between object mass, object width, and joint angle, are plotted in \figref{figure:theoretical_payload}.
The theoretical payload limit on this platform was identified as approximately \SI{1.42}{kg}.
Additionally, the payload decreased as the width of the object increased.
This is because, with larger joint angles, the mass of the grasped object causes the system's overall CoG to be more likely to be located near the boundary of the polygon formed by the thrusters, resulting in thrust saturation.
The average computation time for all configurations for which feasible solutions were obtained was \SI{0.2413}{ms}.
}

\subsection{\revise{Analysis for Force at End-effector}}
\label{section:theoretical_external_force}
\begin{figure}[t]
 \centering
 \includegraphics[width=1.0\columnwidth]{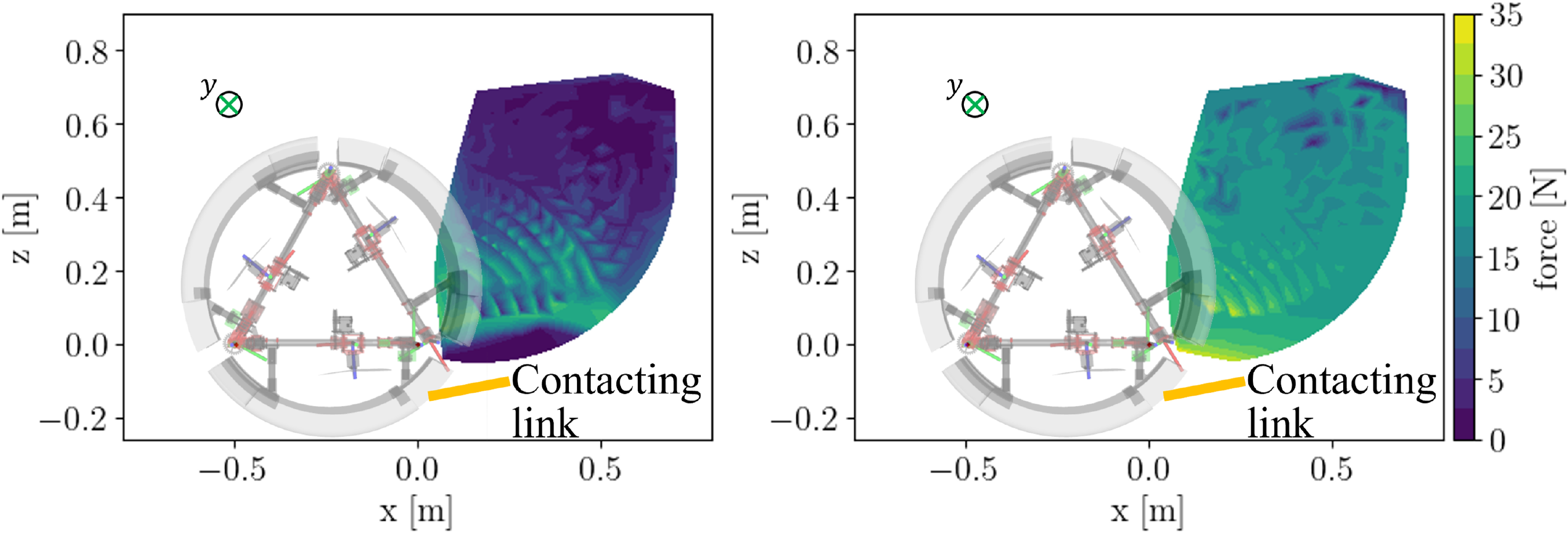}
 \caption{\revise{Theoretical exertable force at end-effector in each direction when the origin of contacting link is in visualized position. Left: In \(+x\) direction. Right: In \(+y\) direction.}}
 \vspace{-5mm}
 \label{figure:theoretical_external_force}
\end{figure}

\switchlanguage{
\revise{
We also analyze the manipulation capability on the ground.
In this analysis, we treat not only the contact force at \(\{c_0\}\) but also the contact force at the end-effector as optimization variables, in order to clarify the maximum feasible end-effector force.
As the end-effector frame, we use \(\{c_1\}\) shown in \figref{figure:robot model}(a), and assume that a contact force \(\bm{f}_{c_1}\) acts at this frame.
The rotation matrix of \(\{c_1\}\) is defined same as that of \(\{CoG\}\) frame.
Let the desired end-effector force be denoted by \(\bm{f}_{c_1}^{des}\).
We add a constraint \( \bm{f}_{c_1}^{des} = \bm{f}_{c_1}\) to the optimization problem defined in \eqref{eq:sqp_original_cost}--\eqref{chap4::eq::sqp_joint_torque_limit}, and \eqref{chap4::eq::rolling_gimbal_range_constraint}.
Then, we  optimize the thrusts, thrust vectoring angles, joint torques, and the two contact forces.
}
\revise
{In the analysis, we consider exerting forces in the \(+x\) and \(+y\) directions of the \(\{c_1\}\) frame in \figref{figure:robot model}(a).
}

\revise{
The results of this analysis are shown in \figref{figure:theoretical_external_force}.
In \figref{figure:theoretical_external_force}, the origin of the ground contacting link is fixed at the position indicated in the figure, and the plot represents the end-effector positions that can be reached by joint deformation and varying root link attitude, together with the corresponding maximum end-effector force.
For all configurations where a feasible solution exists, the optimization problem could be solved in within \SI{2}{ms}.
The maximum feasible forces were about \SI{26.3}{N} in the \(x\)-direction and about \SI{32.9}{N} in the \(y\)-direction.
As can be seen from \figref{figure:theoretical_external_force}, in the pushing direction (the \(x\)-direction), the exertable force becomes smaller when the links are highly extended.
This is because the moment generated by the end-effector force makes it difficult to satisfy the attitude control in the rolling direction.
In contrast, the lateral force in the horizontal direction (the \(y\)-direction) can be generated mainly through thrust vectoring, so relatively large forces can be exerted even when the end-effector is extended.
}}{\revise{
地上におけるマニピュレーション動作においてもスペック解析を行う.
ここでは, 地上との接地点である\(\{c_0\}\)における接触力だけでなく, 手先における接触力も最適化変数として扱うことで, 発揮可能な最大手先力を明らかにする.
手先フレームとしては，\figref{figure:robot model} に示す \(\{c_1\}\) を用い，このフレームに接触力 \(\bm{f}_{c_1}\) が作用すると仮定する。
目標手先力を\(\bm{f}_{c_1}^{des}\)として, 以下の制約条件を\eqref{eq:sqp_original_cost}--\eqref{chap4::eq::sqp_joint_torque_limit} に示す最適化問題に追加し, 推力, 推力方向角, 関節トルク, 2つの接触力を最適化する.
\begin{equation}
 \bm{f}_{c_1}^{des} = \bm{f}_{c_1}
\end{equation}
}

\revise{
解析では\figref{figure:robot model}における\(\{c_1\}\)座標系の\(+x\)方向および\(+y\)方向に力を発揮する場合を対象とする.}

\revise{
解析の結果を\figref{figure:theoretical_external_force}に示す.
\figref{figure:theoretical_external_force}は, root link原点を図中に示す場所に固定した状態で関節変形とroot linkの姿勢変化により到達可能な手先位置とその位置における最大手先力を表している.
力の発揮方向を \(x\) 方向・\(y\) 方向のいずれに設定した場合でも，feasible solution が存在するコンフィギュレーションでは，おおよそ \SI{2}{ms} 以内で目標アクチュエータ入力を算出することができた。
最大発揮力を比較すると，\(x\) 方向では約 \SI{26.3}{N}，\(y\) 方向では約 \SI{32.9}{N} となった。
また，\figref{figure:theoretical_external_force} から分かるように，押し方向（\(x\) 方向）の力については，リンクを大きく伸ばした姿勢では発揮力に起因するモーメントが増大し，転がり方向の姿勢制御が難しくなるため，発揮可能な力が小さくなる傾向がある。
一方で，水平方向の力（\(y\) 方向）は推力偏向によって生成できるため，手先を伸ばした位置においても比較的大きな力を発揮可能であることが明らかになった。

\revise{また, \secref{section:payload_analysis}で明らかにしたペイロードと比較を行うと, }
}}\\

\revise{From the analysis in \secref{section:payload_analysis} and \secref{section:theoretical_external_force}, we can also compare the achievable manipulation forces in aerial and terrestrial modes.
In aerial object grasping with two end-effectors and a relatively symmetric, stable payload shape, the maximum feasible force was less than \SI{15}{N}.
In contrast, during terrestrial manipulation, the proposed robot can exert more than \SI{30}{N} in the direction for the tasks comprising motion primitive of Swipe mentioned in \secref{section:design methodology}.
If we hypothetically tried to apply a force of around \SI{30}{N} at the end-effector in flight, it would be difficult to simultaneously satisfy gravity compensation for hovering and moment equilibrium within the thrust limit described in \secref{section:hardware}.
These results indicate that exploiting contact with the environment significantly increases the feasible manipulation forces and highlights the concept of the proposed air-ground hybrid system.
}

\section{System Architecture}
\label{section:system}
In this section, we \revise{describe} \revise{the hardware and software} architecture of the \revise{implemented} multilinked multirotor prototype.

 \subsection{Hardware}
\label{section:hardware}
\begin{figure}[t]
    \centering
    \includegraphics[width=0.99\columnwidth]{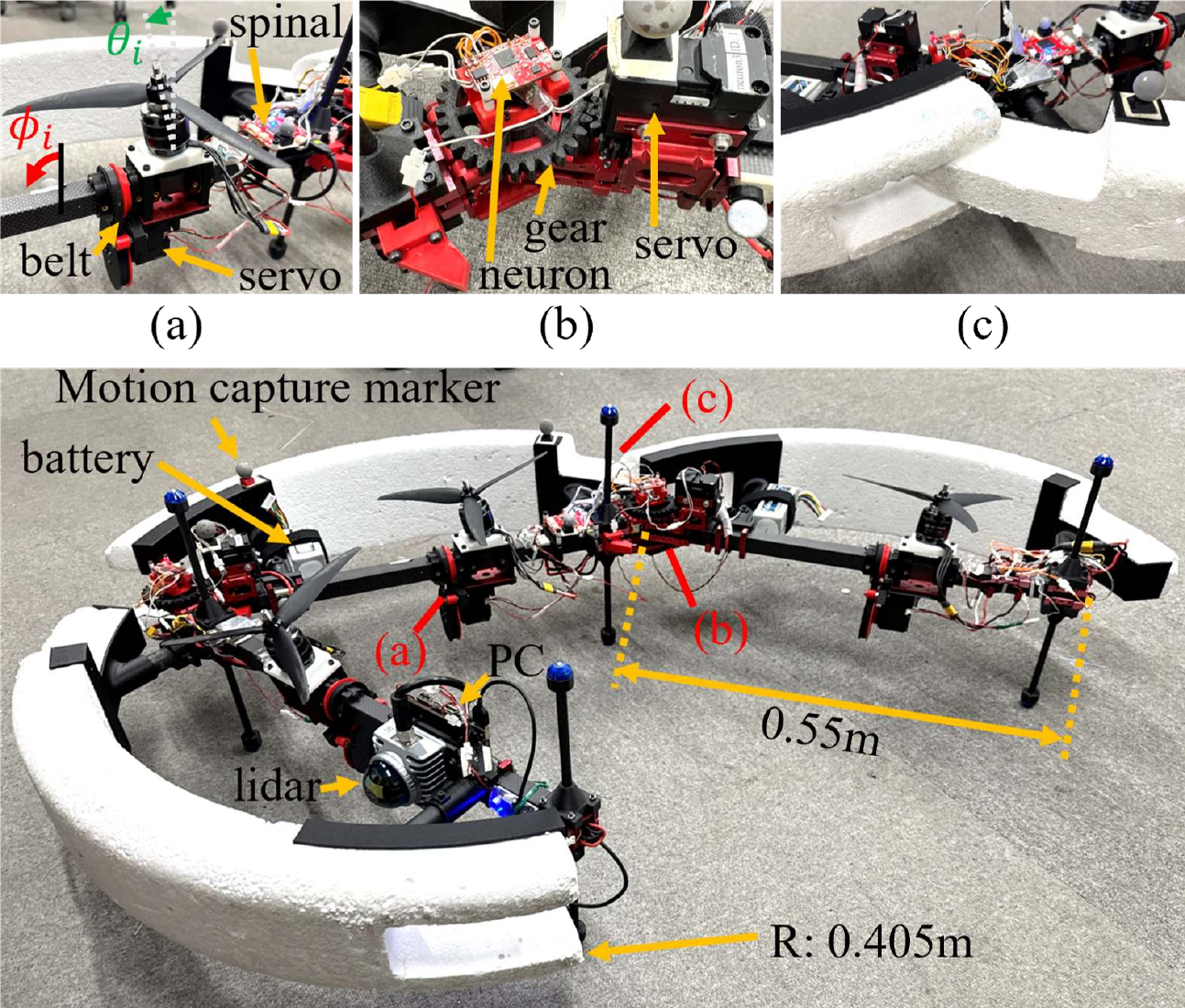}
    \caption{Implemented multilinked multirotor. (a): Thruster. (b): Link joint. (c): Outer frame for environmental contact.}
    \vspace{-5mm}
    \label{figure:prototype figure}
 \end{figure}

\switchlanguage{
\revise{The implemented prototype} is shown in \figref{figure:prototype figure}.
For \revise{a} lightweight design, \SI{20}{mm} square carbon pipes with \revise{a} thickness of \SI{1}{mm}, a mass of \SI{55}{g}, and a length of \SI{550}{mm} were used for the link skeleton.
The \revise{contact frame had a} circular shaped \revise{side and} was made of cut styrofoam, with an outer side radius of \SI{405}{mm}, a mass of \SI{64}{g}, and a volume of \SI{4366}{cm^3}.
\revise{The total mass of robot was \SI{3.65}{kg}.}
In addition, in order to avoid interference with \revise{the} frames of adjacent links during \revise{transformation}, there was a slit as shown in \figref{figure:prototype figure}(c).
\revise{The} other structural \revise{components} \revise{consist} of machined and sheet parts of aluminum or 3D-printed plastic parts.
Among the 3D printed parts, \revise{those} that require \revise{high} rigidity \revise{or} heat resistance, as well as large parts, \revise{were} made of PA6-CF, and the \revise{remaining parts were} made of PLA.

In the thruster shown in \figref{figure:prototype figure}(a), a brushless motor T-Motor AT2814 (KV 900) and \revise{Electronic Speed Controller (ESC)} T-Motor Air 40A were used.
\revise{With a 9-inch three-blade Gemfan9045 propeller}, the maximum thrust \revise{was} \SI{26.5}{N} at a \revise{supply voltage} of \SI{25.2}{V}.
\revise{The ESCs} controlled the rotation speed of b\revise{r}ushless motor according to the received PWM command.
\revise{Before the experiments}, we measured the relationship between the applied voltage, PWM command, \revise{generated} thrust, and current \revise{consumption} with these components.
\revise{The vectorable thrusters used servo motors Dynamixel XL430-W250-T, with 3D-printed PLA pulleys and timing belts to transmit rotation between different axes and provide gear ratio of 1:1.5.
}
\revise{In order to place the CoG of the rotatable part closer to the thrust vectoring axis, the brushless motor and servo motor were deployed in opposite sides of the link.}

The joint between the links is shown in \figref{figure:prototype figure}(b).
Dynamixel XH430-W350-T \revise{servo motors with gear reduction were used to drive the joints}.
The design requirements described in \secref{section:design_joint} were met.

A single-board computer \revise{(}Khadas VIM4\revise{;} Arm Cortex-A73 4 Cores at \SI{2.2}{GHz}, and Cortex-A53, 4 Cores at \SI{2.0}{GHz}\revise{)} and a 3D \revise{L}idar \revise{(}DJI Livox-MID360\revise{)} were \revise{mounted} on the third link.
A real-time attitude control board called Spinal, equipped with an STM32H746 microcontroller (1 Core, \SI{480}{MHz}), was \revise{mounted} on the second link.
Small relay boards called Neuron, equipped with an STM32F413 microcontroller (1 Core, \SI{100}{MHz}), were \revise{mounted} on each link.
\revise{Spinal and neurons are original boards using commercial microcontrollers.}
}{
\figref{figure:prototype figure}に示す, ロボットのハードウェアについて述べる.
リンクの骨格は軽量化のために\SI{20}{mm}角, 厚さ\SI{1}{mm}, 質量\SI{55}{g}のカーボンパイプを用いた.
環境接触のための円弧状のフレームは切削した発泡スチロールを用いており, 外側側面の半径は\SI{405}{mm}, 質量は\SI{64}{g}である.
また, 関節変形時に隣り合うリンクのフレームとの干渉を避けるため, \figref{figure:prototype figure}(c)に示すようにスリットが入っている.
その他の構造材にはアルミニウム製の切削部品および板金部品または3Dプリントしたプラスチック製部品を使用した.
3Dプリント部品のうち剛性や熱への耐性が必要な部分や, 大型な部品はPA6-CF製, それ以外はPLA製である.

ロボット全体と使用したアクチュエータのスペックを\tabref{chap6::table::actuators}に示す.

\figref{figure:prototype figure}(a)に示す推力装置ではT-Motor社製のブラシレスモータAT2814(KV値900)とESC T-Motor Air 40Aを用いて回転翼を回す.
9インチの3枚羽根プロペラGemfan9045 3wingを搭載した際の最大推力は\SI{26.5}{N}である.
モータドライバであるESC(Electronic Speed Controller)にはT-Motor Air40Aを用いており, 受け取ったPWM司令により回転数を制御している.
われわれは予めこれらの構成要素を推力装置に使用したときの印加電圧, PWM司令値, 発揮推力, 消費電流の関係を計測した.
推力方向を制御するためのアクチュエータにはサーボモータDynamixel XL430-W250-Tを用い, サーボモータの外部でPLA製のプーリおよびタイミングベルトで1.5倍に加速している.
これにより推力方向角の角速度を上げている.

\figref{figure:prototype figure}(b)にリンク間関節を示す.
リンク間の関節角を制御するためのアクチュエータにはサーボモータDynamixel XH430-W350-Tを用い, ギアによって減速している.
なお, \secref{section:design_joint}で述べた設計要件を満たせるようにした. 

3番目のリンクにはArm Cortex-A73(4Cores, \SI{2.2}{GHz})とCortex-A53(4Cores,\SI{2.0}{GHz})を搭載した小型シングルボードコンピュータであるKhadas VIM4(以下, オンボードPC)および三次元lidarセンサであるDJI Livox-MID360を搭載している.
また2番目のリンクにはSTM32H746マイコン(1Core, \SI{480}{MHz})を搭載したSpinalと呼ばれる中央制御基板が搭載されている.
各リンクにはSTM32F413マイコン(1Core, \SI{100}{MHz})を搭載したNeuronと呼ばれる小型中継基板が搭載されている. 
}

\begin{figure}[t]
    \centering
    \includegraphics[width=0.99\columnwidth]{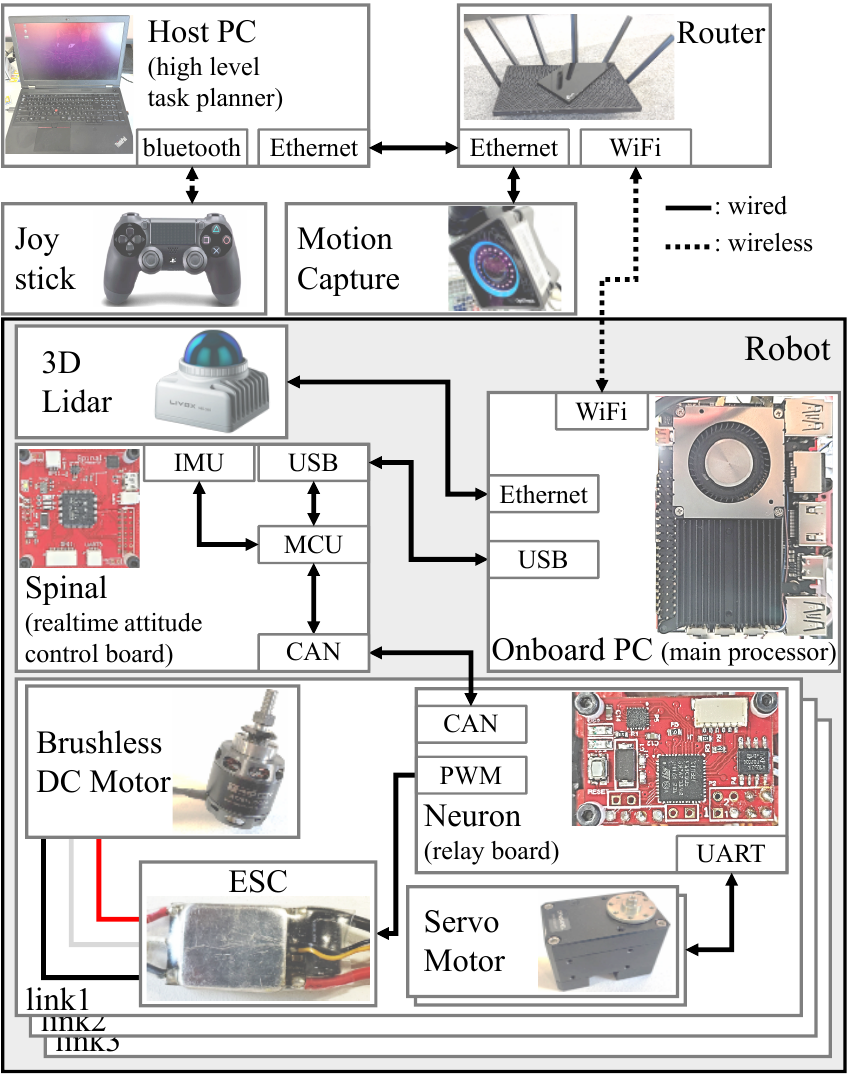}
    \caption{\revise{Architecture of hardware and software system. Actuators are distributed on each link. A communication system consisting of local network, serial, and CAN controls the actuators.}}
    \label{figure:system diagram}
    \vspace{-5mm}
 \end{figure}

 \subsection{Software}
\switchlanguage{
\revise{The architecture of hardware and software system} is shown in \figref{figure:system diagram}.
\revise{A} high\revise{-}level motion planner runs on the user-operable Host PC.
Data \revise{are} sent \revise{between the host PC and} the onboard PC via a local network.

\revise{The o}nboard PC estimates the robot state, \revise{performs} kinematics \revise{calculations}, and \revise{computes} target actuator \revise{commands} at a rate of \SI{100}{Hz}.
Spinal estimates \revise{the} attitude using the IMU \revise{mounted} on the board\revise{.}
\revise{It also} performs attitude control \revise{at \SI{500}{Hz} \revise{by adjusting the} thrust\revise{s} \revise{based on the current} configuration received from PC}.
\revise{The} Neuron \revise{boards} receive target PWM command \revise{for} each thruster from Spinal and send it to ESC.
They also communicate with \revise{the} servo motors via UART, send\revise{ing} target commands to servo motors and \revise{returning their} state to Spinal.

\revise{The} onboard PC and Spinal communicate via \revise{a} serial \revise{connection} (\SI{921600}{bps}), and Spinal and Neuron \revise{boards} communicate via the CAN (Control Area Network) \revise{bus at} \SI{1}{Mbps}.
These \revise{boards and} details of \revise{their} internal communication system are described in \cite{anzai2017multilinked}.

The nonlinear optimization problem described in \secref{section:nonlinear wrench allocation} is solved \revise{on the onboard PC} using the NLOPT \cite{NLopt} library with \revise{the} LD\_SLSQP (Sequential Least Squares Quadratic Programming) algorithm.
This algorithm can solve nonlinear constrained optimization problems \revise{by providing} gradients of \revise{the} cost \revise{function} and \revise{the} constraints.

\revise{During} indoor experiments, the pose of Spinal in the world frame was measured using the  \revise{OptiTrack} motion capture system (sampling frequency: \SI{120}{Hz}).
\revise{During} outdoor experiments, the deviation from the initial pose of the \revise{L}idar was estimated by integrating the point cloud\revise{s} from \revise{L}ivox-MID360 and IMU data of Spinal using FAST-LIO \cite{xu2021fastlio}.
}{
\figref{figure:system diagram}にしめすシステム構成を紹介する.
高次の動作計画器はユーザの操作可能なHost PC上で動いている.
データはローカルネットワークを通してonboard PCに送受信される。

オンボードPC上ではロボットの状態推定, 運動学計算, および目標のアクチュエータ入力計算などを\SI{100}{Hz}の周期で行う.
Spinalでは基板上に搭載したIMUの情報を利用してロボットの姿勢推定を行い, 推力装置の出力の制御による姿勢制御を\SI{500}{Hz}の周期で行う.
NeuronはSpinalから受信した各推力装置への目標PWM司令をESCに送信する,
また, UARTでサーボモータと通信し, サーボへの角度司令とspinalへの状態の送信をする.

オンボードPCとSpinalはシリアル通信(\SI{921600}{bps})により通信を行い, 
SpinalとNeuron群はCAN(Control Area Network)プロトコル(\SI{1}{Mbps})により通信を行う.
この体内通信系の詳細は\cite{anzai2017multilinked}に詳述されている. 

\secref{section:nonlinear wrench allocation}で述べた非線形最適化問題はオンボードPC上でNLOPT\cite{NLopt}ライブラリを用いてLD\_SLSQP(逐次二次計画法)アルゴリズムにより解く.
SLSQPは勾配のある非線形制約付き最適化問題を解くことができる. 
また, \secref{section:nonlinear wrench allocation}で述べた, 
SLSQPの初期解を求めるための線形二次計画問題の求解には交互方向乗数法に基づく二次計画問題のソルバーであるOSQP\cite{osqp}を用いた. 

屋内実験においてはモーションキャプチャシステムOptiTrack(サンプリング周波数: \SI{120}{Hz})を用いて世界座標系におけるSpinalの位置姿勢を計測した. 
屋外実験においては, livox-MID360から得られる点群データと内蔵IMUの情報をFast-LIO\cite{xu2021fastlio}で統合してlidarの初期の位置姿勢からの偏差を推定した.
}

 \begin{figure}[t]
    \centering
    \begin{tabular}{cc}
        \begin{minipage}[t]{0.5\columnwidth}
            \centering
            \includegraphics[width=1.0\columnwidth]{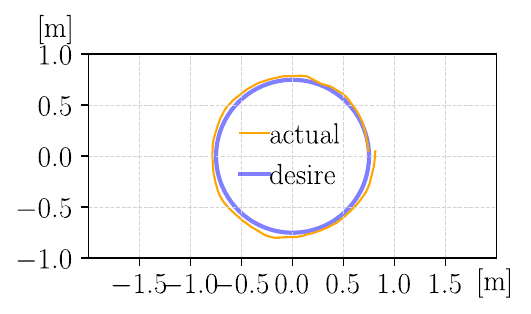}
            \subcaption{Desired circle trajectory and actual trajectory}
            \label{figure:hovering_plot_cog_xy_pos}
        \end{minipage}%
        \begin{minipage}[t]{0.5\columnwidth}
            \centering
            \includegraphics[width=1.0\columnwidth]{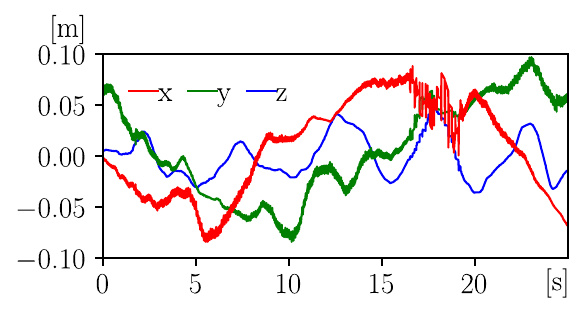}
            \subcaption{Position errors}
        \end{minipage}%
        \\
        \begin{minipage}[t]{0.5\columnwidth}
            \centering
            \includegraphics[width=1.0\columnwidth]{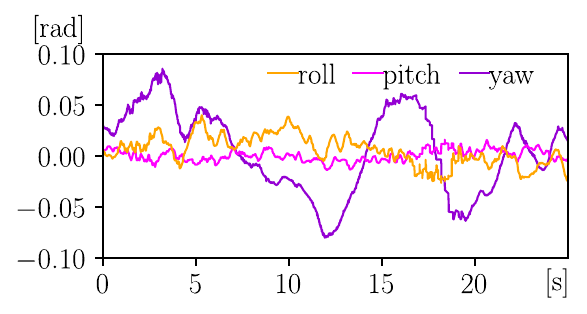}
            \subcaption{Orientation errors}
        \end{minipage}%
        \begin{minipage}[t]{0.5\columnwidth}
            \centering
            \includegraphics[width=1.0\columnwidth]{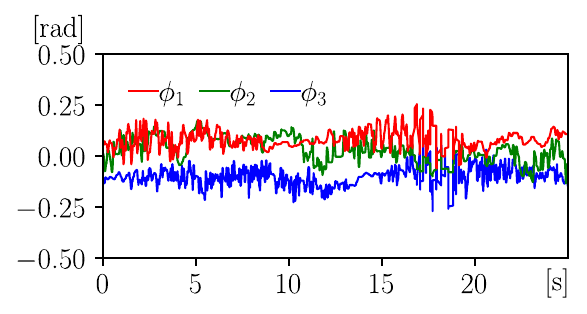}
            \subcaption{Thrust vectoring angles}
        \end{minipage}%
    \end{tabular}
    \caption{Plots related to circle trajectory tracking flight.}
    \vspace{-5mm}
    \label{figure:hovering_plot}
\end{figure}

\section{Experiment\revise{s}}
\label{section:experiment}
\switchlanguage{
    In this section, we evaluate \revise{the} proposed methods by the following experiments \revise{using the implemented prototype}.
    \begin{enumerate}
        \item Multimodal locomotion by flight and rolling, \revise{including} verif\revise{ication} of the locomotion\revise{-}mode transition and stability against disturbances. (\secref{section:locomotion_experiment})
        \item Manipulation on the ground and in the air using joint motion. (\secref{section:manipulation_experiment})
        \item \revise{An} \revise{i}ntegrated object transportation experiment that \revise{combines} locomotion and manipulation \revise{across} multiple domains. (\secref{section:integrated_experiment})
    \end{enumerate}
}{
   本節では以下の実験を行い, 提案手法を評価する.
   \begin{enumerate}
      \item 飛行と転がりによるマルチモーダルロコモーション. それに必要な姿勢遷移動作や外乱に対する安定性も検証する. (\secref{section:locomotion_experiment})
      \item 地上と空中の領域における, 関節変形を利用したマニピュレーション. (\secref{section:manipulation_experiment})
      \item ロコモーションとマニピュレーションの能力を統合した, 物体運搬統合実験. (\secref{section:integrated_experiment})
   \end{enumerate}
}

\subsection{Locomotion Ability Evaluation}
\label{section:locomotion_experiment}
\subsubsection{Circle Trajectory Tracking Flight Experiment}
\label{section:circle_trajectory_flight_experiment}
\switchlanguage{
To \revise{evaluate} \revise{the} locomotion ability in the air, a circular trajectory tracking flight experiment was conducted.
During hovering, \revise{the} target translational position and velocity were continuously sent to \revise{the} robot to follow \revise{a} \revise{counterclockwise (}CCW\revise{)} circular trajectory with a radius of \SI{0.75}{m}. 
The PID gains for flight control were set to \(K_{r,p} = \mathrm{diag}\qty(5.0, 5.0, 5.0)\), \(K_{r,i}=\mathrm{diag}\qty(1.0,1.0,1.5)\), \(K_{p,d}=\mathrm{diag}\qty(3.0, 3.0, 2.5)\), \(K_{\tau, p}=\mathrm{diag}\qty(15.0, 15.0, 4.0)\), \(K_{\tau, i}=\mathrm{diag}\qty(1.0, 1.0, 2.0)\), \(K_{\tau, d}=\mathrm{diag}\qty(2.0, 2.0, 6.0)\).
The plots in \figref{figure:hovering_plot} show the target and actual translational \revise{trajectories}, \revise{the} position tracking errors, and attitude tracking errors, \revise{and the} thrust vectoring angles.
\revise{Please note that the center of \revise{the} desired circle trajectory is plotted as \(\qty(0, 0)\) in \figref{figure:hovering_plot_cog_xy_pos}.}

\revise{The root mean square (RMS)} of these errors were [0.0466, 0.0478, 0.0209]\(\mathrm{m}\) and [0.0154, 0.00057, 0.0411]\(\mathrm{rad}\).
The maximum absolute values of these errors were [0.0885, 0.0973, 0.0479]\(\mathrm{m}\) and [0.0402, 0.0132, 0.0858]\(\mathrm{rad}\).
The robot was able to \revise{perform} translational \revise{movement} while keeping \revise{its} body horizontal, and \revise{it successfully} follow\revise{ed} the target circular trajectory.
\revise{Compared with the tracking performance for the roll and pitch angles, the yaw angle, which has low control responsiveness, exhibited a relatively large tracking error.
This low control responsiveness is due to three factors: (1) the larger inertia around the \(z\)-axis of CoG compared to other axes; (2) rotational moment around the \(z\)-axis of CoG is generated by the thrust vectoring DoF, whose tracking performance is lower than that of the thrust; (3) the short moment arm of the horizontal force component generated by thrust vectoring.
}
\revise{
The power consumption of the thrust motors, estimated from the \revise{thrust data}, the voltage measured by the voltage sensor, and PWM commands, was approximately \revise{\SI{1025}{W}}\revise{.}
Considering the duration and distance of the experiment, the energy required for aerial locomotion per unit distance was approximated to \SI{5438}{J/m}.
}}{
    空中での移動能力を実証するため, 円軌道追従飛行実験を行った.
    ホバリング状態にある機体に, 半径\SI{0.75}{m}の半時計回りの円軌道に追従するように, 目標の並進位置, 並進速度を連続的に与えた. 
    なお, 飛行動作におけるPIDゲインは, \(K_{r,p} = diag\qty(5.0, 5.0, 5.0)\), \(K_{r,i}=diag\qty(1.0,1.0,1.5)\), \(K_{p,d}=diag\qty(3.0, 3.0, 2.5)\), \(K_{\tau, p}=diag\qty(15.0, 15.0, 4.0)\), \(K_{\tau, i}=diag\qty(1.0, 1.0, 2.0)\), \(K_{\tau, d}=diag\qty(2.0, 2.0, 6.0)\)とした.
    \figref{figure:hovering_plot}に, 目標と実際の並進位置, 位置制御誤差, 姿勢誤差, 推力方向角を示す.

    RMS of these errors were 
    \(\begin{bmatrix}
        0.0466 & 0.0478 & 0.0209
    \end{bmatrix} \mathrm{m}\) and 
    \(\begin{bmatrix}
        0.0154 & 0.00057 & 0.0411
    \end{bmatrix} \mathrm{rad}\).
    The maximum absolute values of these errors were 
    \(\begin{bmatrix}
        0.0885 & 0.0973 & 0.0479
    \end{bmatrix} \mathrm{m}\) and 
    \(\begin{bmatrix}
        0.0402 & 0.0132 & 0.0858
    \end{bmatrix} \mathrm{rad}\).
    全駆動飛行制御により, 機体を水平近傍に保った状態で並進方向に力を発揮することができ, 目標の円軌道に追従することができた.
    一方で, 重心座標系に寄与するモーメントアームが短くなりやすいyaw角の制御はroll, pitch角の制御に対して比較的長い周期での振動が見られた.
消費電力平均:1025W
所要時間:25s
距離: 2 * 0.75 * pi = 4.712m
}

\begin{figure}[t]
    \centering
    \includegraphics[width=1.0\columnwidth]{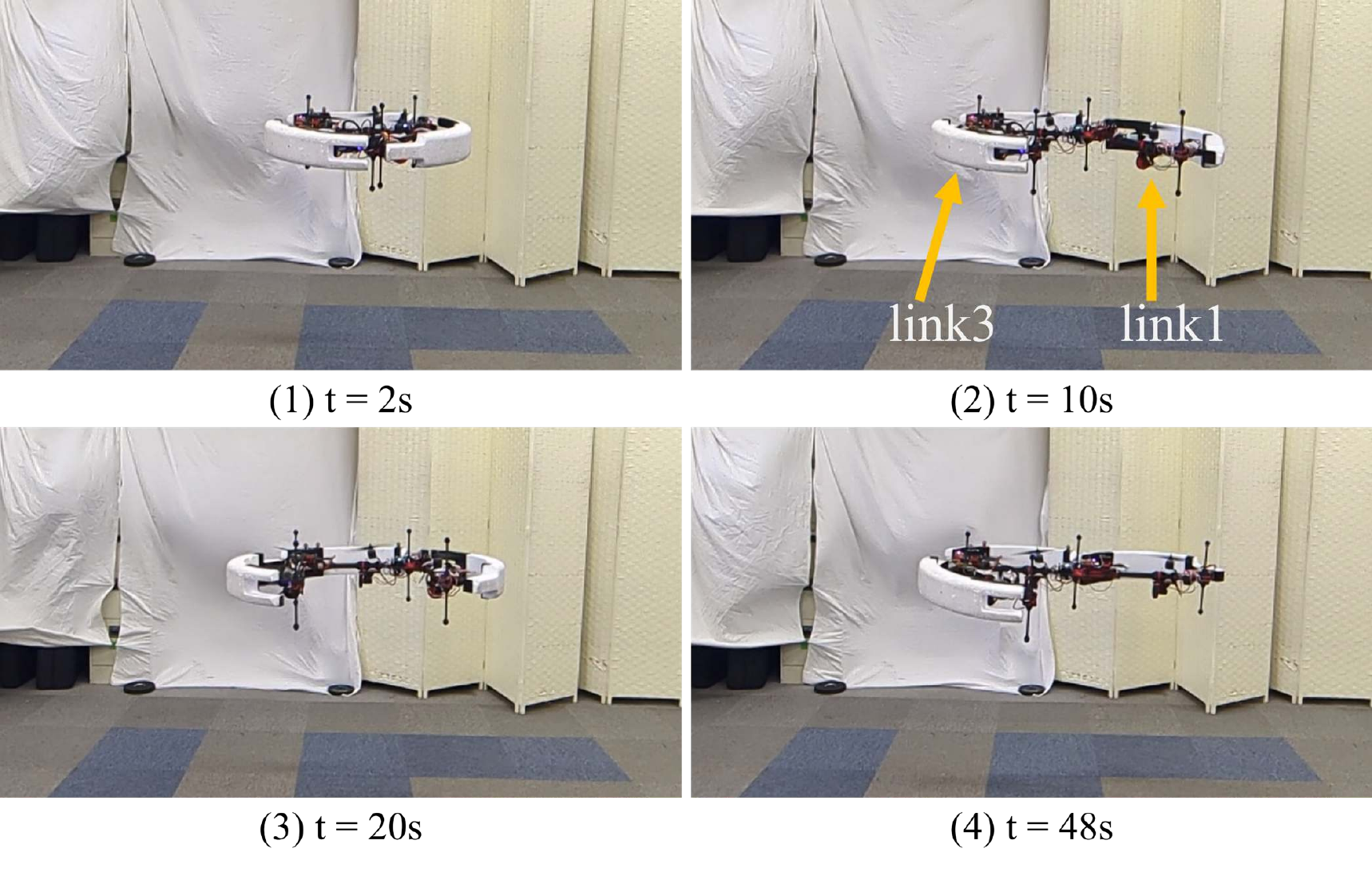}
    \caption{Aerial transformation experiment.}
    \vspace{-5mm}
    \label{figure:aerial transformation experiment}
 \end{figure}

\begin{figure}[t]
    \begin{tabular}{ll}
        \begin{minipage}[t]{0.5\columnwidth}
            \centering
            \includegraphics[width=1.0\columnwidth]{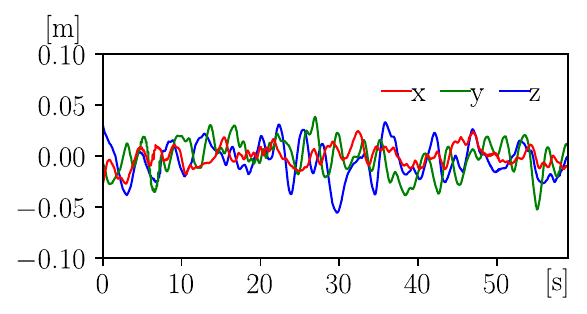}
            \subcaption{Position errors}
        \end{minipage}%
        \begin{minipage}[t]{0.5\columnwidth}
            \centering
            \includegraphics[width=1.0\columnwidth]{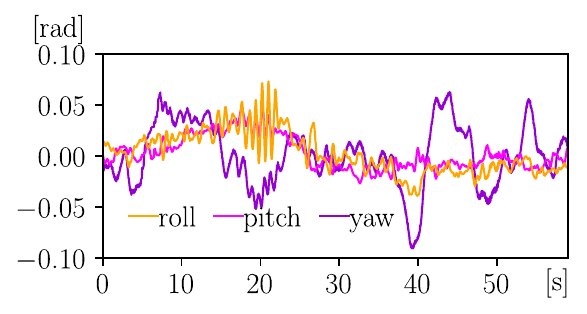}
            \subcaption{Orientation errors}
        \end{minipage}%
        \\
        \begin{minipage}[t]{0.5\columnwidth}
            \centering
            \includegraphics[width=1.0\columnwidth]{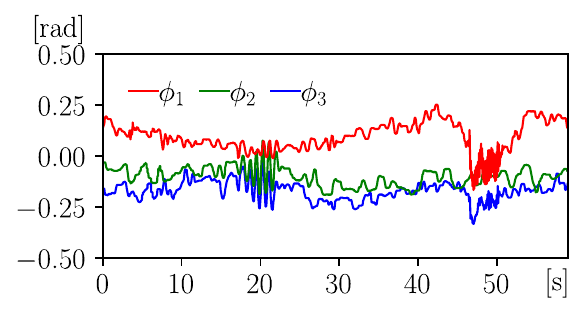}
            \subcaption{Thrust vectoring angles}
        \end{minipage}%
        \begin{minipage}[t]{0.5\columnwidth}
            \centering
            \includegraphics[width=1.0\columnwidth]{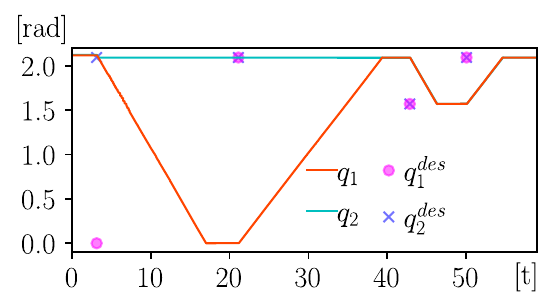}
            \subcaption{Joint angles}
        \end{minipage}%
        \\
        \begin{minipage}[t]{0.5\columnwidth}
            \centering
            \includegraphics[width=1.0\columnwidth]{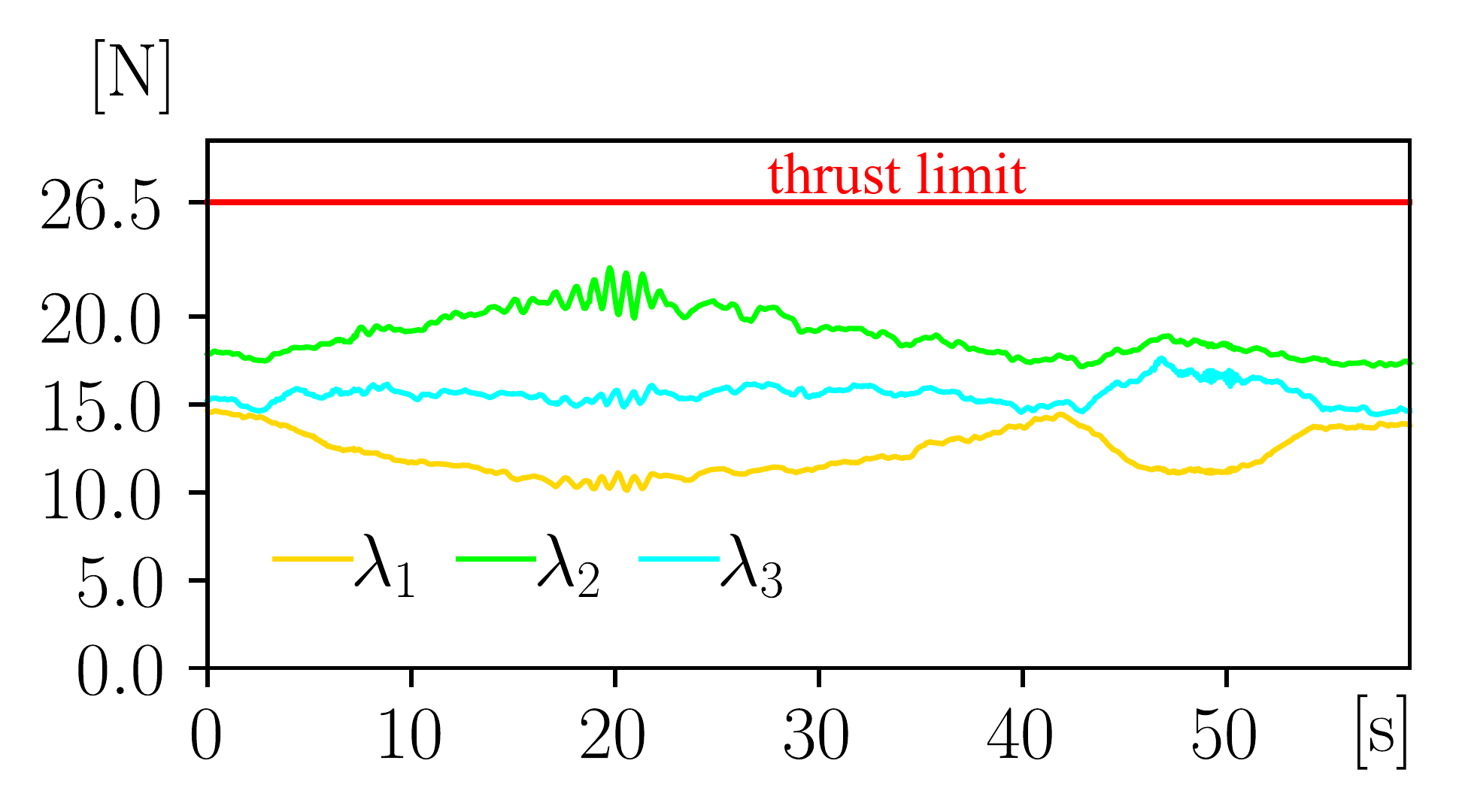}
            \subcaption{Target thrusts}
            \label{figure:aerial_transformation_pwm}
        \end{minipage}%
        \begin{minipage}[t]{0.5\columnwidth}
            \centering
            \includegraphics[width=1.0\columnwidth]{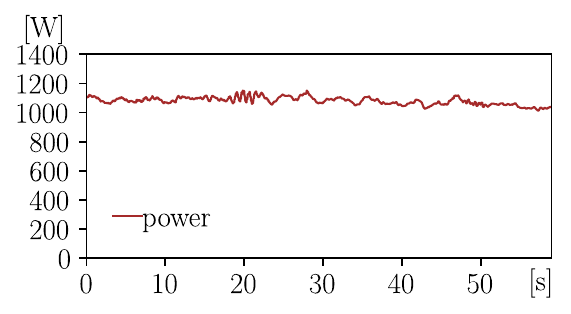}
            \subcaption{\revise{Power consumption}}
        \end{minipage}%
    \end{tabular}
    \caption{Plots related to \figref{figure:aerial transformation experiment}.}
    \vspace{-5mm}
    \label{figure:aerial transformation plot}
\end{figure}

\subsubsection{Aerial Transformation Experiment}
\label{section:aerial_transformation_experiment}
\switchlanguage{
A\revise{n} aerial transformation experiment as shown in \figref{figure:aerial transformation experiment} was conducted.
\revise{T}arget joint angle\revise{s} \revise{were commanded during hovering}.
We sent target joint angles of \(\qty[\frac{2}{3}\pi, \frac{2}{3}\pi]\), \(\qty[0, \frac{2}{3}\pi]\), and \(\qty[\frac{1}{2}\pi, \frac{1}{2}\pi]\).
\revise{The} plots in \figref{figure:aerial transformation plot} show the \revise{position} tracking errors, the \revise{attitude} tracking errors, \revise{the} thrust vectoring angles, \revise{the} joint angles, \revise{the} target thrust of each thruster, \revise{and power consumption}.

\revise{The} RMS of these errors were [0.00971, 0.0171, 0.0179]\(\mathrm{m}\) and [0.0204, 0.0164, 0.0305]\(\mathrm{rad}\).
The maximum absolute values of these errors were [0.0270, 0.0522, 0.0554]\(\mathrm{m}\) and [0.0733, 0.0493, 0.0904]\(\mathrm{rad}\).
The robot was able to \revise{maintain its} position and attitude even when the \revise{body} configuration  changed due to joint motion.
\revise{The average estimated power consumption by thrust motors during this experiment was \SI{1079}{W}.}
\revise{
One reason for this relatively high value is that the tilted propellers and the thrust vectoring DoF reduce the effective thrust component in the anti-gravity direction.
In addition, the use of servo motors for the joints, the contact frame structure, and relatively small propellers (chosen to avoid mechanical interference) also lowers the propulsion efficiency.
Although the power consumption was high, the results indicate that the thrusts are not fully saturated.
Each thruster is capable of generating thrust up to \SI{26.5}{N}, whereas in this experiment, the maximum thrust exerted even in the challenging configuration was approximately \SI{23}{N}, leaving about a \SI{10}{\%} margin.
}
As shown in \figref{figure:aerial_transformation_pwm}, the exerted thrust of first thruster decreased as the configuration changed from \figref{figure:aerial transformation experiment}(1) to \figref{figure:aerial transformation experiment}(3).
This is because the distance from the \revise{CoG} to the first \revise{thruster} increased, and the rotational moment generated per unit thrust increased, so the effect was suppressed.
In addition, in the attitude shown in \figref{figure:aerial transformation experiment}(3), the roll angle control of CoG became oscillatory.
\revise{The} thrust vectoring angle\revise{s} and thrust command value\revise{s} \revise{also} became oscillatory.
This \revise{occurred} because the distance of each \revise{thruster} \revise{from the CoG in} the \(y\) direction became small, \revise{reducing} the control performance compared \revise{to} other axes.
}{
    A aerial transformation experiment as shown in \figref{figure:aerial transformation experiment} was conducted.
    空中において目標の関節角度を与える. 
    与えた目標関節角は, \(\qty[\frac{2}{3}\pi, \frac{2}{3}\pi]\), \(\qty[0, \frac{2}{3}\pi]\), \(\qty[\frac{1}{2}\pi, \frac{1}{2}\pi]\)である.
    Plots in \figref{figure:aerial transformation plot} show the tracking errors of positions, tracking errors of orientations, 推力方向角, joint angles, and input PWM commands to each thrusters.

    RMS of these errors were 
    \(\begin{bmatrix}
       0.00971 & 0.0171 & 0.0179
    \end{bmatrix} \mathrm{m}\) and 
    \(\begin{bmatrix}
       0.0204 & 0.0164 & 0.0305
    \end{bmatrix} \mathrm{rad}\).
    The maximum absolute values of these errors were 
    \(\begin{bmatrix}
       0.0270 & 0.0522 & 0.0554
    \end{bmatrix} \mathrm{m}\) and 
    \(\begin{bmatrix}
       0.0733 & 0.0493 & 0.0904
    \end{bmatrix} \mathrm{rad}\).
    関節変形により機体のフィギュレーションが変化しても位置と姿勢の制御を安定化することができている。
    予め測定したスラスタのデータと, 電圧センサで測定した電圧, PWM司令値より, power consumption in flight was approximately \SI{1300}{W}.
    \figref{figure:aerial_transformation_pwm}に示すように, \figref{figure:aerial_transformation_experiment_1}から\figref{figure:aerial_transformation_experiment_3}に変化する過程で, 1番目のリンクの発揮推力が小さくなっていった. 
    これは1番目のリンクが重心から遠くなり, 単位推力あたりに発生する回転モーメントが大きくなったため, その影響を抑えるためである.
    また, \figref{figure:aerial_transformation_experiment_3}に示す姿勢では, 重心座標系のroll角の制御が振動的になった.
    また, それに一致して推力方向角度や推力司令値も振動的になった.
    これは重心からみた, 各ロータの\(y\)方向の距離が小さくなり, 他の軸と比較して制御性が低下したためであると考えられる.
}

\begin{figure}[t]
    \centering
    \includegraphics[width=1.0\columnwidth]{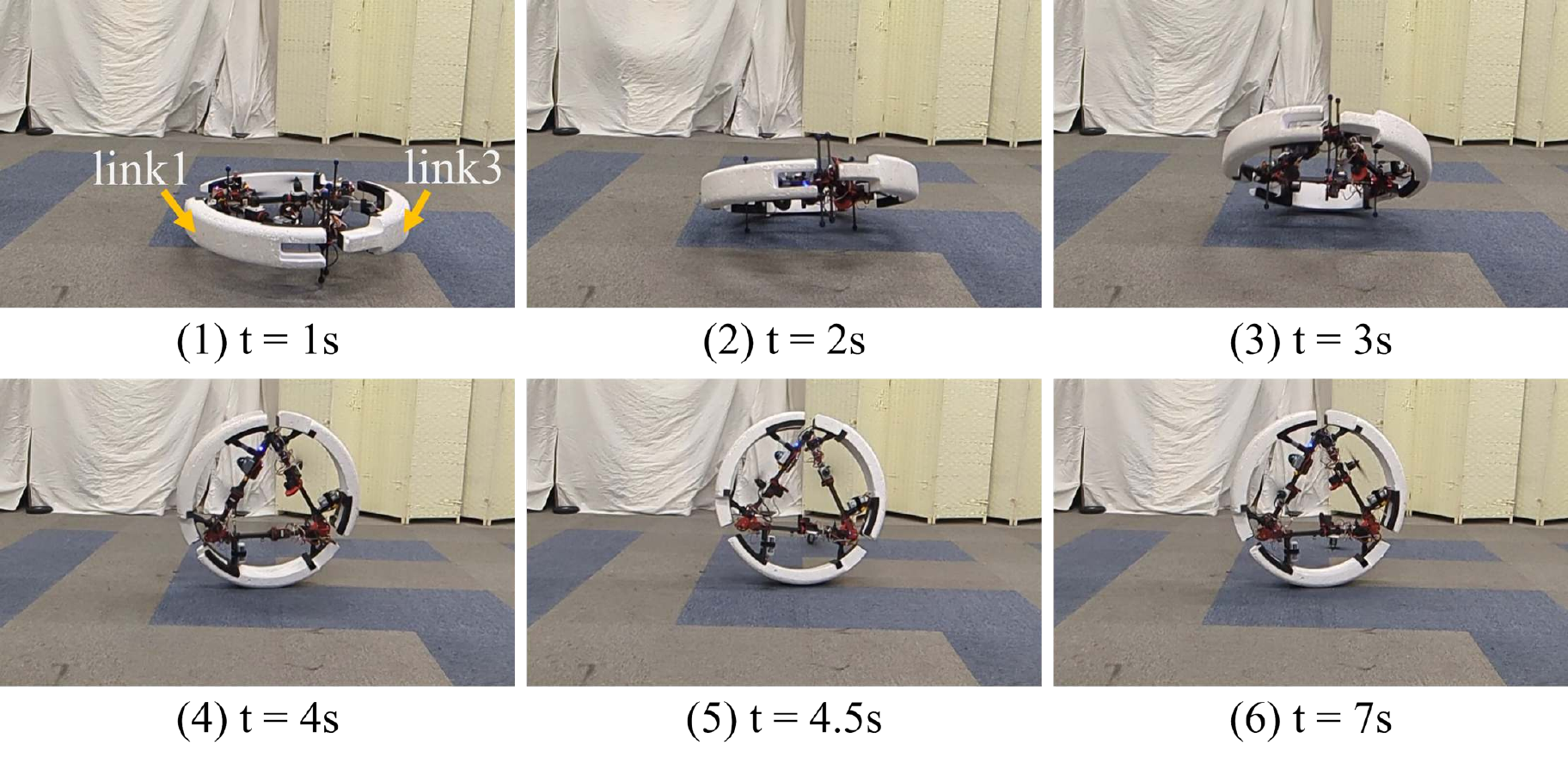}
    \caption{Standing experiment.}
    \label{figure:standing_experiment}
    \vspace{-5mm}
\end{figure}

\begin{figure}[t]
    \begin{tabular}{ll}
        \begin{minipage}[t]{0.5\columnwidth}
            \centering
            \includegraphics[width=1.0\columnwidth]{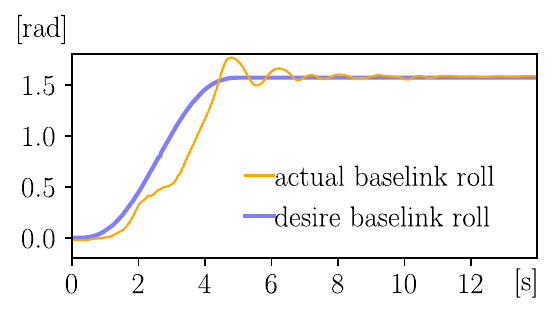}
            \subcaption{Target and actual roll angle of baselink (minimum jerk trajectory)}
            \label{figure:standing_plot:baselink_roll}
        \end{minipage}%
        \begin{minipage}[t]{0.5\columnwidth}
            \centering
            \includegraphics[width=1.0\columnwidth]{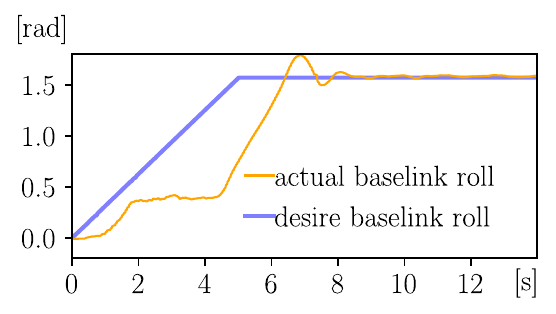}
            \subcaption{Target and actual roll angle of baselink (linear interpolation)}
            \label{figure:standing_plot:baselink_roll_linear}
        \end{minipage}%
        \\
        \begin{minipage}[t]{0.5\columnwidth}
            \centering
            \includegraphics[width=1.0\columnwidth]{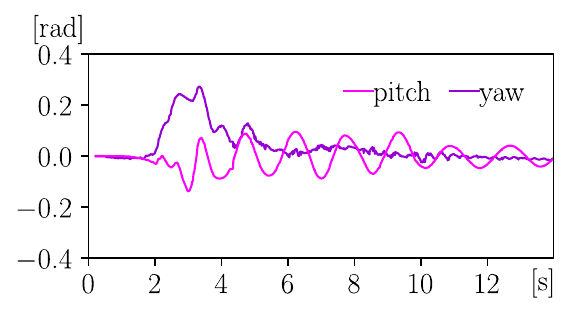}
            \subcaption{Tracking error of pitch and yaw angle}
            \label{figure:standing_plot:pitch_yaw_error}
        \end{minipage}%
        \begin{minipage}[t]{0.5\columnwidth}
            \centering
            \includegraphics[width=1.0\columnwidth]{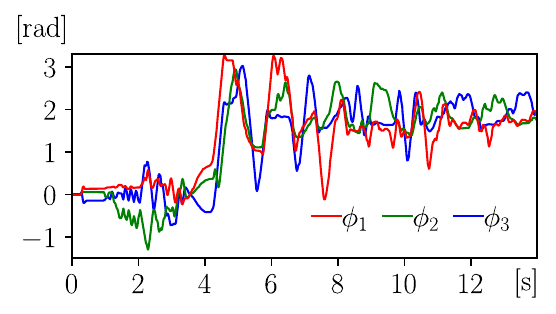}
            \subcaption{Thrust vectoring angles}
            \label{figure:standing_plot:gimbal}
        \end{minipage}%
        \\
        \begin{minipage}[t]{0.5\columnwidth}
            \centering
            \includegraphics[width=1.0\columnwidth]{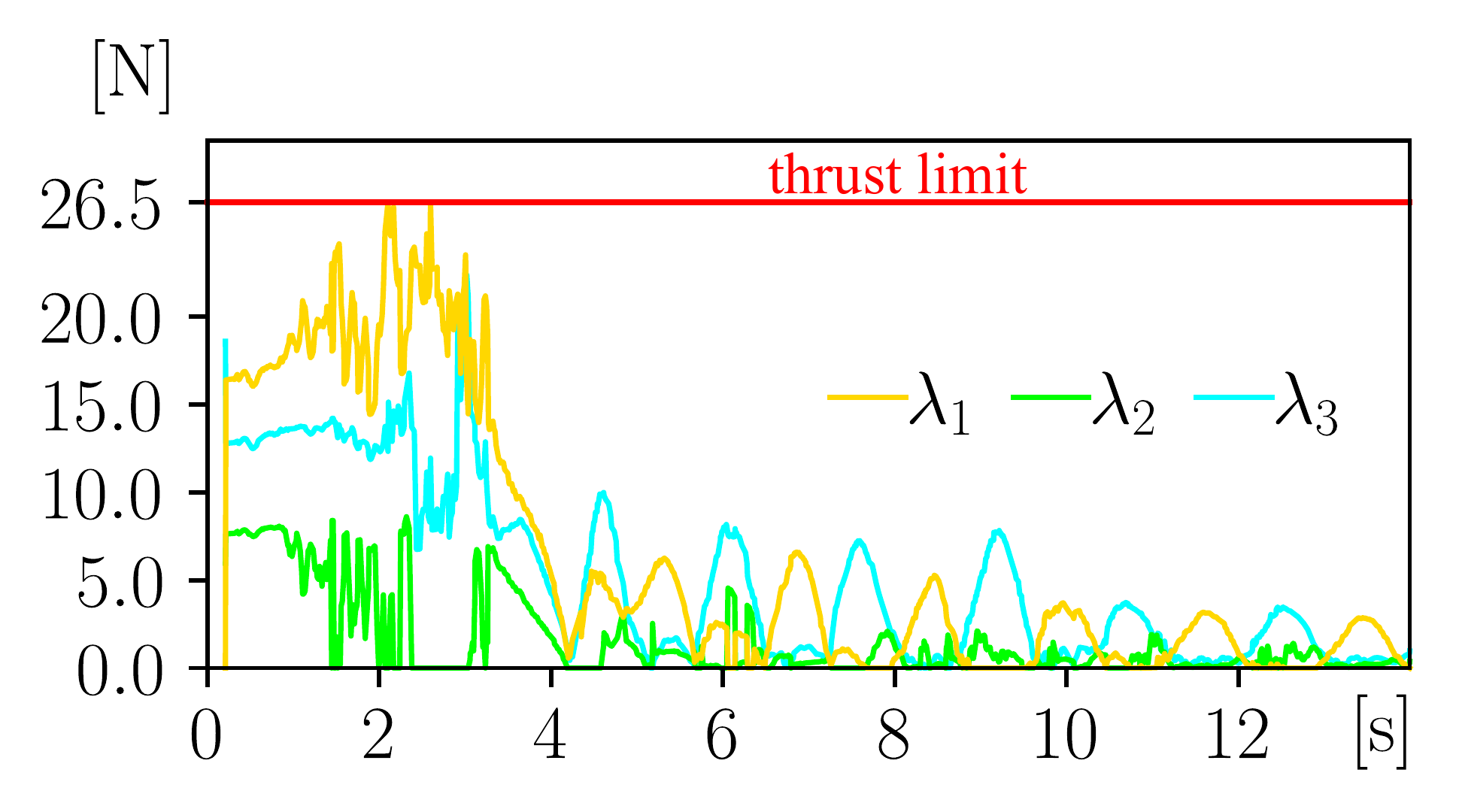}
            \subcaption{Target thrust\revise{s}}
            \label{figure:standing_plot:pwm}
        \end{minipage}%
    \end{tabular}
    \caption{Plots related to \figref{figure:standing_experiment}.}
    \label{figure:standing_plot}
    \vspace{-5mm}
\end{figure}

\begin{figure}[t]
  \centering
  \includegraphics[width=1.0\columnwidth]{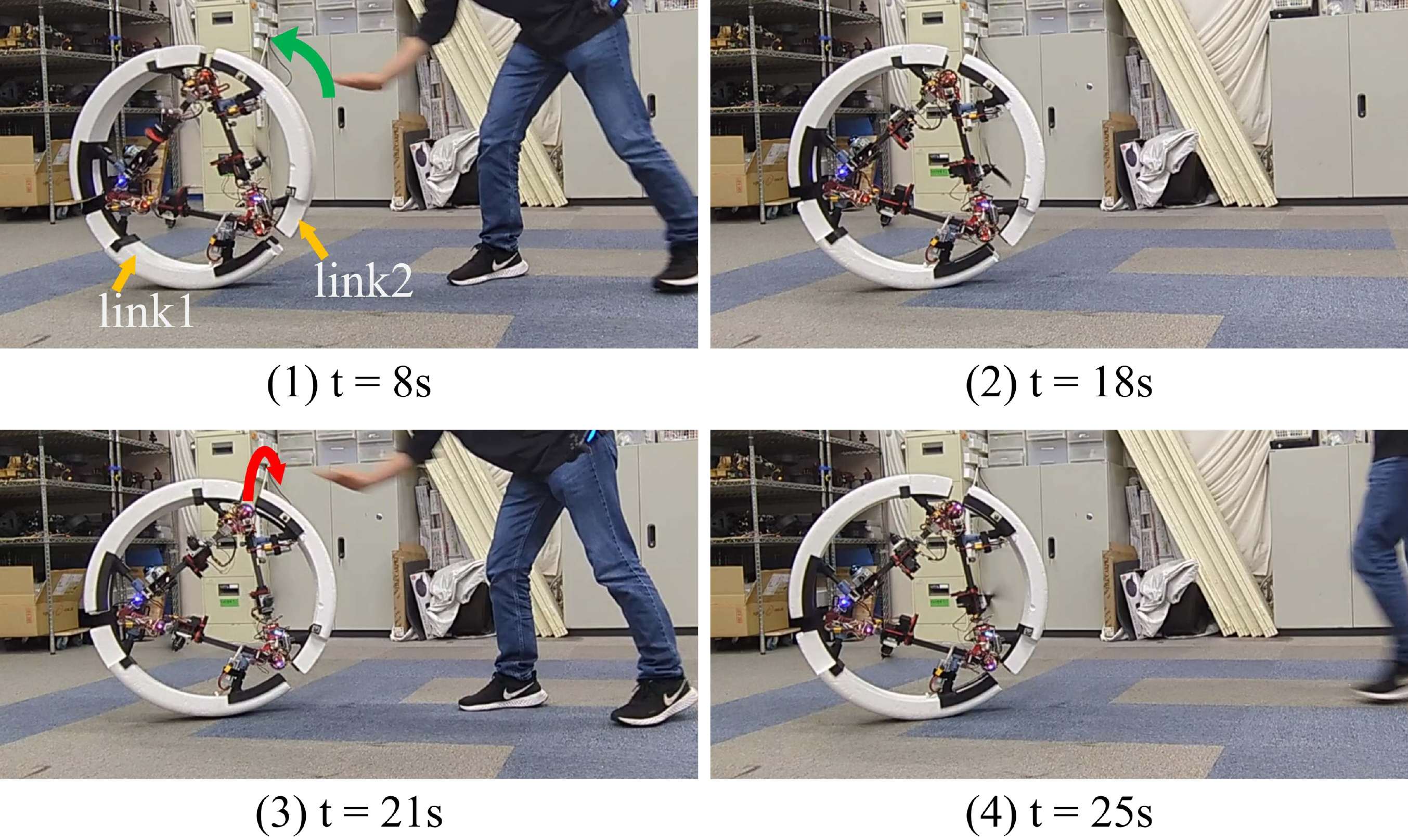}
    \caption{Ground balance experiment.}
    \label{figure:ground_balance_experiment}
    \vspace{-5mm}
\end{figure}

\begin{figure}[t]
    \begin{tabular}{ll}
        \begin{minipage}[t]{0.5\columnwidth}
            \centering
            \includegraphics[width=1.0\columnwidth]{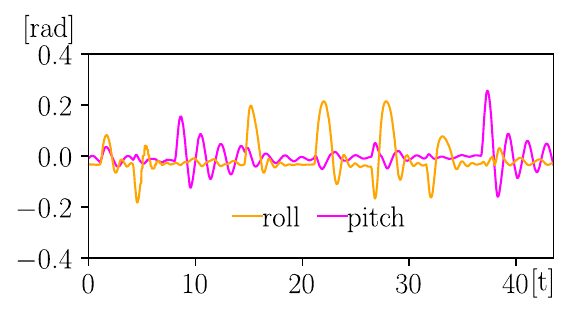}
            \subcaption{Orientation errors}
        \end{minipage}%
        \begin{minipage}[t]{0.5\columnwidth}
            \centering
            \includegraphics[width=1.0\columnwidth]{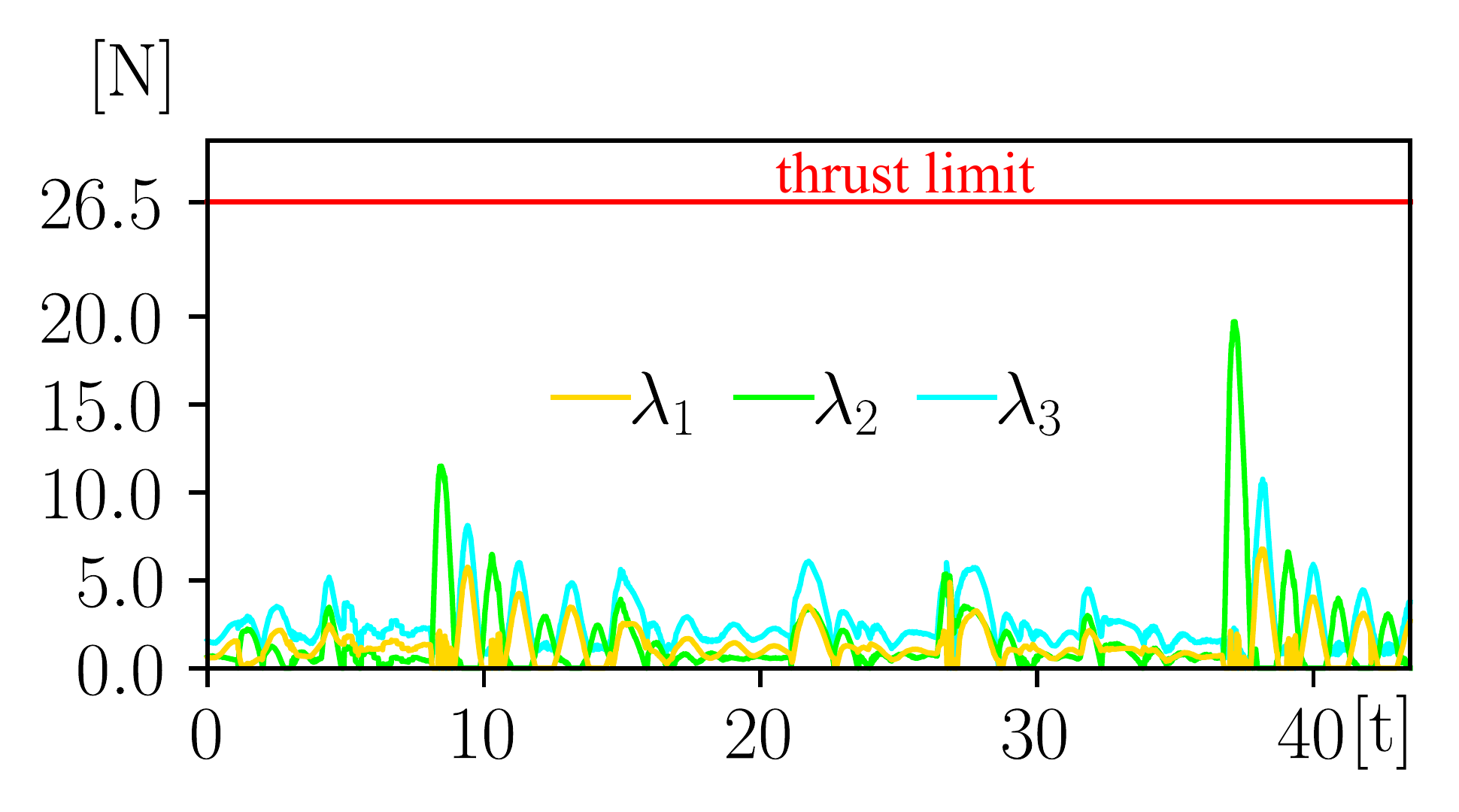}
            \subcaption{Target thrust\revise{s}}
            \label{figure:ground_balance_pwm}
        \end{minipage}%
        \\
        \begin{minipage}[t]{0.5\columnwidth}
            \centering
            \includegraphics[width=1.0\columnwidth]{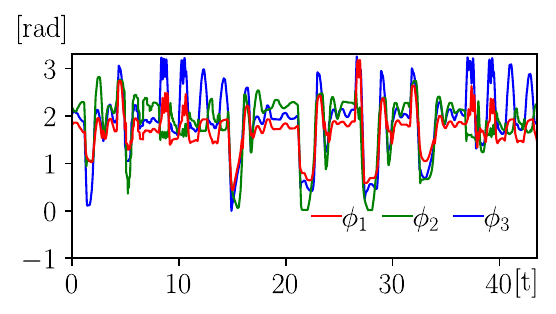}
            \subcaption{Thrust vectoring angles}
        \end{minipage}%
        \begin{minipage}[t]{0.5\columnwidth}
            \centering
            \includegraphics[width=1.0\columnwidth]{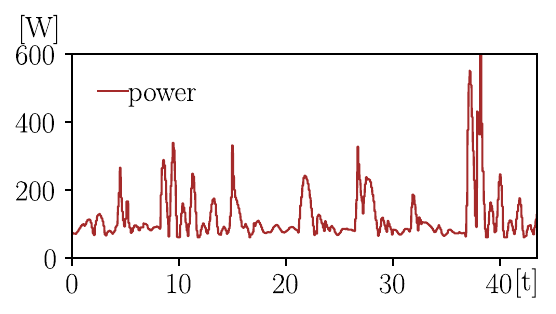}
            \subcaption{\revise{Power consumption}}
            \label{figure:ground_balance_power}
        \end{minipage}%
    \end{tabular}
    \caption{Plots related to \figref{figure:ground_balance_experiment}.}
    \label{figure:ground_balance_plot}
    \vspace{-5mm}
\end{figure}

\subsubsection{Standing-up Experiment}
\label{section:standing_experiment}
\switchlanguage{
A standing up motion \revise{experiment} on the ground \revise{to switch from aerial mode to terrestrial mode} was conducted.
The experiment is shown in \figref{figure:standing_experiment}.
The threshold of \revise{baselink} roll angle for \revise{the} state transition \(\phi_{\alpha}\) described in \subsecref{section:state transition} was set to \SI{0.2}{rad}.
The minimum jerk attitude trajectory in \equref{eq:baselink_trajectory} and \equref{eq:minjerk_trajectory} was calculated with \(N=11\), \(t_1=5s\).
For verification, we also conducted an experiment using linear\revise{ly} interpolat\revise{ed} trajectory instead of minimum jerk trajectory with the same duration.
\revise{The} PID gains were set to \(K_{\tau,p}=diag\qty(6.5, 5.0, 6.0)\), \(K_{\tau,i}=diag\qty(0.1, 0.1, 0.0)\), \(K_{\tau,d}=diag\qty(5.0, 4.0, 3.0)\). 
The plots in \figref{figure:standing_plot:baselink_roll} and \figref{figure:standing_plot:baselink_roll_linear} show the target and actual \revise{baselink} roll angle when using minimum jerk trajectory, and when using linear interpolation trajectory\revise{, respectively}.
\revise{The} plots in \figref{figure:standing_plot:pitch_yaw_error}--\figref{figure:standing_plot:pwm} show the tracking errors of pitch and yaw angle angle of the CoG frame, \revise{the} thrust vectoring angles, and \revise{the target thrust of} each thruster, when we use minimum jerk trajectory.

For the first few seconds, the robot exerted approximately \SI{25}{N} thrust to follow the target roll angle and \revise{stood} up.
As shown in \figref{figure:standing_plot:pwm}, the \revise{thrust} was larger in the thruster \revise{on links} 1 and 3.
This is because they can contribute more to the rotation of the body due to the large distance from the ground contact point.
As shown in \figref{figure:standing_plot:baselink_roll}, there was a delay of about \SI{1}{s} in following the roll angle during standing up.
Especially around \SI{2}{s}, this time delay is increased because the outer frame started to contact the ground, and the external force prevented the rotation.
\revise{To investigate the effect of the reference attitude trajectory, we compared the minimum-jerk trajectory with a linearly interpolated attitude trajectory.
As shown in \figref{figure:standing_plot:baselink_roll_linear}, the trajectory based on linear interpolation exhibited a larger tracking delay, with a maximum delay of about \SI{3}{s} and a remaining delay of about \SI{1}{s} at the end of the motion.
These delay occur because no explicit feedforward terms for the desired angular velocity and angular acceleration are provided.
These results suggest that, for relatively large and high-inertia multirotor systems such as our prototype, generating smooth reference trajectories is crucial when executing motions that involve large attitude changes.
}

After standing up, the robot showed an overshoot of about \SI{0.25}{rad}, but it responded by controlling the thrust vectoring angles and finally stabilized in the vertically standing state.
In this process, the yaw angle showed an error of about \SI{0.25}{rad} at most.
\revise{The} target thrust vectoring angles was commanded, but there was a time delay in following the actual state, and the difference between the target and actual state affected around another axis.
The pitch angle also oscillated with an amplitude of about \SI{0.1}{rad}, but \revise{it was finally} converged.
In addition, the state transition described in \secref{section:state transition} worked.
During the standing up, the thrust vectoring angle\revise{s} became negative to satisfy the constraint of roll angle control and friction cone constraint, but after reaching the vertical, the thrust vectoring angle\revise{s} \revise{were} controlled to satisfy \equref{chap4::eq::rolling_gimbal_range_constraint}.
}{
地上における移動形態の変化に必要な, 機体を起き上がらせる動作実験を行った.
実験の様子を\figref{figure:standing_experiment}に示す.
\subsecref{section:state transition}で述べた, state transitionのためのベースリンクのroll角のthreshold\(\phi_{\alpha}\)は\SI{0.2}{rad}とした。
\subsecref{sec::minimum_jerk}の姿勢遷移軌道を\(N=11\), \(t_1=5s\)で計算を行った.
また, 検証のため, minimum jerk軌道の代わりに, 姿勢だけに対して線形補間の軌道を用いた場合の実験も行った.
PIDゲインは\(K_{\tau,p}=diag\qty(6.5, 5.0, 6.0)\), \(K_{\tau,i}=diag\qty(0.1, 0.1, 0.0)\), \(K_{\tau,d}=diag\qty(5.0, 4.0, 3.0)\)とした. 
Plots in \figref{figure:standing_plot:baselink_roll} and \figref{figure:standing_plot:baselink_roll_linear} show the target and actual roll angle of baselink when using minimum jerk trajectory, and when using linear interpolation trajectory.
Plots in \figref{figure:standing_plot:pitch_yaw_error}--\figref{figure:standing_plot:pwm} show the tracking errors of pitch and yaw angle angle of the CoG frame, thrust vectoring angles, and input PWM commands to each thrusters.

最初の数秒間, 最大でおよそ\SI{25}{N}の推力を発揮することで, 機体が目標のroll角に追従し, 立ち上がることができた.
このとき, \figref{figure:standing_plot:pwm}に示すように, 発揮推力はlink1とlink3の推力装置が大きくなった.
これは接地点に対して距離が大きく, 機体の回転に多く寄与できるためであると考えられる. 
\figref{figure:standing_plot:baselink_roll}に示すように, 立ち上がり中の姿勢の追従にはおよそ\SI{1}{s}程度の遅れが発生した.
特に\SI{2}{s}付近で, 追従が遅れており, 足のみで接地していた状態からフレームが接地し, 外力が回転を妨げたためであると考えられる.
minimum jerk軌道と線形補間軌道で比較をすると, 線形補間軌道では, とくにこの遅れが大きくなっていて, 最大\SI{3}{s}程度停滞し, 最終的に追従するのに\SI{1}{s}程度の遅れが発生した.
これは, フィードフォワード的な目標角加速度の設定と目標角速度の設定をしていないため, 角度制御の項が十分大きくなるまで十分な出力が出なかったためである.
さらに姿勢遷移に伴う角速度が逆にその運動を妨げる方向にフィードバックがかかったことが原因だと考えられる.
また, 起き上がった後, 約\SI{0.25}{rad}程度のオーバーシュートが起きたが, それに対し, 推力方向角を制御することによって応答し,最終的に垂直近傍の姿勢に安定化することができた.

また, この過程で, yaw角が最大で\SI{0.25}{rad}程度の誤差を示した.
目標の推力方向角が司令されるが, 実際に追従するまでには時間遅れが存在していて, 目標と実際の状態の違いにより別の軸に影響が出たのだと考えられる.
pitch角も\SI{0.1}{rad}程度の振幅で振動していたが, 収束していった.
また, \secref{section:state transition}で述べたstate transitionが機能しており, 起き上がりの最中は大きなroll角のモーメントと滑りを防ぐ制約を満たすために推力方向角が負になることがあったが, 
垂直近傍になったあとは\equref{chap4::eq::rolling_gimbal_range_constraint}を満たすように推力方向角が制御された.
}

\begin{figure}[t]
    \centering
    \includegraphics[width=1.0\columnwidth]{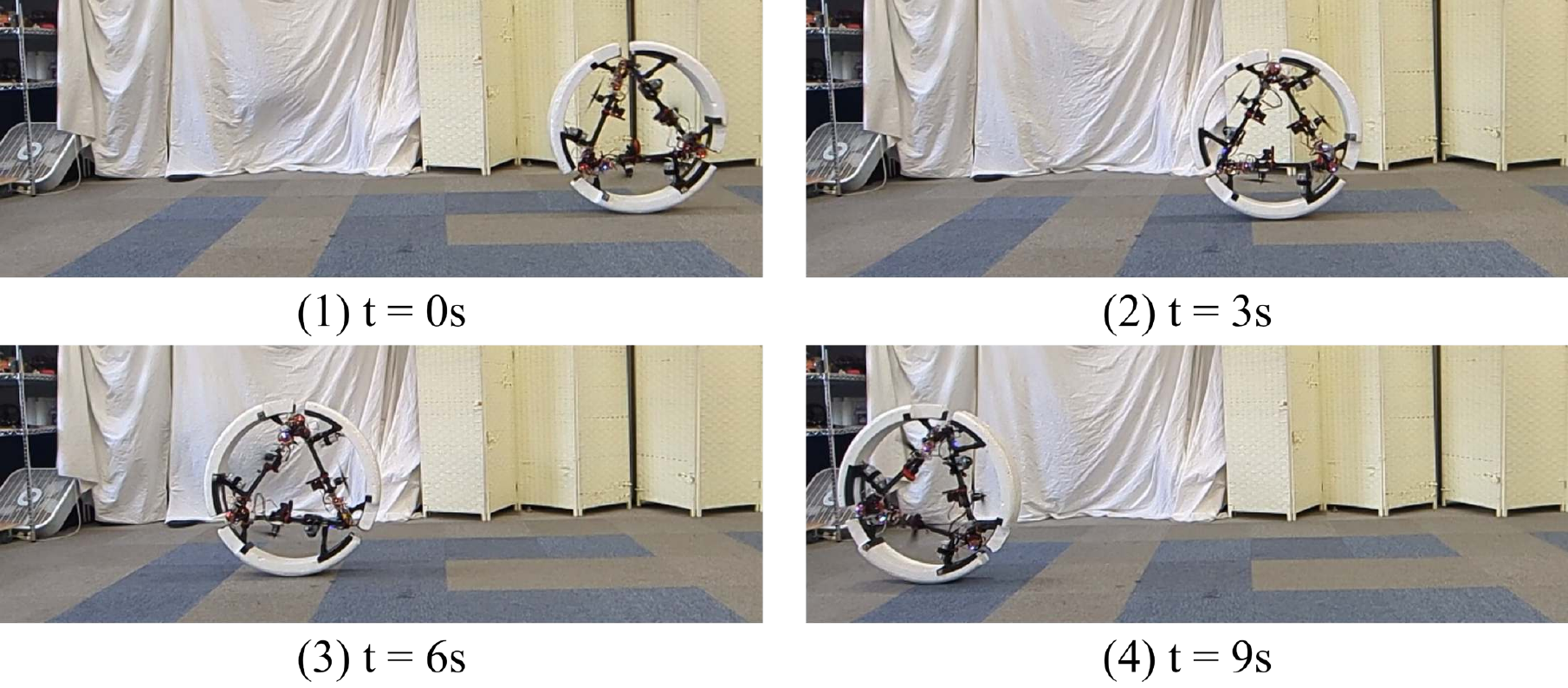}
    \caption{Rolling experiment.}
    \vspace{-5mm}
    \label{figure:rolling experiment}
\end{figure}

\begin{figure}[t]
    \begin{tabular}{cc}
        \begin{minipage}[t]{0.5\columnwidth}
            \centering
            \includegraphics[width=1.0\columnwidth]{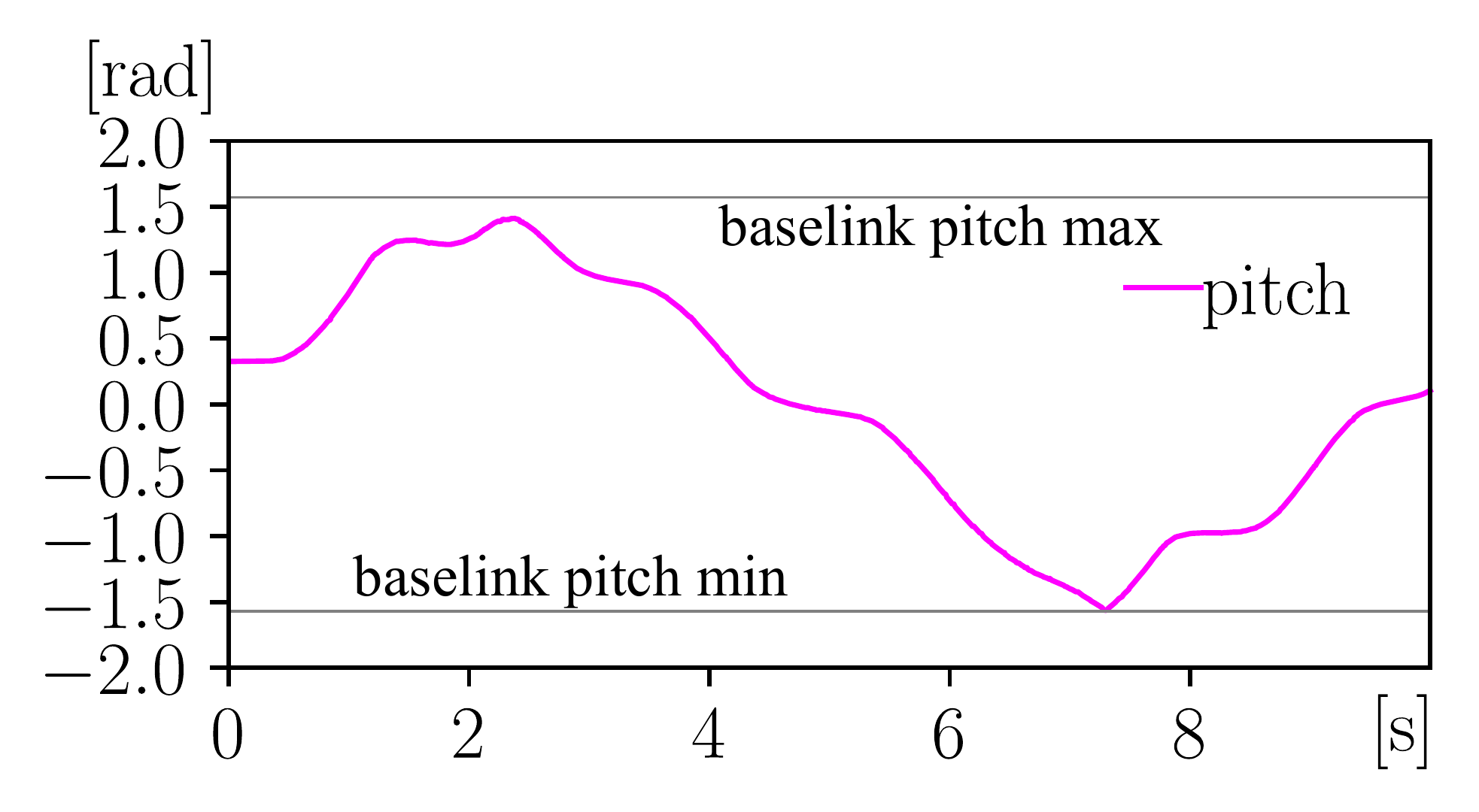}
            \subcaption{Pitch angle of baselink}
        \end{minipage}%
        \begin{minipage}[t]{0.5\columnwidth}
            \centering
            \includegraphics[width=1.0\columnwidth]{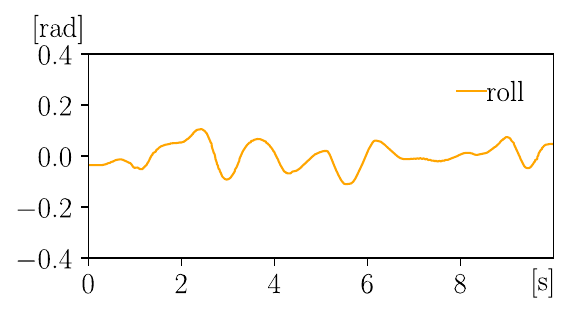}
            \subcaption{Roll angle tracking error}
        \end{minipage}%
        \\
        \begin{minipage}[t]{0.5\columnwidth}
            \centering
            \includegraphics[width=1.0\columnwidth]{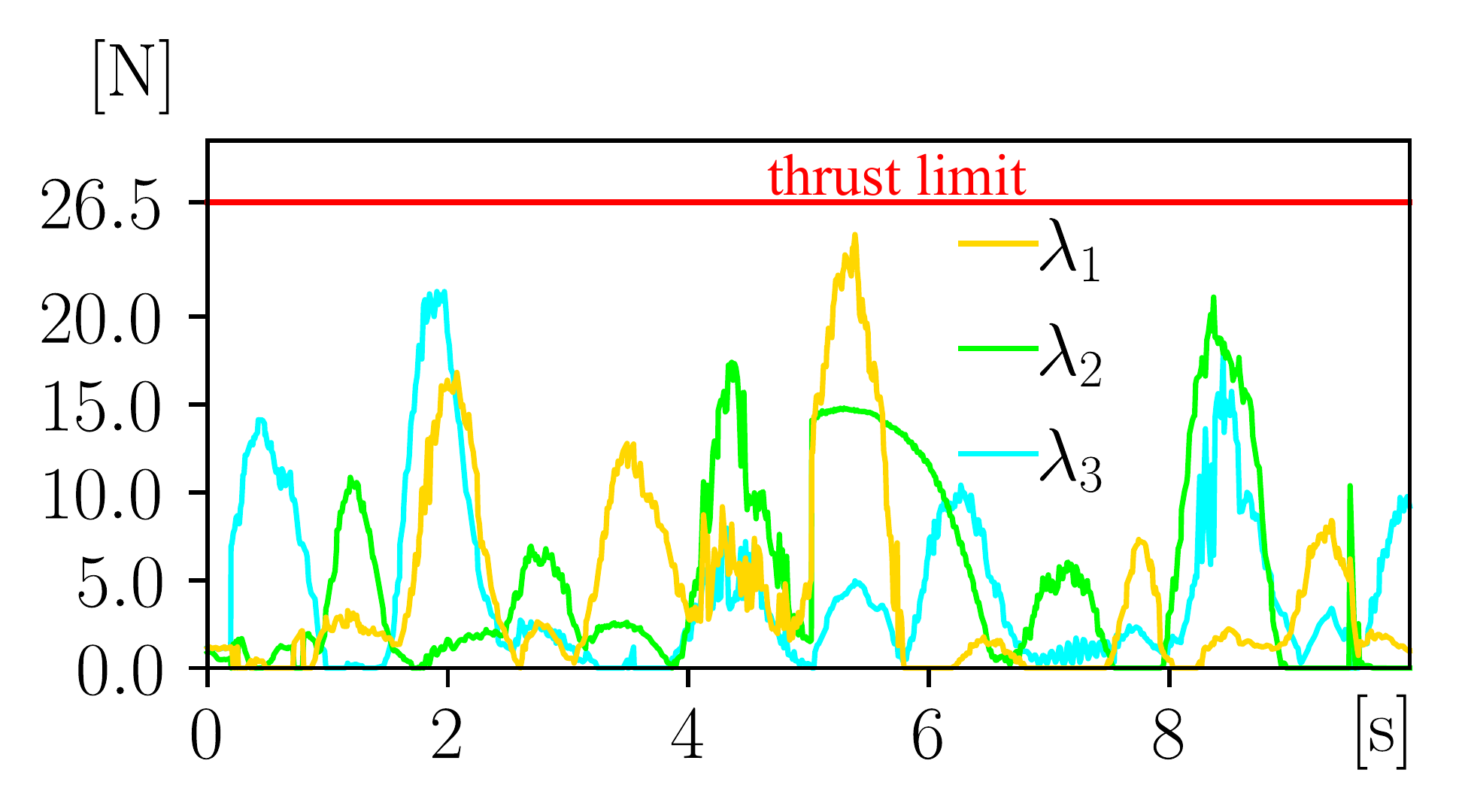}
            \subcaption{Target thrust}
            \label{figure:rolling_plot:pwm}
        \end{minipage}%
        \begin{minipage}[t]{0.5\columnwidth}
            \centering
            \includegraphics[width=1.0\columnwidth]{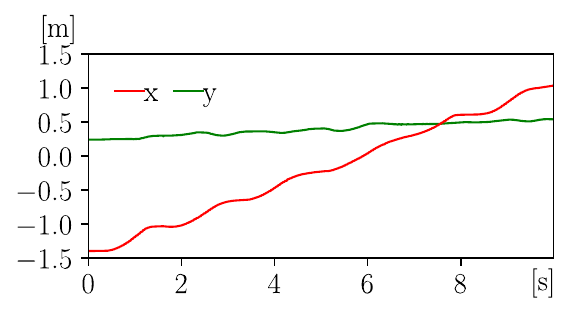}
            \subcaption{Position of CoG in world frame}
        \end{minipage}%
    \end{tabular}
    \caption{Plots related to \figref{figure:rolling experiment}.}
    \label{figure:rolling plot}
    \vspace{-5mm}
\end{figure}

\subsubsection{Stability against Disturbance}
\label{section:disturbance_experiment}
\switchlanguage{
To verify the stability against external forces during motion on the ground, a balancing experiment with disturbance\revise{s,} as shown in \figref{figure:ground_balance_experiment}, was conducted.
Disturbances were manually applied \revise{about} the roll and pitch angles \revise{of \(\{CoG\}\)}.
\revise{The p}lots in \figref{figure:ground_balance_plot} show the tracking errors of roll and pitch angle\revise{s}, the target thrust of each thruster, thrust vectoring angles, and power consumption.

Even when disturbances were applied, the robot \revise{could be} stabilized by \revise{the} control\revise{ler}.
Comparing the tracking errors of roll and pitch angles with the changes in thrust vectoring angles, it can be seen that when the tracking error of the roll angle is large, the thrust vectoring angle changes significantly.
At this time, the thrust of second and third links increased.
This is because the distance from the contact point is large, and the moment per unit thrust generated around the contact point is large.
Also, when the tracking error of the pitch angle is large, attitude control was mainly performed by thrust as shown in \figref{figure:ground_balance_pwm}.
\revise{As shown in \figref{figure:ground_balance_power}, when the attitude was sufficiently following the target, the power consumption was kept to about \SI{100}{W}, while when a large disturbance was applied, it increased to about \SI{300}{W}.}
}{
地上領域における動作時の外力に対する安定性を検証するため, \figref{figure:ground_balance_experiment}に示す, 外乱を与えるバランス実験を行った.
roll角とpitch角周りに手動で外乱を与えた.
Plots in \figref{figure:ground_balance_plot} show the tracking errors of roll and pitch angle, the input command to each thruster, and thrust vectoring angles.

外乱が加えられた際にも, 制御により安定化させることができた.
roll角, pitch角の外乱と推力方向角の変化を比較すると, roll角の追従誤差が大きいときに特に推力方向角が大きく変化していることがわかる.
またこのとき, link2とlink3のスラストが大きくなった.
これは, 接地点からの距離が大きく, 接地周りに発生させられる単位推力辺りのモーメントが大きいためであると考えられる
また, pitch角の追従誤差が大きいときには主に推力により, 姿勢制御が行われた.
姿勢が目標に十分追従しているときには消費電力が\SI{100}{W}程度に抑えられていたのに対し, 大きな外乱が加えられた際には\SI{300}{W}程度まで増加した.
}

\begin{figure}[t]
    \centering
    \includegraphics[width=1.0\columnwidth]{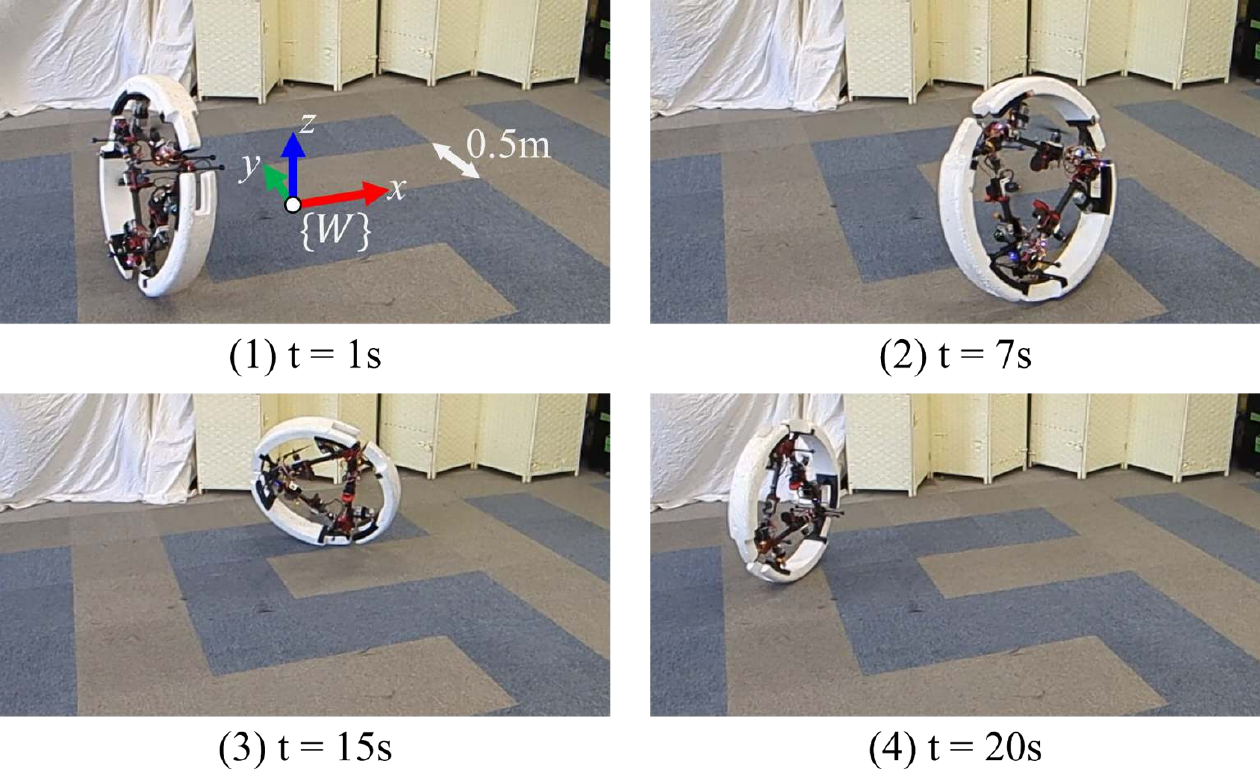}
    \caption{Circle trajectory tracking rolling experiment.}
    \label{figure:circle_rolling_experiment}
    \vspace{-5mm}
\end{figure}

\begin{figure}[t]
    \begin{tabular}{cc}
        \begin{minipage}[t]{0.5\columnwidth}
            \centering
            \includegraphics[width=1.0\columnwidth]{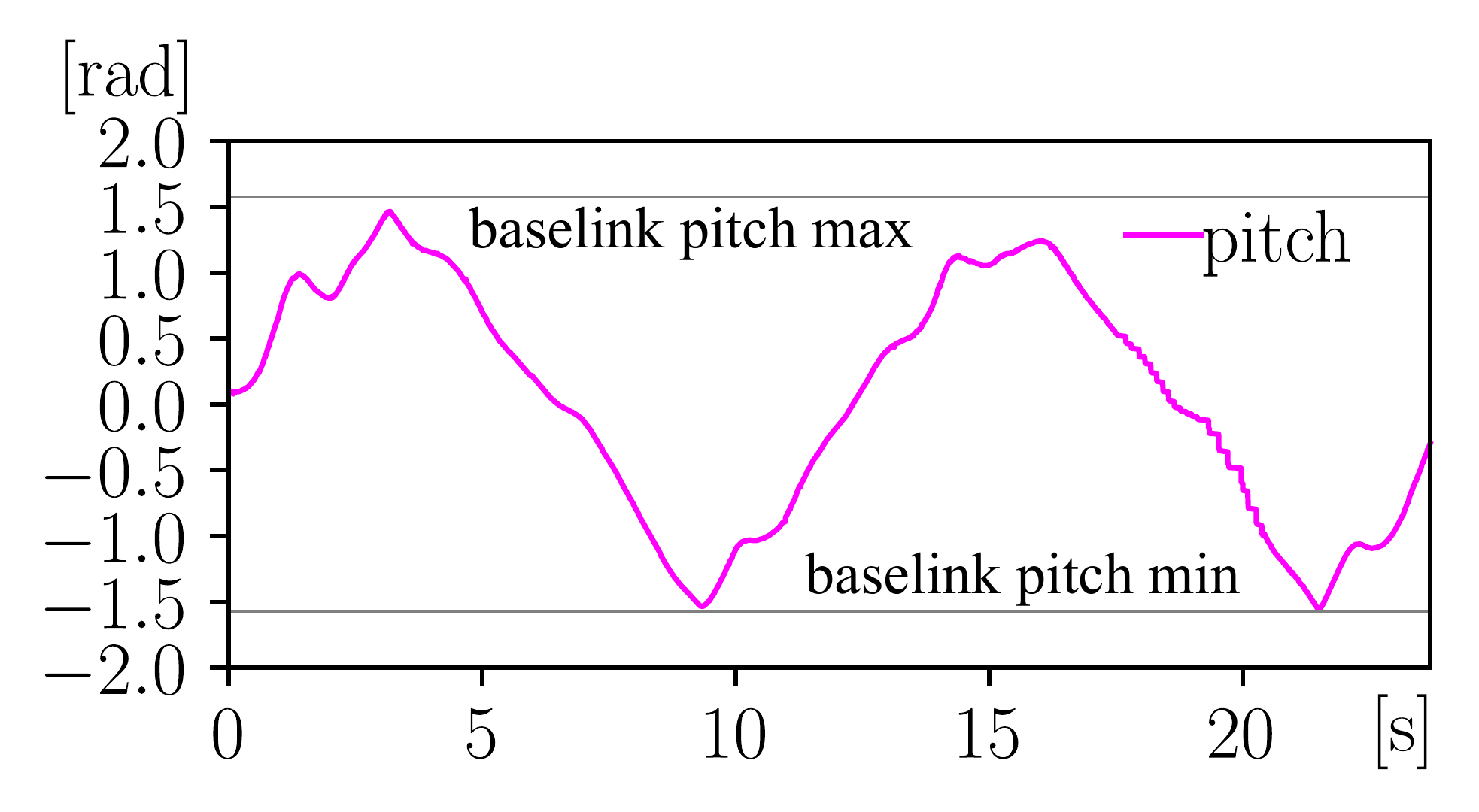}
            \subcaption{Pitch angle of baselink}
        \end{minipage}%
        \begin{minipage}[t]{0.5\columnwidth}
            \centering
            \includegraphics[width=1.0\columnwidth]{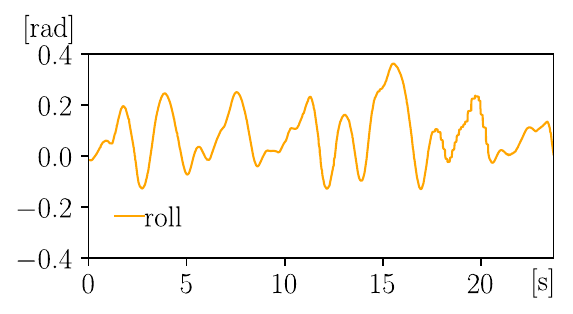}
            \subcaption{Roll angle tracking error}
        \end{minipage}%
        \\
        \begin{minipage}[t]{0.5\columnwidth}
            \centering
            \includegraphics[width=1.0\columnwidth]{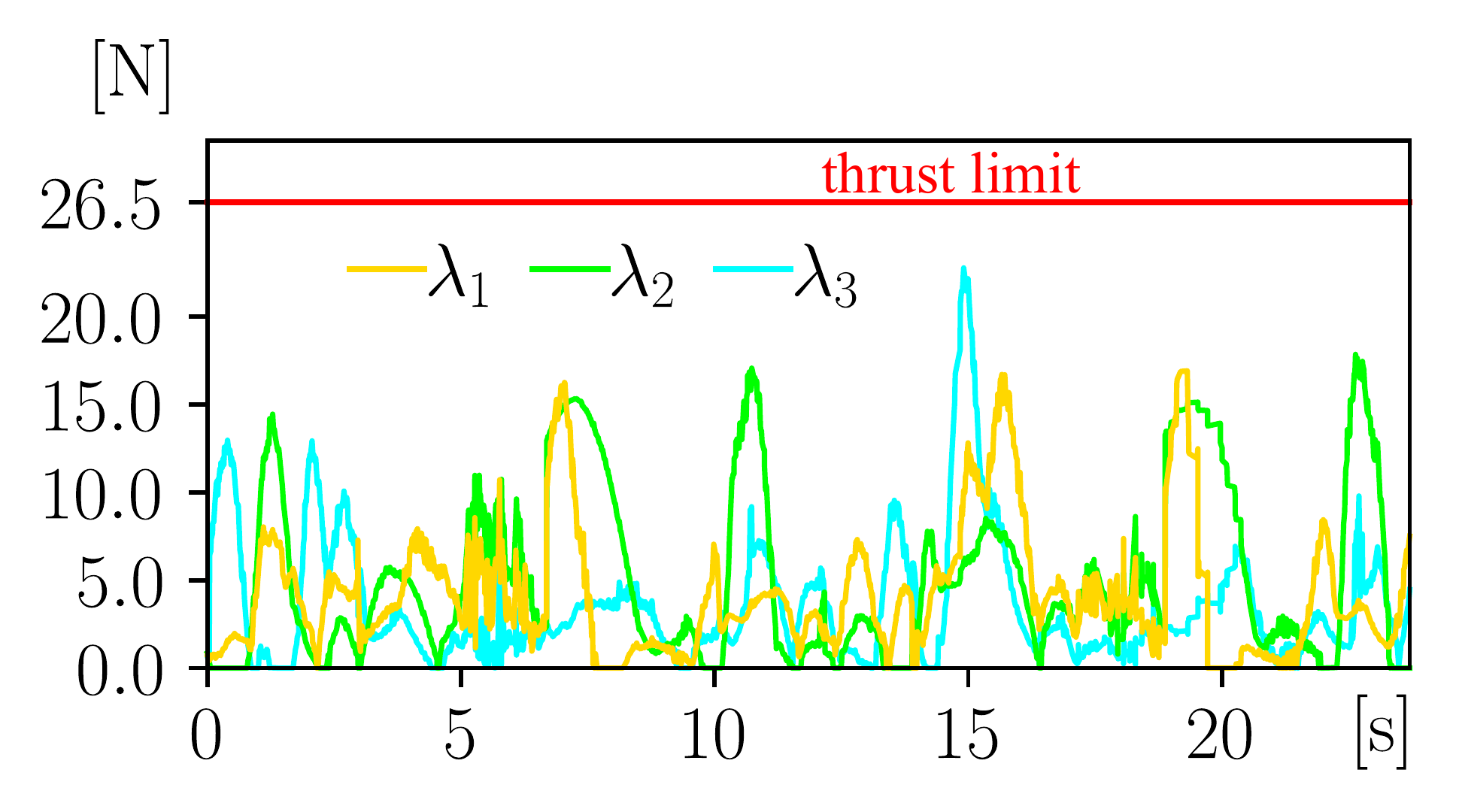}
            \subcaption{Target thrust}
            \label{figure:circle_rolling_plot:pwm}
        \end{minipage}%
        \begin{minipage}[t]{0.5\columnwidth}
            \centering
            \includegraphics[width=1.0\columnwidth]{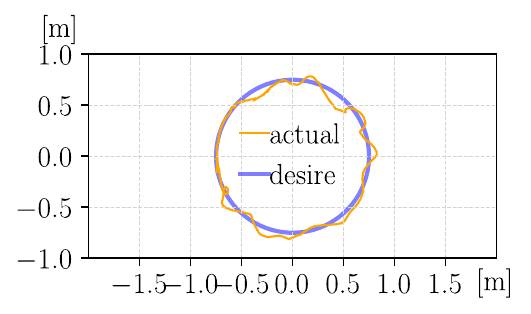}
            \subcaption{Desire and actual \(xy\) trajectory of CoG}
        \end{minipage}%
    \end{tabular}
    \caption{Plots related to \figref{figure:circle_rolling_experiment}.}
    \label{figure:circle_rolling_plot}
    \vspace{-5mm}
\end{figure}

\subsubsection{Rolling Experiment}
\label{section:rolling_experiment}
\switchlanguage{
Rolling locomotion experiment\revise{s} as shown in \figref{figure:rolling experiment} and \figref{figure:circle_rolling_experiment} \revise{were} conducted.
In \figref{figure:rolling experiment}, a target angular velocity around \revise{the} \(y\)-axis \revise{of the CoG} (rolling \revise{axis}) was \revise{commanded} to make the robot roll along a straight trajectory.
In \figref{figure:circle_rolling_experiment}, a target angular velocity around \(z\)-axis \revise{of the world frame} (steering \revise{axis}) was also sent to make the robot roll along a CCW circular trajectory with a radius of \SI{0.75}{m}.
Note that no position feedback was used in the control \revise{so that we could} evaluate the control\revise{ler} \revise{under} the friction cone constraint\revise{s}.
\revise{The} plots in \figref{figure:rolling plot} \revise{show the pitch angle of baselink, roll angle tracking error, target thrust of each thruster, and the position of the CoG in the world frame}.
\revise{The plots in \figref{figure:circle_rolling_plot}(1)--(3) show the same physical quantities in as in \figref{figure:rolling plot}(1)--(3) and \figref{figure:circle_rolling_plot}(4) show the desired and actual CoG trajectory.
Please note that the center of the desired circle trajectory is plotted as \(\qty(0, 0)\) in \figref{figure:circle_rolling_plot}(4).}
The pitch angle of the baselink is clamped to the range of \(\qty[-\frac{\pi}{2}, \frac{\pi}{2}]\).

In the rolling motion along the straight trajectory shown in \figref{figure:rolling experiment}, the robot was able to roll at \SI{0.25}{m/s}.
The tracking error of the roll angle was about \SI{0.1}{rad}.
Also, in the rolling motion along the circular trajectory, this tracking error was about \SI{0.4}{rad} at most, and the robot tended to fall in the direction of the inner side of the circular trajectory.
As a result, the error with the target trajectory was at most \SI{0.2}{m}.
This \revise{behavior is attributed} to \revise{the fact} that, while changing the thrust vectoring angle\revise{s} to \revise{generate} the yaw angular acceleration,  the rotational motion around \revise{other axes is affected} until it converges to \revise{the} target angle.
As shown in \figref{figure:circle_rolling_plot}, the exerted thrust was larger in the direction that easily contributes to the angular momentum change for rolling locomotion, and the output of each thruster changed periodically.
The power consumption \revise{of thrust motors} during \revise{the circle trajectory} rolling locomotion experiment was about \revise{\SI{284}{W}}.
\revise{Considering the duration and distance of the experiment, the energy for terrestrial locomotion per unit distance was approximately \SI{1428}{J/m}.}
}{
\figref{figure:rolling experiment}と\figref{figure:circle_rolling_experiment}に示す, 地上での転がり移動動作実験を行った.
\figref{figure:rolling experiment}では, pitch角の角速度(rolling angular vel)を送り, 直線軌道に沿って転がるようにした.
\figref{figure:circle_rolling_experiment}では, 同時にyaw角の目標角速度(streering velocity)も送り, 半径\SI{0.75}{m}の円軌道に沿って転がるようにした.
摩擦錐制約を考慮した制御の評価のため, どちらも位置によるフィードバックは行わず, 目標角速度のみを司令していることに注意されたい.
\figref{figure:rolling plot}と\figref{figure:circle_rolling_plot}のプロットはそれぞれベースリンクのpitch角, roll角の追従誤差, 各スラスタへの入力PWM, 重心の世界座標系における位置を示している.
なお, ベースリンクのpitch角は\(\qty[-\frac{\pi}{2}, \frac{\pi}{2}]\)の範囲にクランプされている.

\figref{figure:rolling experiment}に示す直線軌道に沿った転がり動作において, \SI{0.25}{m/s}での転がり動作が実現できた。
またこのときのroll角の追従誤差は\SI{0.1}{rad}程度となった.
また, 円軌道に沿った動作時には最大で\SI{0.4}{rad}程度となり, 円軌道の内側に倒れる方向への誤差が大きくなった.
これに起因して, 目標軌道との誤差が最大で\red{\SI{0}{m}}程度発生した.
yaw角周りの加速度を発揮するために目標の推力方向角を変化させている間は別の軸の回転運動に影響を及ぼすためであると考えられる.
hogeに示すように, 発揮推力は転がり移動のための角運動量変化に寄与しやすい方向を向いた推力装置出力が大きくなり, 各推力装置の出力が周期的に変化する結果となった.
また, 転がり移動時の消費電力は約\SI{300}{W}であり, 飛行時よりも低い値となった.
}

\begin{figure}[t]
    \centering
    \includegraphics[width=1.0\columnwidth]{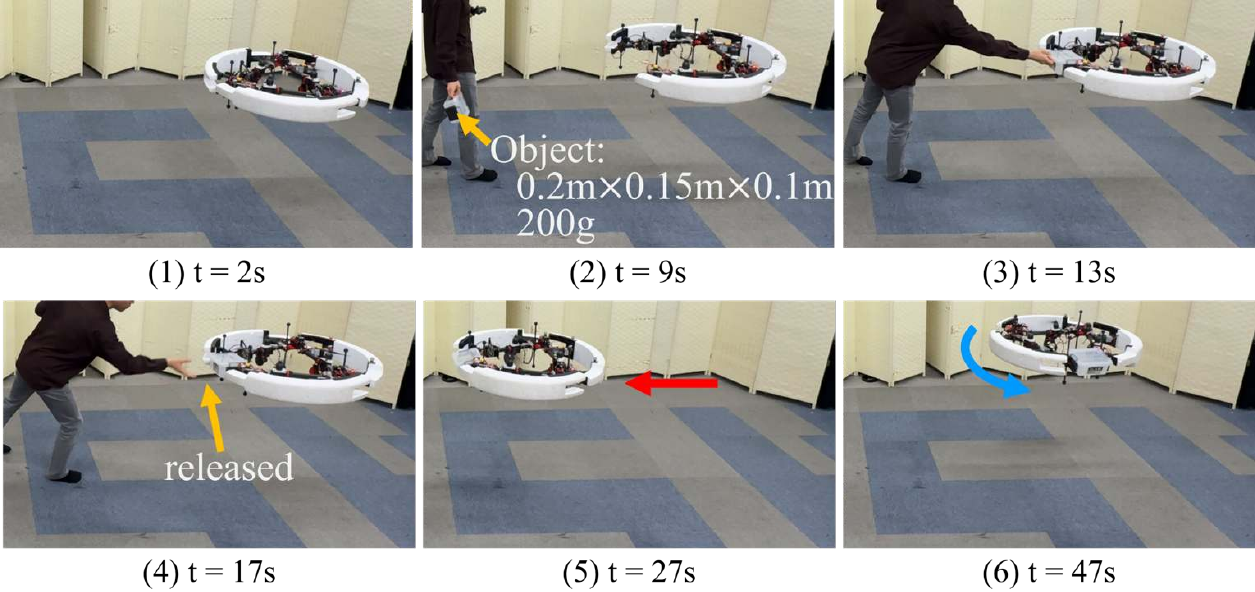}
    \caption{Object grasping flight experiment.}
    \label{figure:object_grasping_experiment}
    \vspace{-5mm}
\end{figure}

\begin{figure}[t]
    \centering
    \begin{tabular}{cc}
        \begin{minipage}[t]{0.5\columnwidth}
            \centering
            \includegraphics[width=1.0\columnwidth]{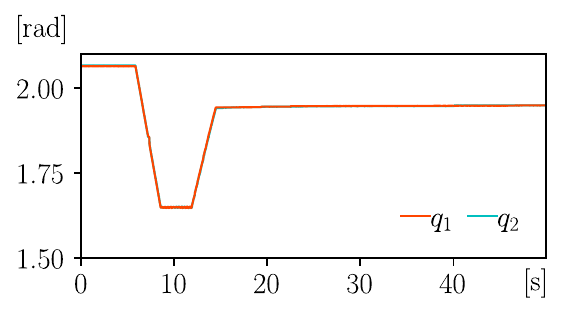}
            \subcaption{Joint angles}
        \end{minipage}%
        \begin{minipage}[t]{0.5\columnwidth}
            \centering
            \includegraphics[width=1.0\columnwidth]{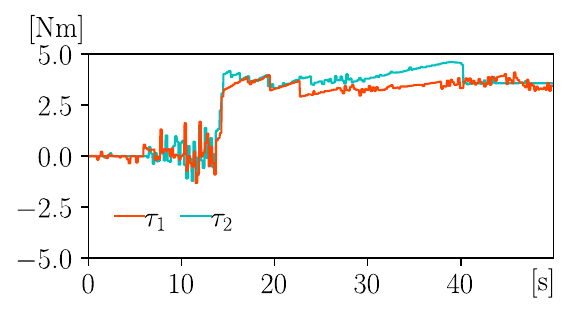}
            \subcaption{Joint torques}
        \end{minipage}%
        \\
        \begin{minipage}[t]{0.5\columnwidth}
            \centering
            \includegraphics[width=1.0\columnwidth]{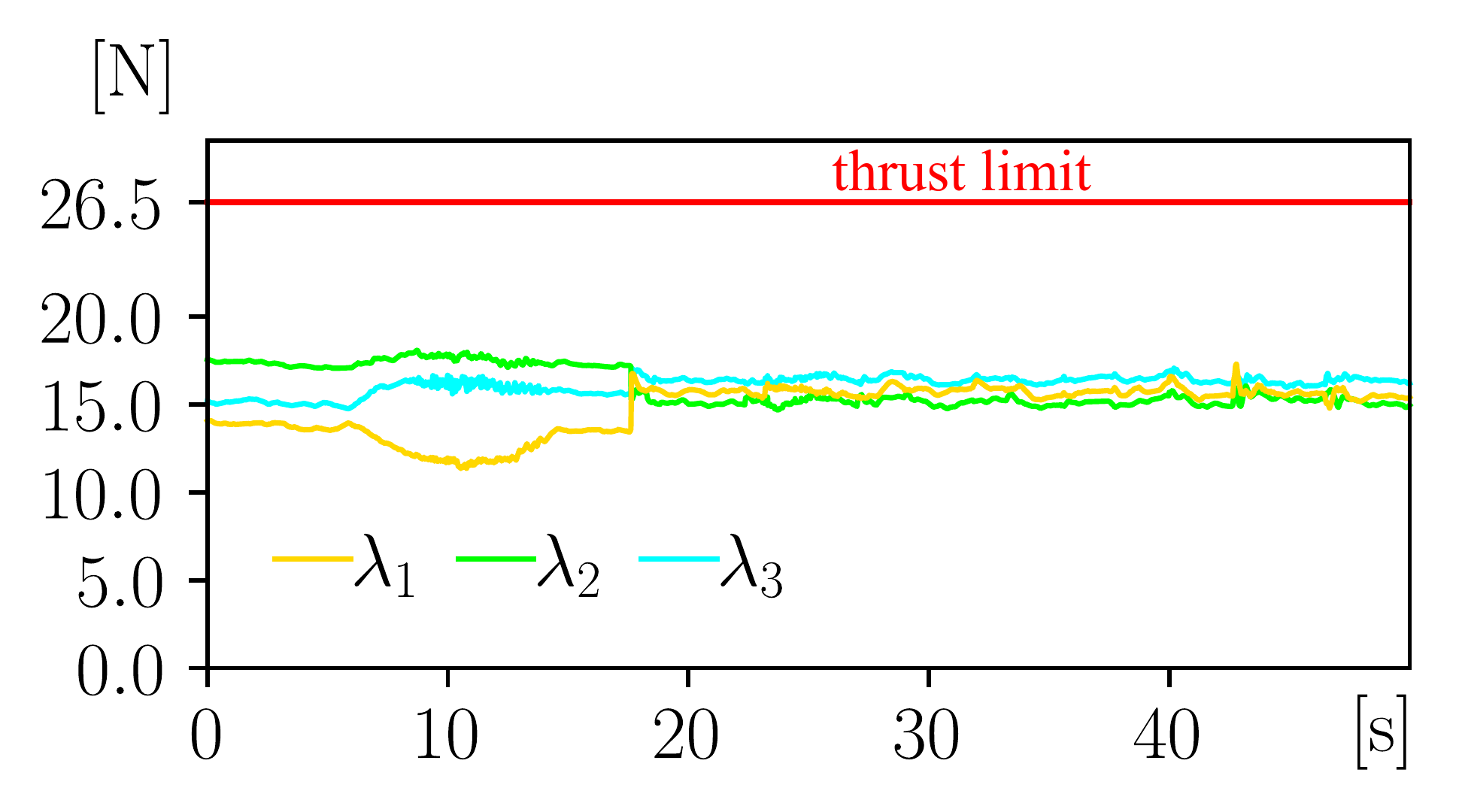}
            \subcaption{Target thrusts}
        \end{minipage}%
        \begin{minipage}[t]{0.5\columnwidth}
            \centering
            \includegraphics[width=1.0\columnwidth]{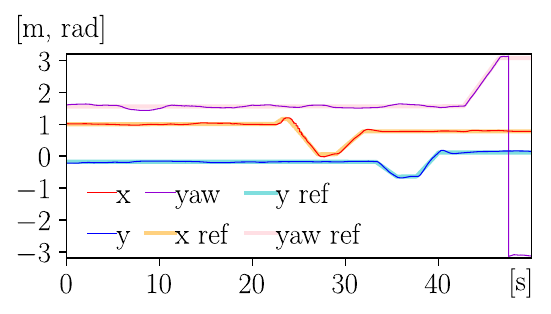}
            \subcaption{Target and actual value of CoG \(xy\) position and yaw angle}
            \label{figure:object_grasping_plot:cog_ref_curr}
        \end{minipage}%
    \end{tabular}
    \caption{Plots related to \figref{figure:object_grasping_experiment}.}
    \label{figure:object_grasping_plot}
    \vspace{-5mm}
\end{figure}

\switchlanguage{
\revise{From the viewpoint of energy consumption, terrestrial locomotion is 3.8 times more efficient than aerial locomotion.
In \cite{kalantari2013hytaq,cao2023doublebee} and \cite{zhao2023spidar}, this ratios were 35, 4.9, and 2.2, respectively, based on the reported values or estimation from the experimental results.
This ratio tends to decrease as the system becomes more complex, and the value obtained in this work lies between those reported in \cite{cao2023doublebee} and \cite{zhao2023spidar}.}
}{
消費電力平均: 284W
時間: 23.7s
距離: 2 * 0.75 * pi = 4.712m
その他のプラットフォームとの比較も行う.
\cite{kalantari2013hytaq,cao2023doublebee,zhao2023spidar}では, 報告された実験結果から推定すると, 地上ロコモーションによる航続時間の延長倍率はそれぞれ, 35, 4.9, 2.2であった.
これらに対し, 本研究では3.8倍となった.
システムが複雑化していくほどこの倍率は下がっていく傾向にあると考えられ, \cite{cao2023doublebee}と\cite{zhao2023spidar}の間の値になった.
hytaq: 論文4-Bで報告
double bee: Fig.5, 454/92
spidar: table1 20/9
}

\begin{figure}[t]
    \centering
    \includegraphics[width=1.0\linewidth]{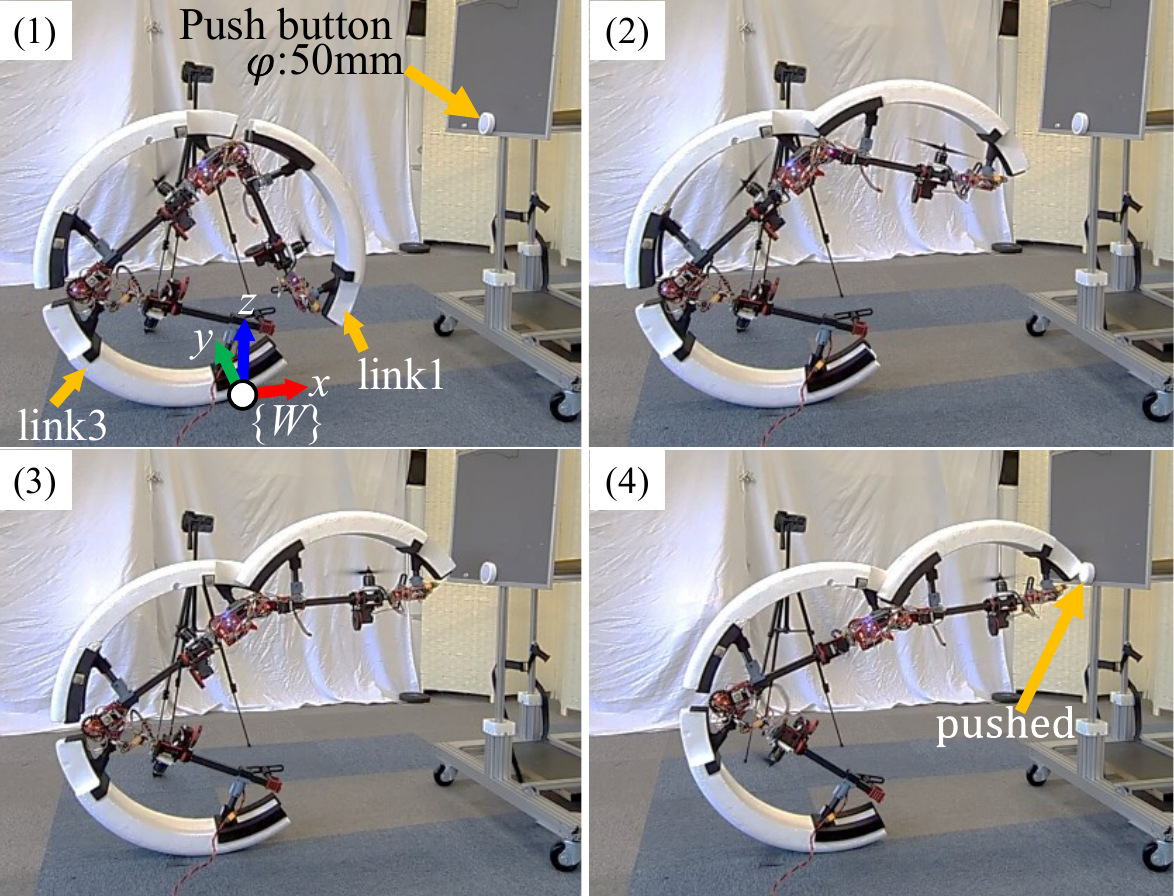}
    \caption{Button push experiment}
    \label{figure:button_push_experiment}
    \vspace{-5mm}
\end{figure}

\begin{figure}[t]
    \begin{tabular}{cc}
        \begin{minipage}[t]{0.5\columnwidth}
            \centering
            \includegraphics[width=1.0\columnwidth]{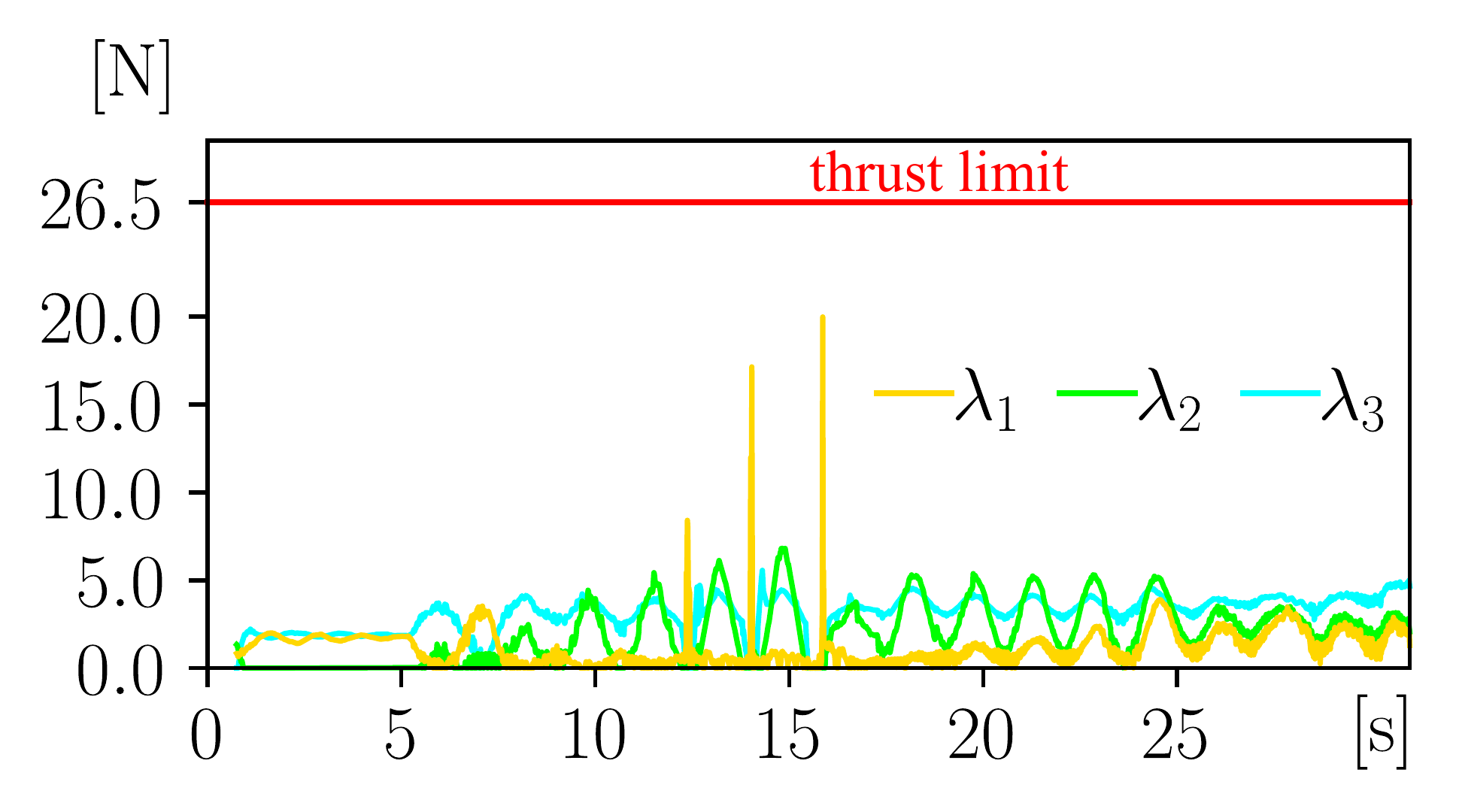}
            \subcaption{Target thrust\revise{s}}
        \end{minipage}%
        \begin{minipage}[t]{0.5\columnwidth}
            \centering
            \includegraphics[width=1.0\columnwidth]{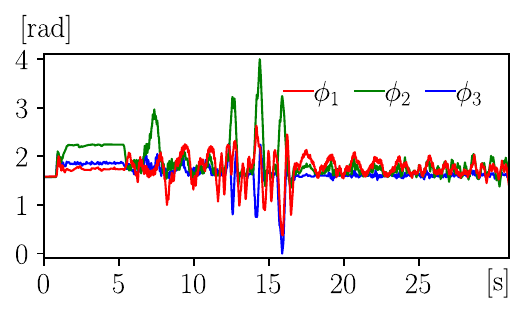}
            \subcaption{Thrust vectoring angles}
        \end{minipage}%
        \\
        \begin{minipage}[t]{0.5\columnwidth}
            \centering
            \includegraphics[width=1.0\columnwidth]{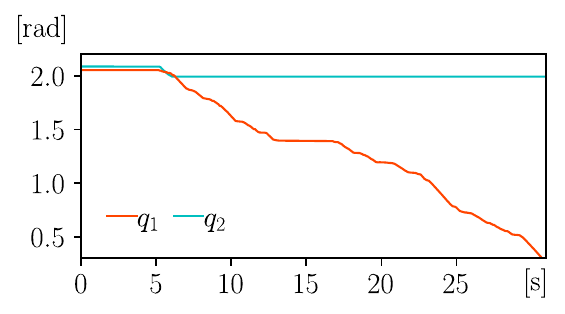}
            \subcaption{Joint angles}
        \end{minipage}%
        \begin{minipage}[t]{0.5\columnwidth}
            \centering
            \includegraphics[width=1.0\columnwidth]{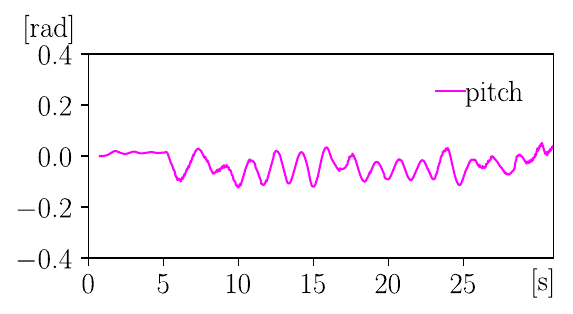}
            \subcaption{Pitch angle error}
        \end{minipage}%
        \\
        \begin{minipage}[t]{0.5\columnwidth}
            \centering
            \includegraphics[width=1.0\columnwidth]{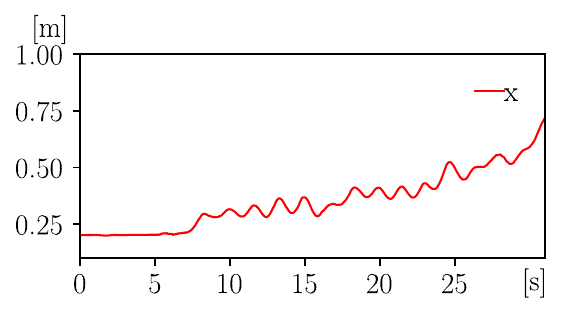}
            \subcaption{End effector \(x\) position in \revise{\(\{W\}\)}}
        \end{minipage}%
        \begin{minipage}[t]{0.5\columnwidth}
            \centering
            \includegraphics[width=1.0\columnwidth]{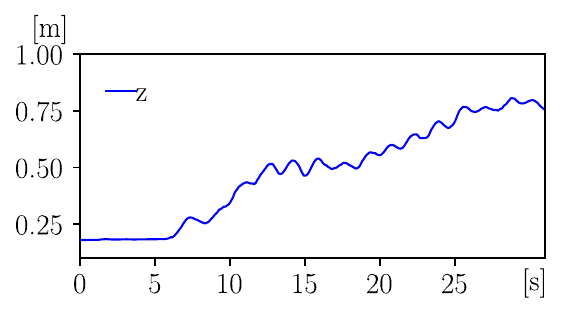}
            \subcaption{End effector \(z\) position in \revise{\(\{W\}\)}}
        \end{minipage}%
    \end{tabular}
    \caption{Plots related to \figref{figure:button_push_experiment}.}
    \label{figure:button_push_plot}
    \vspace{-5mm}
\end{figure}

\begin{figure}[t]
    \centering
    \includegraphics[width=1.0\linewidth]{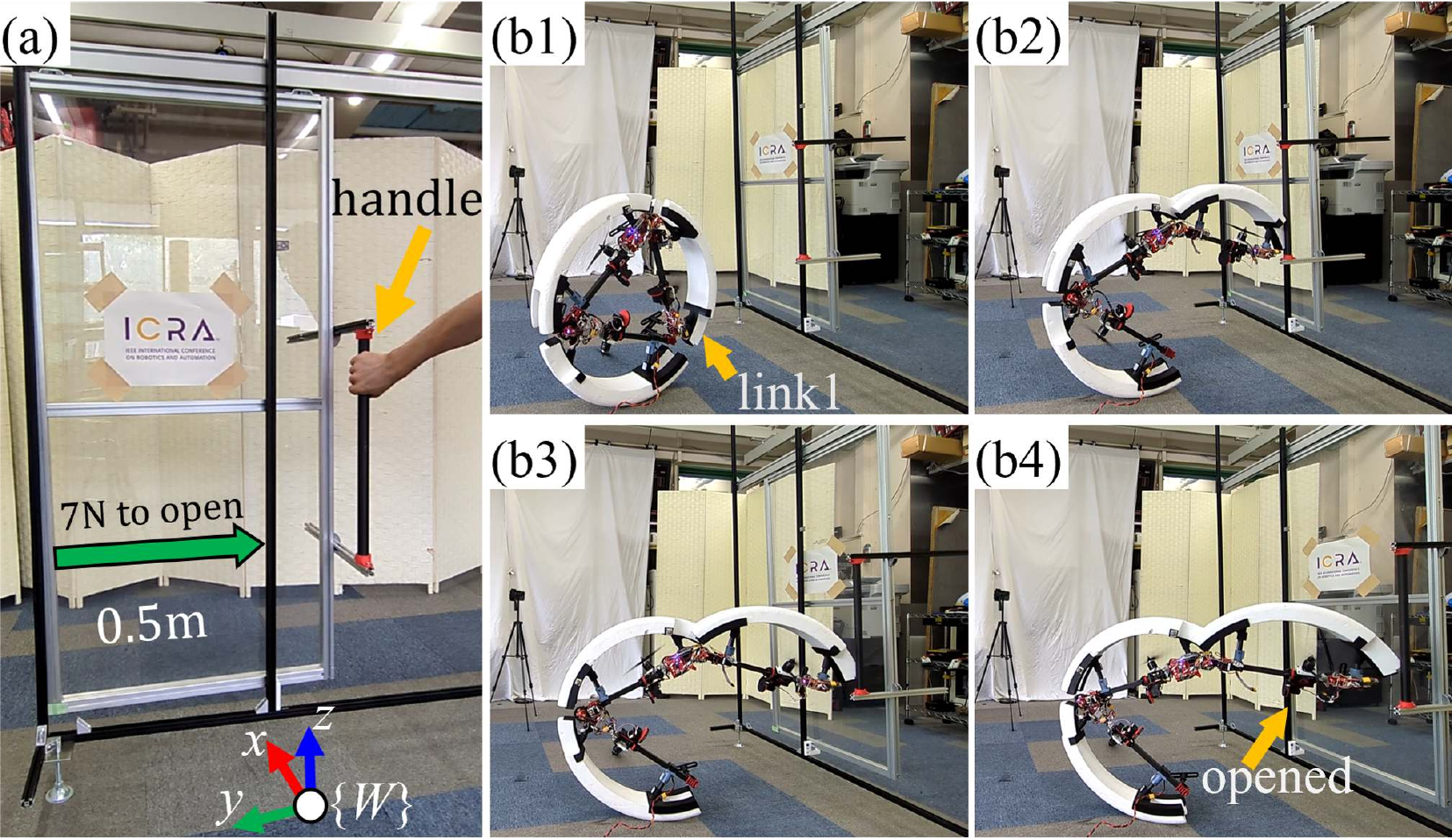}
    \caption{Door opening experiment}
    \label{figure:door_opening_experiment}
    \vspace{-5mm}
\end{figure}

\begin{figure}[t]
    \begin{tabular}{ll}
        \begin{minipage}[t]{0.5\columnwidth}
            \centering
            \includegraphics[width=1.0\columnwidth]{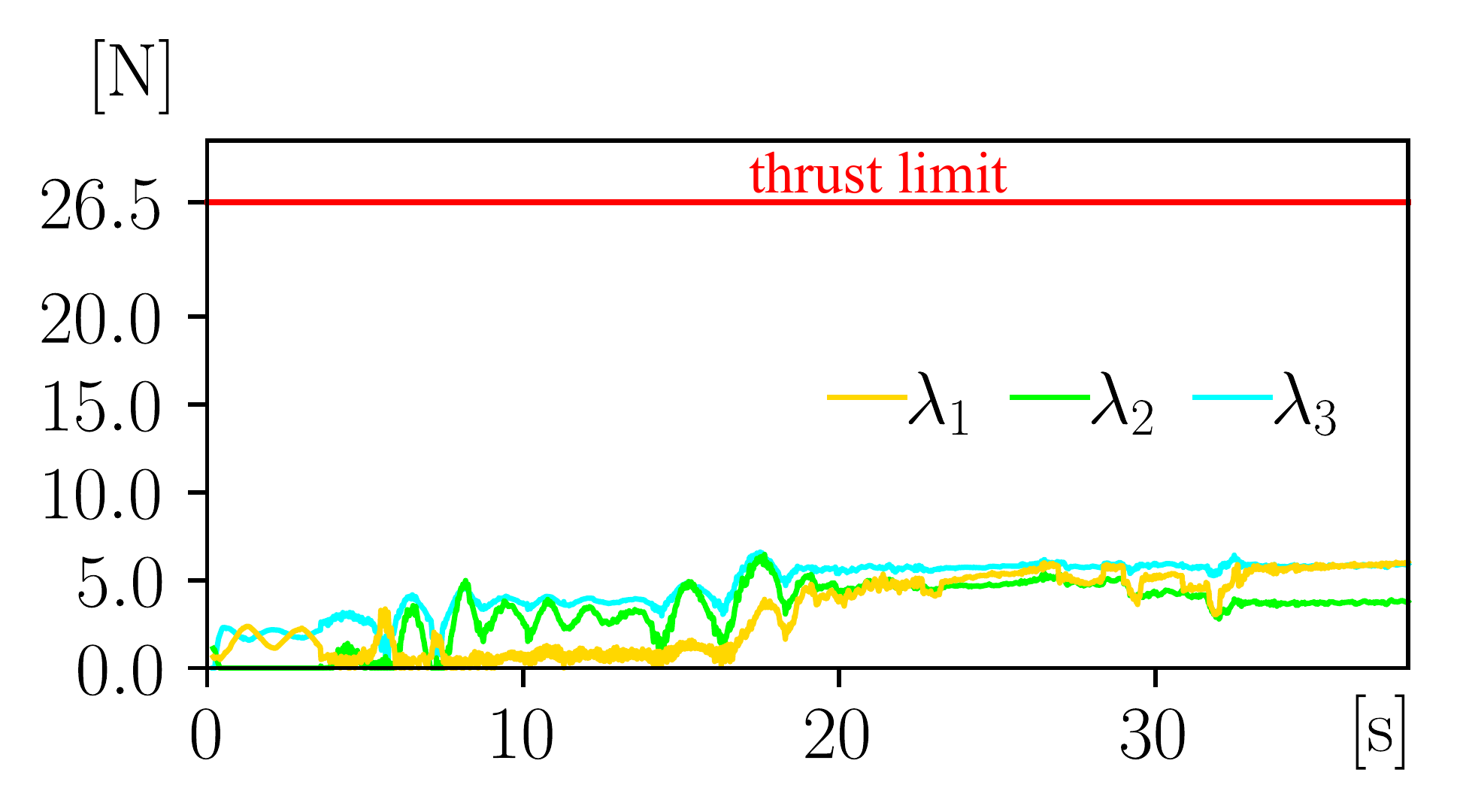}
            \subcaption{Target thrust\revise{s}}
            \label{figure:door_opening_plot::pwm}
        \end{minipage}%
        \begin{minipage}[t]{0.5\columnwidth}
            \centering
            \includegraphics[width=1.0\columnwidth]{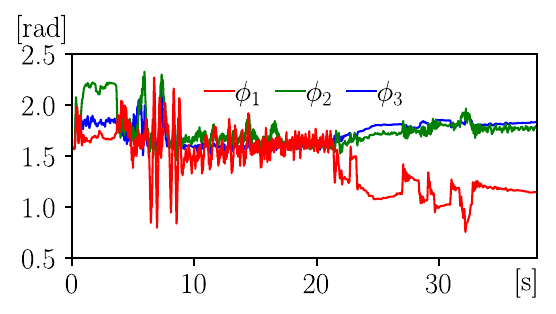}
            \subcaption{Thrust vectoring angles}
            \label{figure:door_opening_plot::gimbal_angle}
        \end{minipage}%
        \\
        \begin{minipage}[t]{0.5\columnwidth}
            \centering
            \includegraphics[width=1.0\columnwidth]{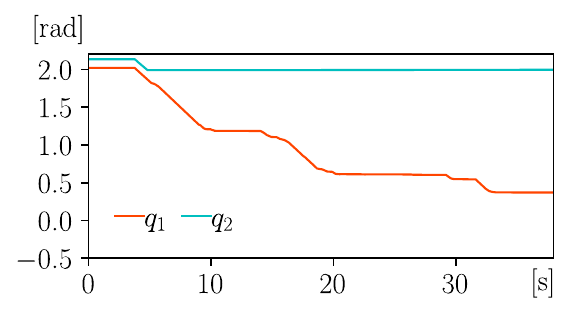}
            \subcaption{Joint angles}
            \label{figure:door_opening_plot::joint_angle}
        \end{minipage}%
        \begin{minipage}[t]{0.5\columnwidth}
            \centering
            \includegraphics[width=1.0\columnwidth]{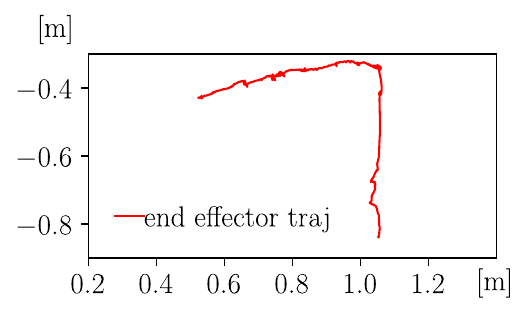}
            \subcaption{End effector trajectory in \(xy\) plane}
        \end{minipage}%
        \\
        \begin{minipage}[t]{0.5\columnwidth}
            \centering
            \includegraphics[width=1.0\columnwidth]{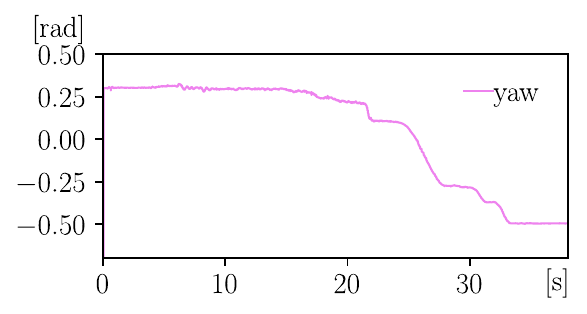}
            \subcaption{\revise{Yaw angle of \(\{CoG\}\) frame}}
        \end{minipage}%
        \begin{minipage}[t]{0.5\columnwidth}
            \centering
            \includegraphics[width=1.0\columnwidth]{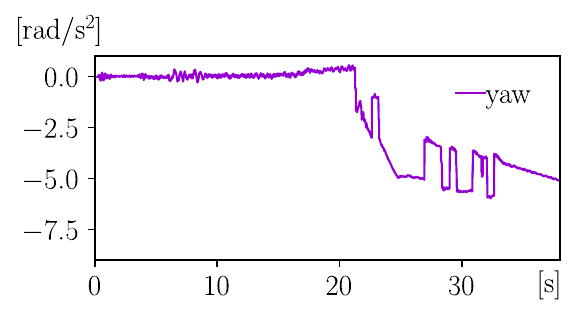}
            \subcaption{Target acceleration of yaw angle}
            \label{figure:door_opening_plot::yaw_pid_total}
        \end{minipage}%
     \\
     \begin{minipage}[t]{0.5\columnwidth}
      \centering
      \includegraphics[width=1.0\columnwidth]{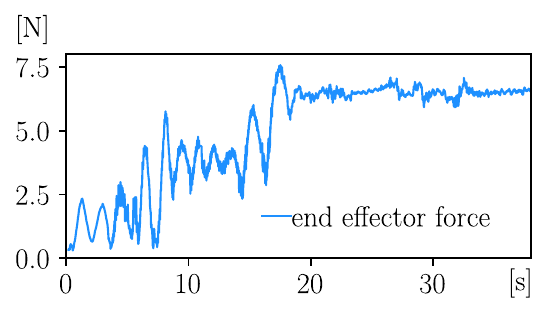}
      \subcaption{\revise{Estimated magnitude of force exerted at end-effector.}}
        \label{figure:door_opening_plot::external_force}
     \end{minipage}
    \end{tabular}
    \caption{Plots related to \figref{figure:door_opening_experiment}.}
    \label{figure:door_opening_plot}
    \vspace{-5mm}
\end{figure}

\subsection{Manipulation Ability Evaluation}
\label{section:manipulation_experiment}
\subsubsection{Object Grasping Flight Experiment}
\switchlanguage{
To \revise{evaluate} the manipulation ability in aerial domain using the joint motion, an object grasping \revise{flight} experiment as shown in \figref{figure:object_grasping_experiment}, was conducted.
During hovering, a \SI{200}{g} rectangular object with dimensions of \SI{200}{mm} \(\times\) \SI{150}{mm} \(\times\) \SI{100}{mm} was \revise{handed to the robot} by \revise{a} human.
Then, by actuating the joints \revise{to sandwich} the object between \revise{the two end} links, the robot grasped it.
\revise{After grasping}, the human released the object.
At the moment when the human released the object, \revise{previously measured object's inertia} was added to the control model to compensate for the influence of the grasped object.
Then, the target values of the \(xy\) position and yaw angle were changed to verify \revise{the ability to transport the} object.
The joint angles were set to be slightly narrower than the \revise{object width at contact}, and the current limit of the servo motor was set.
To ensure sufficient compliance of the object, a sponge was attached.

The plot\revise{s} in \figref{figure:object_grasping_plot} show the joint angles, joint torques, target thrust of each thruster, and the target and actual values of the \(xy\) position and yaw angle.
Note that the \revise{the range of} yaw angle in \figref{figure:object_grasping_plot:cog_ref_curr} is clamped to \(\qty[-\pi, \pi]\).
At around \(t=14s\), the joint torque increased as the object was sandwiched and the contact began.
At \(t=17s\), when the human released the object, the outputs of \revise{thrusters on first and third links} which are near the object, increased, and the thrust of the second link decreased.
Even at this moment, by adding the inertia of the object to the control model, the grasping state of the object was maintained without divergence of the position and attitude control.
After that, the commands for \(xy\) position and yaw angle were sent, and the robot successfully followed the target values while grasping the object.
Through this experiment, the tracking errors of CoG position and orientation were under \SI{0.15}{m} and \SI{0.15}{rad}, respectively.
}{
    関節変形能力を利用した, 空中での操作能力を実証するため, \figref{figure:object_grasping_experiment}に示す, 空中で物体を受け取る実験を行った.
    ホバリング中に\SI{200}{mm}\(\times\)\SI{150}{mm}\(\times\)\SI{100}{mm}, \SI{200}{g}の直方体の直方体を関節を変形させて両端のリンクで挟み込み, 人間が手を離す.
    物体を渡す人間が手を離した瞬間に制御の運動モデルに同一の慣性を有する剛体を追加し, 把持物体の影響を補償し始める.
    そのあと, \(xy\)方向, yaw角の目標値を変化させ, 物体を運搬可能かを検証した.
    なお, 関節角度は実際に物体が接触するよりも少し狭くなるように設定しつつ, サーボモータに電流のリミットを設定した.
    物体には十分なコンプライアンスを確保するためにスポンジを貼り付けた.

    \figref{figure:object_grasping_plot}に示すプロットは, 関節角度, 関節トルク, PWM値, \(xy\)位置とyaw関する目標値と実値を示している.
    なお, \figref{figure:object_grasping_plot:cog_ref_curr}におけるyaw角は\(\qty[-\pi, \pi]\)の範囲になる.
    \(t=14s\)付近で物体を挟み込み接触を始めた時点で, 関節トルクが増加している.
    \(t=17s\)において, 人間が手を離すと同時に物体に近い位置にある1番目と3番目のリンクの推力が増加し, 2番目のリンクの推力は減少した.
    なお, この瞬間においても, 物体の慣性を制御モデルに追加することで, 位置や姿勢の制御が発散することなく物体の把持状態を維持することができた.
    また, その後, \(xy\)位置やyaw角に関する司令を送ったが, 物体を把持した状態で, 目標値に追従することができた.
}

\begin{figure}[t]
    \centering
    \includegraphics[width=1.0\linewidth]{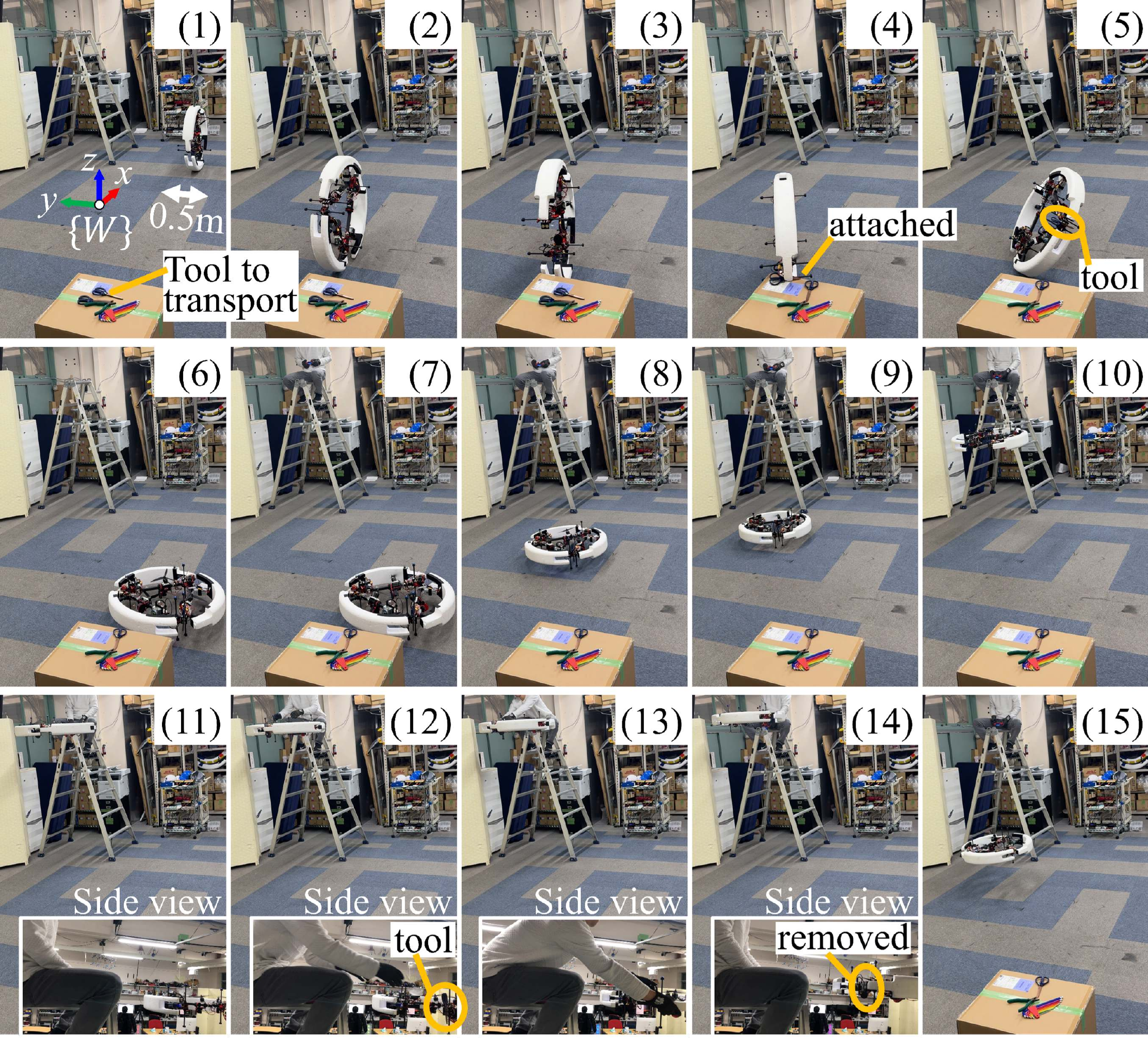}
    \caption{Object transportation experiment.}
    \label{figure:object_transportation_experiment}
    \vspace{-5mm}
\end{figure}

\begin{figure}[t]
    \begin{tabular}{c}
        \begin{minipage}[t]{0.5\columnwidth}
            \centering
            \includegraphics[width=1.0\columnwidth]{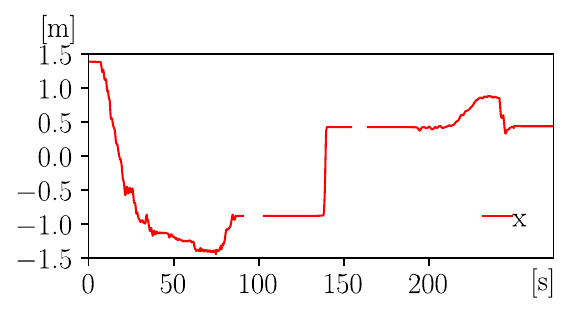}
            \subcaption{CoG \revise{\(x\)} position}
            \label{fig::object_transportation::cog_x_pos}
        \end{minipage}%
        \begin{minipage}[t]{0.5\columnwidth}
            \centering
            \includegraphics[width=1.0\columnwidth]{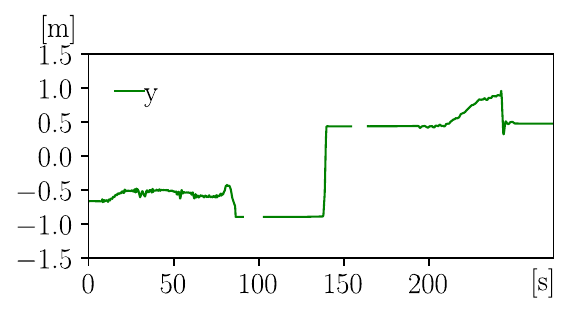}
            \subcaption{CoG \revise{\(y\)} position}
            \label{fig::object_transportation::cog_y_pos}
        \end{minipage}%
        \\
        \begin{minipage}[t]{0.5\columnwidth}
            \centering
            \includegraphics[width=1.0\columnwidth]{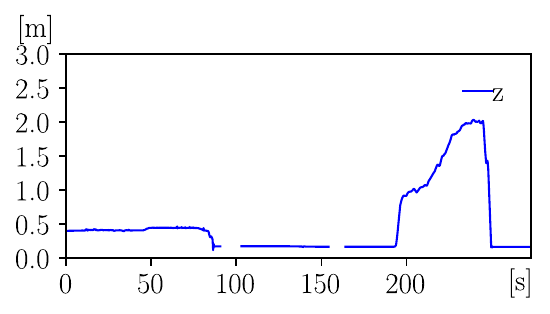}
            \subcaption{CoG \revise{\(z\)} position}
            \label{fig::object_transportation::cog_z_pos}
        \end{minipage}%
        \begin{minipage}[t]{0.5\columnwidth}
            \centering
            \includegraphics[width=1.0\columnwidth]{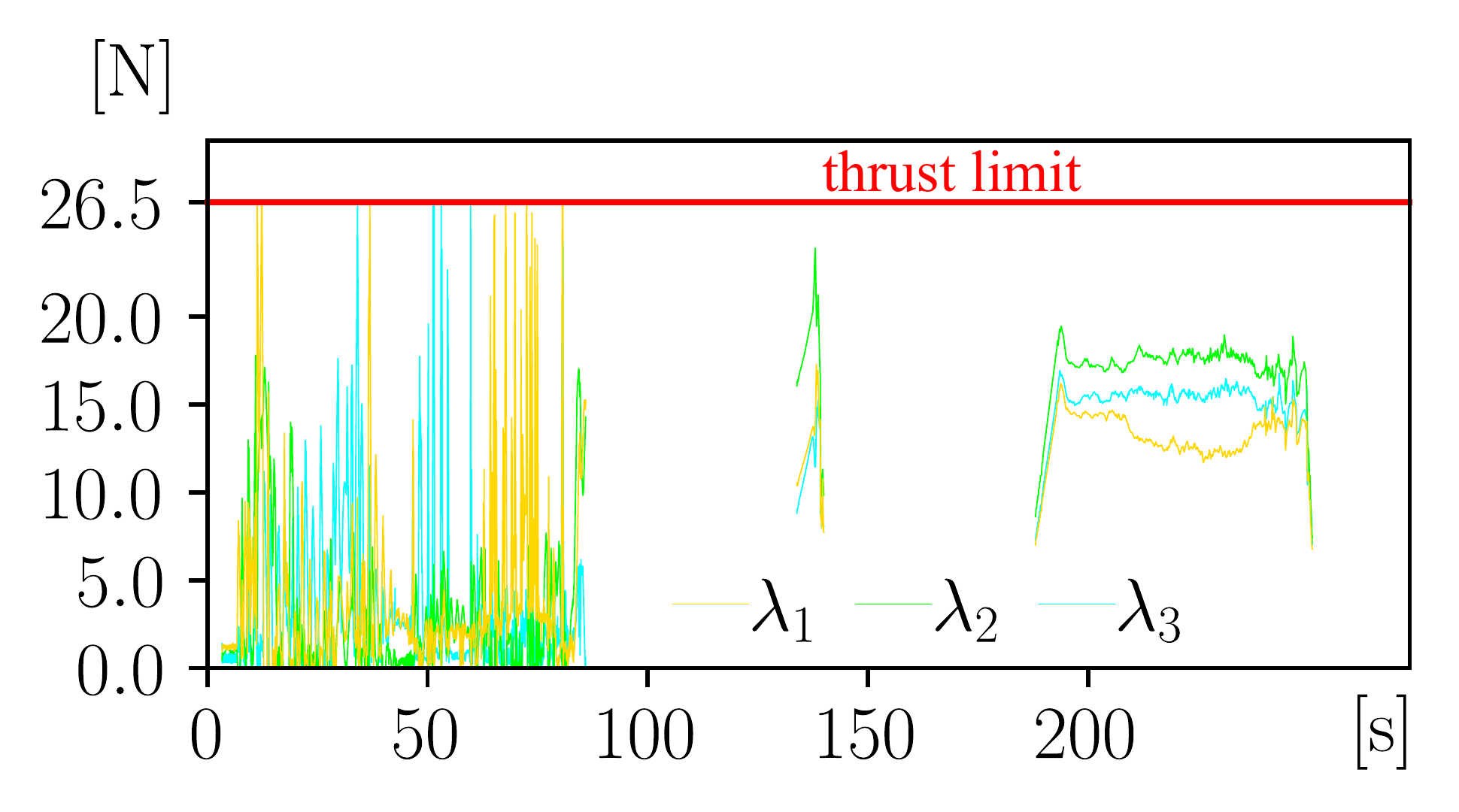}
            \subcaption{\revise{Target thrusts}}
            \label{fig::object_transportation::pwm}
        \end{minipage}
    \end{tabular}
    \caption{Plots related to \figref{figure:object_transportation_experiment}.}
    \label{fig::object_transportation_plot}
    \vspace{-5mm}
\end{figure}

\subsubsection{Button Push Experiment}
\label{section:button_push_experiment}
\switchlanguage{
Based on the proposed thrust control and fullbody inverse kinematics, \revise{a button push experiment} on the ground, as shown in \figref{figure:button_push_experiment}, was conducted to evaluate whether \revise{such a} precise manipulation task can be achieved.
The button size was \SI{50}{mm} in diameter, and the height from the ground was \SI{0.78}{m}.
The target position of the end\revise{-}effector was updated sequentially by teleoperation using a joystick.

By updating the target position of the end effector by teleoperation, the \revise{robot successfully pushed the button} as shown in \figref{figure:button_push_experiment}(4).
\revise{Considering that the button diameter was} \SI{50}{mm} and \revise{that accuracy of position control} during hovering is less than \SI{0.05}{m}, \revise{this motion} is considered \revise{to have been} realized by stabilizing the \revise{robot's} position \revise{through ground contact} rather than \revise{hovering in the air}.
The plots in \figref{figure:button_push_plot} show target thrust of each thruster, thrust vectoring angles, joint angles, pitch angle tracking error, and the \(x\) and \(z\) components of the end effector in the world frame.
The tracking error of the pitch angle oscillated within a range of about 10 degrees.
The cause is considered to be that the position of the CoG of the robot changes continuously due to the \revise{transformation}.
\revise{In addition, the dead zone for thrust commands below \SI{2.0}{N} is also considered to be a contributing factor.}
\revise{In our prototype}, conversion of the target thrust to the PWM command to the ESC (motor driver for thruster\revise{s}) is performed \revise{on the flight controller, where IMU based attitude estimation and feedback control are executed on a microcontroller.}
\revise{Due to the limited computational resources of this device, the thrust characteristics are modeled by a simple continuous 4th-order polynomial.
Moreover, due to the characteristics of ESCs, thrust commands below \SI{2.0}{N} fall into a dead zone in which the propeller does not rotate.}
Therefore, the modeling error of thrust becomes large when operating in the ground \revise{mode}, where the target thrust\revise{s} tend to be small.
\revise{In our prototype, this dead zone is not explicitly compensated for and remains a limitation.
As future work, we plan to improve the  communication bandwidth and integrate the control architecture so that all process can be executed centrally on the onboard PC, allowing us to explicitly consider the nonlinear thrust characteristics, including this dead zone.}

\revise{
In this experiment, the end-effector was commanded to approach object placed above the robot, and as a result, the motion of the joint \revise{on} end-effector side tended to dominate the overall behavior.
This can be attributed to two main reasons.
First, this is because that large angular changes were required primarily at the distal joint to realize the desired end-effector position.
Second, this is due to the difference in load applied to each joint.
The joint near the ground are heavily loaded by the robot's own weight, and under the PD control in servo, the target torque calculated with constant gains may not always be able to generate sufficient torque to track the reference.
In such situations, the distal joint angle changes first, and the end-effector reaches the target mainly through the motion of that joint, which makes it appear as if only a single joint on the end-effector side is being actively controlled.
}}{
提案した推力制御手法および全身逆運動学に基づき, 精度が必要な操作行動を実現可能か評価するため, \figref{figure:button_push_experiment}に示す, 地上でボタンを押す実験を行った. 
ボタンのサイズは直径\SI{50}{mm}で, 地表からの高さは\SI{0.78}{m}である.
JoyStickを用いてエンドエフェクタの目標位置をテレオペレーションによって逐次更新する.

テレオペレーションでエンドエフェクタの目標位置を更新し続けることで, \figref{chap5::fig::button_push_experiment}(4)に示すようにボタンを押すことができた.
また, このボタンの大きさは直径\SI{50}{mm}であり, ホバリング時の機体の位置制御の精度は\SI{0.05}{m}以下であることから考えて, 飛行により浮遊するのではなく地上に接触をして機体位置を安定化することによってできた操作であると考える.
\figref{chap5::fig::button_push_plot}のプロットは, それぞれ推力司令値, 推力方向角度, 関節角度, ピッチ角誤差, 世界座標系におけるエンドエフェクタの\(x\)および\(z\)成分を示している.
機体重心座標系のpitch角の追従誤差が約10度以下の範囲で振動した.
原因として, 関節の変形によって機体重心の位置が変化し続けることが挙げられる.
また, 目標発揮推力とESC(推力装置用モータドライバ)へのPWM司令値の変換は連続関数である4次多項式へのフィッテイングにより行っているが\SI{1200}{\micro s}以下のPWM司令値ではプロペラが回転しない不感帯になる.
そのため, 目標の発揮推力が小さくなりやすい地上領域での動作時には推力のモデル化誤差が大きくなることも原因だと考えられる.
また, 機体の前方上方向に目標のエンドエフェクタ位置が存在していたため, エンドエフェクタの\(+x\)方向, \(+z\)方向への移動への寄与の大きい関節(joint1)が主に動いた.
}

\begin{figure}[t]
    \centering
    \includegraphics[width=\linewidth]{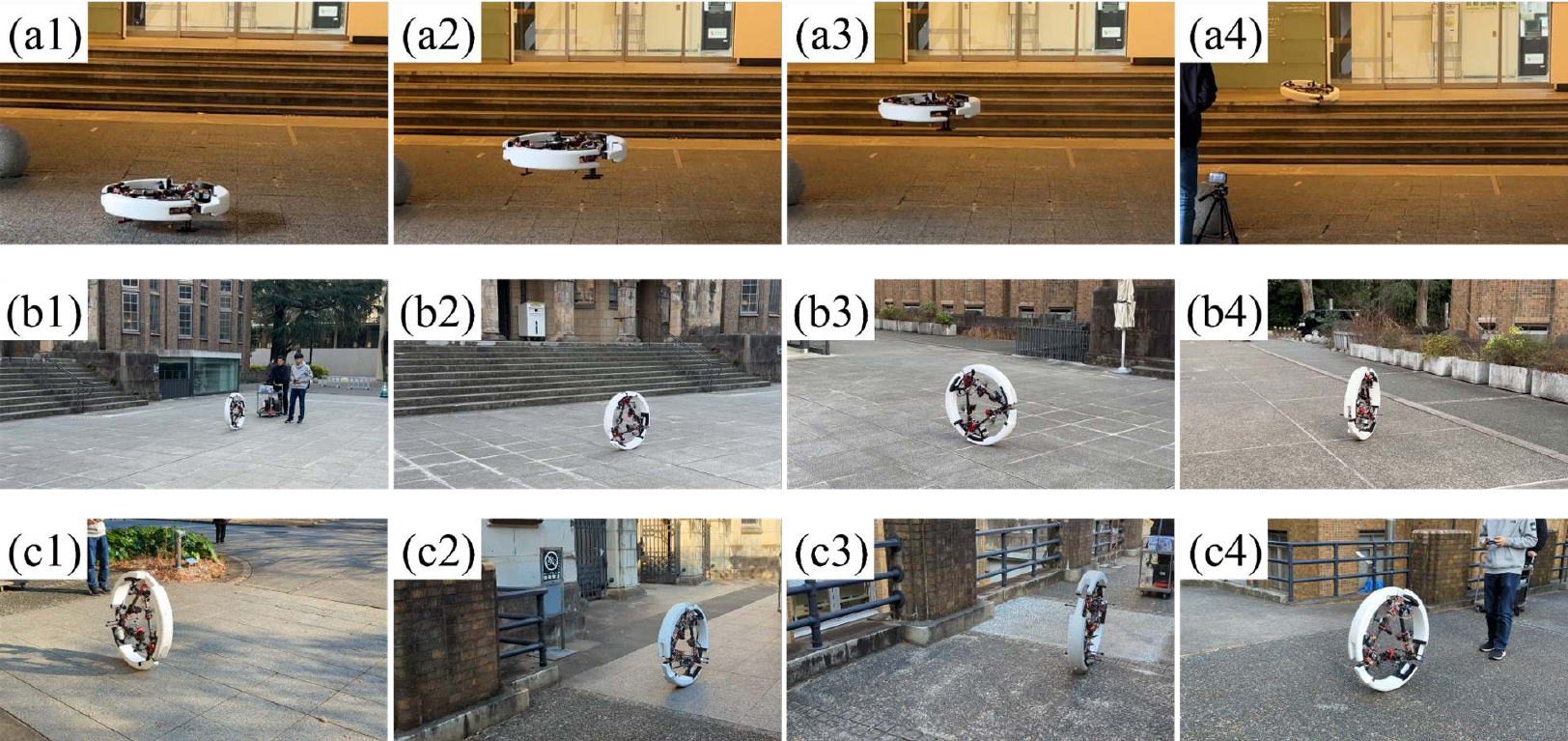}
    \caption{Outdoor locomotion experiments.}
    \label{figure:outdoor_locomotion_experiment}
    \vspace{-5mm}
\end{figure}

\subsubsection{Door Opening Experiment}
\label{section:door_opening_experiment}
\switchlanguage{
To evaluate whether the manipulation task that requires force can be achieved, an experiment of opening a slide door as shown in \figref{figure:door_opening_experiment}(b1)--\figref{figure:door_opening_experiment}(b4), was conducted.
The slide door used for this experiment is shown in \figref{figure:door_opening_experiment}(a), which can be opened with a force of \SI{7}{N} and has an opening width of \SI{0.5}{m}.
The target position of the end\revise{-}effector in the plane \revise{containing the} links was given by teleoperation as in \subsecref{section:button_push_experiment}, and the end effector was \revise{moved} \revise{toward} the door handle.
Furthermore, by sending a command to rotate around the \(z\)-axis of \revise{the world frame}, a force in the translational direction was applied to the door handle.

\revise{The plots in} \figref{figure:door_opening_plot} show target thrust of each thruster, thrust vectoring angles, joint angle\revise{s}, trajectory of the end-effector in the \(xy\)\revise{-}plane of \revise{the world frame}, \revise{yaw angle of \(\{CoG\}\) frame}, angular acceleration exerted around the \(z\)\revise{-}axis of the \revise{\(\{CoG\}\)} frame, \revise{and estimated exerted force at the end-effector}.
\revise{Exerted force shown in \figref{figure:door_opening_plot::external_force} is estimated from the target thrust, configuration of the robot at each time step, and the equilibrium of motion using the least squares method.}
\revise{As shown in \figref{figure:door_opening_experiment},} by sequentially updating the target position of the end effector, even as the door opened and the distance between the door handle and the robot increased, the joint angle\revise{s} w\revise{ere} changed to continue to apply force to the door handle.
Also, by performing thrust control considering the friction cone constraint, \revise{the robot could} perform \revise{this} task with\revise{out slipping}.
By exerting an angular acceleration about \SI{5}{rad/s^2} around the \revise{\(z\)}-axis of the entire robot as shown in \figref{figure:door_opening_plot::yaw_pid_total}, the slid\revise{e} door was opened.
Also, while the door was being opened, as shown in \figref{figure:door_opening_plot::pwm}, the outputs of \revise{the} thrusters on first and third links increased.
The output of the thruster on first link \revise{became} larger because it \revise{was located} far from the contact point and can exert a large yaw moment per unit thrust, which \revise{effectively} contributes to the \revise{door opening motion}.
At the same time, the output of the thruster on third link also increased to satisfy the attitude control constraint of the pitch angle \revise{of the \(\{CoG\}\) frame}.
\revise{
Regarding thrust vectoring angles, vectoring angle of the thruster on first link was drastically changed.
This is because the robot in ground mode generates rotational moment around vertical axis mainly by changing thrust direction.
Besides, the thrust vectoring angles of thruster on second and third links changed in the opposite direction to first one.
Thruster on first link utilized its long moment arm to generate the desired rotational moment, while the other thrusters were considered to be oriented in the opposite direction to balance the roll angle control.}
\revise{Although the robot could conduct this task, joint transformation and whole-body rotation occurred in stages due to teleoperation.
Thus, proposed inverse kinematics method is expected to be improved to consider not only the end-effector position tracking on the plane where the links exist, but also the generation of the whole-body orientation.
}
}{
    地上領域における操作能力において力を要する動作が実現可能か評価するため, \figref{figure:door_opening_experiment}(b1)--\figref{figure:door_opening_experiment}(b4)に示す, スライドドアを開ける実験を行った.
    使用したスライドドアは\figref{figure:door_opening_experiment}(a)に示すもので, \SI{7}{N}の力で開閉することが可能であり, 開きしろは\SI{0.5}{m}である.
    リンクの存在する平面内におけるエンドエフェクタの目標位置は\subsecref{section:button_push_experiment}と同様にテレオペレーションによって与えられ, ドアのハンドルにアプローチさせる.
    さらに, 機体重心をyaw軸周りに回転させる司令を送ることで, スライドドアのハンドルに並進方向の力を作用させた.

    \figref{figure:door_opening_plot}に推力司令値, 推力方向角度, 関節角度, エンドエフェクタの\(xy\)平面上の軌道, エンドエフェクタの\(z\)座標, 接地点座標系\(z\)軸周りに発揮した角加速度を表す.
    エンドエフェクタの目標位置を逐次的に更新することで, ドアが開くにつれドアハンドルと機体の距離が増加しても\figref{figure:door_opening_experiment}, \figref{figure:door_opening_plot::joint_angle}に示すように関節角度を変化させ, ドアハンドルに力を継続的に作用させることができた.
    また, 摩擦制約を考慮した推力制御を行っているため, 機体位置を安定化させながら力を要する操作を行うことができた. 
    \figref{figure:door_opening_plot::yaw_pid_total}に示すように, 接地点座標系\(z\)軸周りに約\SI{5}{rad/s^2}の角加速度を発揮することで, スライドドアを開くことができた.
    また, ドアを開いている最中は\figref{figure:door_opening_plot::pwm}に示すように1番目と3番目の推力装置の出力が大きくなった.
    1番目の推力装置は接地点から遠く, 単位推力あたりに発揮可能な機体全体のyaw角まわりの回転モーメントが大きく, ドアを開く動作に寄与しやすいためであると考えられる.
    それと同時にpitch角の姿勢維持制約を満たすために, 3番目の推力装置の出力も大きくなったのだと考えられる.
    また, \figref{figure:door_opening_plot::yaw_pid_total}の機体全体のyaw角周りの角加速度変化と\figref{figure:door_opening_plot::gimbal_angle}の推力方向角変化を比較すると, yaw角周りの角加速度の発揮に伴い1番目の推力方向角が変化しており, 推力方向角変化を物体操作行動に有効に利用できていることがわかる.
}

\subsection{Integrated Motion Experiment}
\label{section:integrated_experiment}
\switchlanguage{
To \revise{demonstrate} the integrated locomotion and manipulation \revise{capabilities of} the proposed system, an object transportation experiment as shown in \figref{figure:object_transportation_experiment} was conducted.
The experimental environment and the definition of the world frame are \revise{illustrated} in \figref{figure:object_transportation_experiment}(1).
In this experiment, \revise{the} robot picked up a tool on the ground and \revise{took} it to a person on a ladder about \SI{1.8}{m} high.
To attach the object to the robot, a permanent magnet was \revise{attached} to the end effector.
The flow of the experiment is as follows.
\begin{itemize}
    \item Approach the object to be transported by rolling on the ground (\figref{figure:object_transportation_experiment}(1)--\figref{figure:object_transportation_experiment}(2)).
    \item Attach the object to be transported (scissors) to the end effector with a permanent magnet by controlling \revise{the} joint angles and whole-body attitude (\figref{figure:object_transportation_experiment}(3)--\figref{figure:object_transportation_experiment}(4)).
    \item Switch to the flight mode by tilting the robot and transforming to standard joint angles (\figref{figure:object_transportation_experiment}(5)--\figref{figure:object_transportation_experiment}(6)).
    \item Fly to the person on the ladder while changing the joint angles to open the link structure (\figref{figure:object_transportation_experiment}(7)--\figref{figure:object_transportation_experiment}(10)).
    \item The person on the ladder take the tool (\figref{figure:object_transportation_experiment}(11)--\figref{figure:object_transportation_experiment}(14)).
    \item \revise{The robot move} away from the person on the ladder and land (\figref{figure:object_transportation_experiment}(15)).
\end{itemize}
\revise{The} plots in \figref{fig::object_transportation_plot} show the \(x\), \(y\), and \(z\) position of the CoG and the \revise{target thrust of each} thruster during the experiment.

In this experiment, the implemented robot platform, with \revise{both} terrestrial and aerial locomotion \revise{capabilities}, \revise{achieved} object transportation to a high place by utilizing \revise{its} joint \revise{transformation capability}.
\revise{In addition}, by comparing the exerted thrust, \revise{we} demonstrated that the exerted \revise{thrust} is smaller in \revise{the} terrestrial domain than in the aerial domain.
\revise{By utilizing ground contact, the robot could stabilize its position, allowing the end-effector to reach the relatively small object.}
}{
提案システムによるロコモーションとマニピュレーションの統合動作の検証のため, \figref{figure:object_transportation_experiment}に示す, 物体運搬行動実験を行った.
実験環境および世界座標系の定義を\figref{figure:object_transportation_experiment}(1)に示す.
本実験では, 地上の台上にある工具をピックアップし, 約\SI{1.8}{m}の高さの脚立の上に居る人に渡す行動を実現する.
運搬対象物体を機体につけるため, エンドエフェクタに永久磁石を取り付けた.
実験の流れは以下である.
\begin{itemize}
    \item 地上を転がり移動することで運搬する運搬対象物体に接近する(\figref{figure:object_transportation_experiment}(1)--\figref{figure:object_transportation_experiment}(2)).
    \item 関節変形と全身の姿勢変化を利用することで, エンドエフェクタに搭載した永久磁石に運搬対象物体(はさみ)をくっつける(\figref{figure:object_transportation_experiment}(3)--\figref{figure:object_transportation_experiment}(4)).
    ただし, 本実験における運搬対象物体は機体質量と比較して十分軽く, 制御への影響が小さいとみなし, 制御モデルへの反映を行わなかった.
    \item 標準関節角度へ変形しつつ機体を倒し, 飛行時の姿勢に遷移する(\figref{figure:object_transportation_experiment}(5)--\figref{figure:object_transportation_experiment}(6)). 
    \item 脚立上の人の側まで飛行しつつ, 関節を変形させてリンクを開く(\figref{figure:object_transportation_experiment}(7)--\figref{figure:object_transportation_experiment}(10)).
    \item 脚立上の人が工具をとる(\figref{figure:object_transportation_experiment}(11)--\figref{figure:object_transportation_experiment}(14)).
    \item 脚立上の人から離れ, 着陸する(\figref{figure:object_transportation_experiment}(15)).
\end{itemize}
\figref{fig::object_transportation_plot}に実験中のCoGの\(x\),\(y\),\(z\)座標, および推力装置へのPWM司令値を示す.

本実験では, 地上と空中での移動能力を有する本ロボットプラットフォームにより, さらに地上での操作能力を利用して, 地上にある物体の高所への運搬を実現した.
また, 発揮した推力を比較すると, 地上領域での行動時には空中領域での動作時と比較して発揮力が小さく出来るということを一連の実験により実証した.
本実験で運搬対象とした物体は小型なため, 地面との接触によって機体位置が安定していることにより, エンドエフェクタを目標に近づけることができたと考える.
}

\subsection{Outdoor Locomotion Experimen\revise{t}s}
\switchlanguage
{
Finally, to verify locomotion ability in \revise{an} actual environment, outdoor locomotion experiments were conducted using the implemented prototype as shown in \figref{figure:outdoor_locomotion_experiment}.
The robot was able to fly over actual stairs, as shown in \figref{figure:outdoor_locomotion_experiment}(a1)--(a4).
Also, as shown in \figref{figure:outdoor_locomotion_experiment}(b1)--(b4) and (c1)--(c4), the robot was able to roll over on the actual ground and move a distance of more than \SI{50}{m}.
\revise{Although we assumed that the robot works on rigid flat ground, it could traverse some uneven terrain because PID control was applied for attitude control.}
The robot was able to perform locomotion in outdoor environment using sensors on the body for \revise{state} estimation.
However, the accuracy of \revise{state} estimation decreased in outdoor environments compared to indoor environment, and sometimes the motion was unstable due to the influence of wind and uneven ground.
}{
実装したプロトタイプを用いて, 屋外でのロコモーション実験を行った.
にしめすように, 飛行をすることで階段を超えることができた.
また, にしめすように, 転がりにより, \SI{50}{m}以上の距離を移動することができた.

機体上のセンサのみをもちいてegomotion estimationをして動作することができた。
しかし, 屋外環境では屋内環境よりも自己位置推定精度が低下することや,
風の影響や地面の凹凸により不安定化する減少も見られた.
}

\section{Conclusions and Future Work}
\label{section:conclusion}
\switchlanguage{
In this work, we aim\revise{ed} to develop a robot platform with air-ground hybrid locomotion and manipulation \revise{capabilities}.
We propose the design, control and motion \revise{strategies} for this platf\revise{or}m.
Through the experiments, \revise{we} found that locomotion on the ground can be realized with lower energy consumption than in flight, and \revise{we experimentally demonstrated the} effectiveness of combining the\revise{se two modes to} achieve various tasks.
Moreover, \revise{from the analysis of the manipulation limitations, it was found that using contact with the environment can increase the end-effector feasible force, and ground manipulation tasks requiring force were actually achieved}.

\revise{
In this work, our modeling and control approach assumed that the robot works on level ground.
Consequently, the current system has limitations in rolling over large steps and must rely on flight to overcome them.
As future work, the proposed control framework could be extended to handle ground environments with slopes and steps, and the multilink structure could be exploited as leg-like limbs to enable rough-terrain locomotion.
The inverse kinematics method incorporating rolling contact constraints can serve as the basis for progressive loco-manipulation tasks and for path-planning methods with continuous contact and transformation in the confined space.
Moreover, we aim to achieve advanced motion\revise{s} with a configuration \revise{that has more DoFs} and with dynamic joint \revise{transformation} combin\revise{ing joint torque and thrust force.
A motion planning method that takes future flight stability into account will also be investigated to realize advanced tasks.
}}
}{
本研究では, 転がりによる地上移動, 飛行による空中移動, 関節変形によるマニピュレーションが可能な, 環境接触機能をもつマルチリンク型マルチロータを開発した.
提案したシステムに基づき実装したプロトタイプを用いて, 両領域におけるロコモーションとマニピュレーションの能力を実証した.
以下はkey contributiondである.
\begin{enumerate}
    \item 提案したロボットプラットフォームの, リンク数とスラスタのデザインに関する最小構成法を提案した. また, ロボットモデルに基づき, 制御の安定性を向上させるためのスラスタのposeの決定法と, 関節の設計要件を提案した.
    \item 接触と関節変形を考慮した, 非線形最適化に基づく実時間推力制御手法を提案した. これは関節や推力偏向機構の有無によらず他のマルチロータにも適用可能なcontrollerである.
    \item マルチロータによる地上での移動とマニピュレーションのための動作計画法を提案し, 実装したマルチロータは地上と空中のドメインで様々なタスクを実現した.
\end{enumerate}
実機実験をとおして, 地上でのロコモーションは飛行よりも低消費エネルギで実現可能であることがわかり, それらを組み合わせて行動実現することの有効性を示した.
また, マニピュレーション性能の限界解析から, 環境接触を利用することで手先発揮力が大きくできることがわかり, 実際に力を要する地上マニピュレーションタスクを実現した。

今回のモデリングと制御においては水平な地面上での動作を前提としていた.
そのため, 大きな段差を転がって走破するのには限界があり, 飛行して突破することとなる.

今後は環境認識に基づくロコモーションの計画や操作行動の自律化が考えられる.
また, より多自由度な機体構成と推力制御に動的な関節変形制御を組み合わせ,
マルチドメインでの高度な操作行動を実現したいと考える.
}

\addtolength{\textheight}{-0.5cm}   

\section*{\revise{Appendix}}
\subsection{\revise{Approximation of Contact Force}}
\label{sec:contact_force_approx}
\switchlanguage{
\revise{As described in \secref{section:nonlinear wrench allocation},
in the ground mode of the actual machine experiment, the optimization problem was simplified by approximating the contact force \(\bm{f}_{c_0}\) as a function of the control inputs.
In this section, we verify the effectiveness of this approximation through benchmarking.}

\revise{We assume the robot works on a horizontal plane and contacts with the ground at a single point located at the lowest point of the propeller guard.
Under conditions where no slipping occurs, the following relationship exists between the contact force and the exerted force.
}
}{
\secref{section:nonlinear wrench allocation} で述べたように，
実機実験における地上モードでは，接地点である\(\{c_0\}\)に作用する接触力\(\bm{f}_{c_0}\)を制御入力の関数として近似することで，最適化問題の簡略化を行った．
本節では，この近似の効果について，ベンチマークを通して検証する．

ロボットが水平面上で動作し，プロペラプロテクタの最下点にある一点で接地していると仮定する．
このとき，滑りが生じない条件のもとで，接触力と発揮力の間には次の関係が成り立つ．
}
\begin{equation}
\begin{aligned}
 \bm{f}_{c_{\revise{0}}} &= {}^{\{W\}}R_{\{c_{\revise{0}}\}}^{-1} \qty(m\bm{g} - {}^{\{W\}}R_{\{\revise{C}o\revise{G}\}}\bm{f}_{\lambda})\\
 &= m\bm{g} - {}^{\{c_{\revise{0}}\}}R_{\{\revise{C}o\revise{G}\}}\bm{f}_{\lambda}\\
 &= m\bm{g} - {}^{\{c_{\revise{0}}\}}\bm{f}_{\lambda}\revise{.}\label{eq:contact_force}
\end{aligned}
\end{equation}
\switchlanguage{
\revise{Here, we} use the fact that the \(z\)\revise{-}axis of \(\{c_{\revise{0}}\}\) is \revise{oriented} vertically upward.
\(^{\{c_{\revise{0}}\}}\bm{f}_{\lambda}\) represents the force generated by thrust\revise{s} on \(\{c_{\revise{0}}\}\).
\revise{U}sing this \revise{relationship}, the nonlinear optimization problem in \eqref{eq:sqp_control_cost_function}--\eqref{eq:sqp_control_torque_constraint} can be simplified as follows,
}{
ただし\({\{cp\}}\)の\(z\)軸が鉛直上向きを向くことを用いた.
\(^{\{cp\}}\bm{f}_{\lambda}\)は推力によって\(\{cp\}\)に発生する力を表す.
これを利用すると\equref{eq:sqp_control_cost_function}--\equref{eq:sqp_control_torque_constraint}の非線形最適化問題は,
以下のように簡単化できる.
}
\begin{align*}
    &\underset{\bm{\lambda}, \bm{\phi}, \bm{\tau}_{joint}}{\text{minimize}} \hspace{5mm} &&w_1\norm{\bm{\lambda}}^2 + w_2\norm{\bm{\tau}_{joint}}^2 + w_3\norm{\bm{\phi} - \bm{\phi}_{curr}}^2,\\
    & \text{subject to} && \eqref{chap4::eq::joint_motion}, \eqref{chap4::eq::sqp_lambda_limit}, \eqref{chap4::eq::sqp_joint_torque_limit}, \revise{\eqref{eq:ground_mode_attitude_pid}}, \eqref{eq:sqp_control_torque_constraint}, \eqref{eq:contact_force}\\
    &&& {}^{\{c_{\revise{0}}\}}\bm{f}_{\lambda} = {}^{\{c_{\revise{0}}\}}R_{\{\revise{C}o\revise{G}\}}Q_{trans}(\bm{q}, \bm{\phi})\bm{\lambda},\\
    &&& {}^{\{c_{\revise{0}}\}}f_{\lambda, z} < mg\\
    &&& \qty|{}^{\{c_{\revise{0}}\}}f_{\lambda, x}| < \mu \qty(mg - {}^{\{c_{\revise{0}}\}}f_{\lambda, z}),\\
    &&& \qty|{}^{\{c_{\revise{0}}\}}f_{\lambda, y}| < \mu \qty(mg - {}^{\{c_{\revise{0}}\}}f_{\lambda, z}).\label{eq:sqp_friction_cone_y}
\end{align*}
\switchlanguage{
where \(Q_{trans}\qty(\bm{q}, \bm{\phi})\) is the upper \revise{three} rows of \(Q\qty(\bm{q}, \bm{\phi})\) \revise{in \eqref{eq:Q_definition}}.
By \revise{removing} the contact force \revise{from} the optimization variables in \eqref{eq:sqp_control_cost_function}, \revise{we} expect that the problem can be solved faster.
}{
ただし\(Q_{trans}\qty(\bm{q}, \bm{\phi})\)は\(Q\qty(\bm{q}, \bm{\phi})\)の上3行を取り出したものである.
この変形により, \equref{eq:sqp_control_cost_function}で最適化変数に含まれていた接触力を除くことができていて, 高速化が期待できる.
}

\begin{figure}[t]
 \centering
 \begin{tabular}{cc}
  \begin{minipage}[t]{0.48\columnwidth}
   \centering
   \includegraphics[width=1.0\columnwidth]{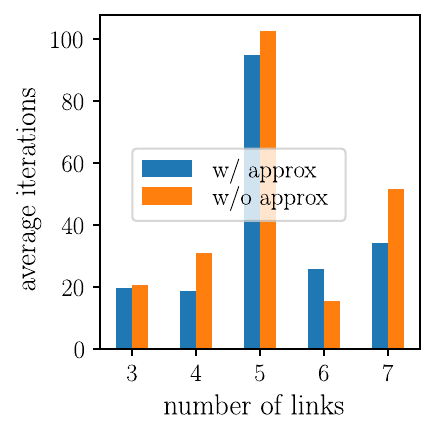}
   \subcaption{Comparison of iterations.}
  \end{minipage}
  \begin{minipage}[t]{0.48\columnwidth}
   \centering
   \includegraphics[width=1.0\columnwidth]{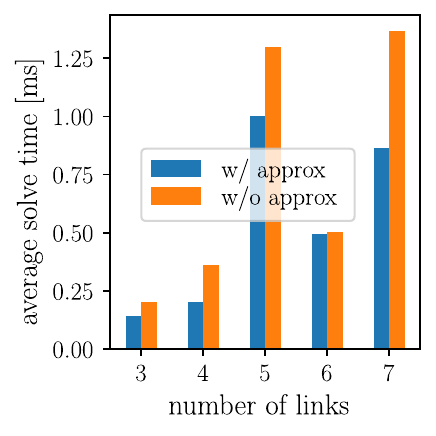}
   \subcaption{Comparison of solve time.}
  \end{minipage}
 \end{tabular}
 \caption{Comparison of the number of SQP iterations and solve time with and without approximating the contact force as a function of control inputs.}
 \label{fig:contact_force_approximation_comparison}
 \vspace{-5mm}
\end{figure}

\switchlanguage{
\revise{We compare following two formulations in terms of the number of SQP iterations and computation time:
(i) the optimization problem in \eqref{eq:sqp_original_cost}--\eqref{chap4::eq::sqp_joint_torque_limit}, where thrusts, thrust vectoring angles, joint torques, and ground contact forces are all treated as optimization variables; and
(ii) the simplified problem in which the ground contact force is approximated as a function of the control inputs, as described above.
In this comparison, we consider not only the three link model developed in this work, but also extended models with up to seven links obtained by concatenating the same link modules in series.
For each condition, we added random noise to the feedback term in \eqref{eq:ground_mode_attitude_pid} and solve the optimization problem 1000 times.
}

\revise{
\figref{fig:contact_force_approximation_comparison} shows the average number of iterations and the average computation time over the trials in which an optimal solution was obtained. In all models, an optimal solution was found in more than \SI{98}{\%} of the trials.
}

\revise{
In most cases, replacing the contact force by a function of the control inputs reduces the number of iterations required for convergence.
Moreover, for all conditions, the computation time is shorter when this approximation is applied.
Even as the number of links increases, the SQP computation converges within \SI{2}{ms}.
Therefore, these results suggest that the proposed control framework is also applicable to other multilink multirotor configurations that involve contact forces and nonlinear constraints induced by thrust vectoring.}
}{
推力・推力方向角・関節トルクに加えて地上との接触力も変数として扱う\eqref{eq:sqp_original_cost}--\eqref{chap4::eq::sqp_joint_torque_limit} の最適化問題と，前述の近似により接触力を制御入力の関数として置き換えた場合とで，SQP による求解におけるイテレーション数および求解時間を比較した．
この比較では，本研究で開発した 3 リンクモデルだけでなく，同じリンクを直列に追加した 7 リンクモデルまでを対象とした．
\eqref{eq:ground_mode_attitude_pid} におけるフィードバック項にランダムノイズを加え，各条件について最適化問題を1000回解いた.
\figref{fig:contact_force_approximation_comparison} に，最適解が得られた試行に対するイテレーション数と求解時間の平均値を示す．
なお，全てのモデルにおいて，試行の \SI{98}{\%} 以上で最適解を得ることができた．

多くの場合，接触力を制御入力の関数として近似した場合のほうが，収束に必要なイテレーション数は少なくなった．
また，全ての条件で，求解に要する時間は近似を用いた場合のほうが短かった．

リンク数が増加した場合でも，SQP の計算は \SI{2}{ms} 以内に収束している．
したがって, 本研究で提案する制御手法が, 接触力と推力偏向に起因する非線形制約を含む他の構成の多リンクマルチロータにも適用可能だとわかった.
}

\bibliographystyle{junsrt}
\bibliography{main}

\end{document}